%% file: main.tex
\documentclass[10pt,twocolumn,letterpaper]{article}

\usepackage[pagenumbers]{cvpr} 
\usepackage{comment}
\usepackage[symbol]{footmisc}
\usepackage{gensymb}
\usepackage{subcaption}
\usepackage{colortbl}
\usepackage{arydshln}
\usepackage{amssymb}
\usepackage{pifont}
\newcommand{\cmark}{\ding{51}}%
\newcommand{\xmark}{\ding{55}}%

\input{preamble}

\definecolor{cvprblue}{rgb}{0.21,0.49,0.74}
\usepackage[pagebackref,breaklinks,colorlinks,citecolor=cvprblue]{hyperref}

\def\paperID{*****} 
\def\confName{CVPR}
\def\confYear{2024}

\title{Free-Moving Object Reconstruction and Pose Estimation with Virtual Camera}

\author{%
{Haixin Shi$^1$\footnotemark[1], ~Yinlin Hu$^2$, ~Daniel Koguciuk$^2$, ~Juan-Ting Lin$^2$, ~Mathieu Salzmann$^{1}$, ~David Ferstl$^2$} \\
{$^1$EPFL, \quad $^2$MagicLeap} \\
Project page: \href{https://haixinshi.github.io/fmov}{https://haixinshi.github.io/fmov} \\
}

\begin{document}

\twocolumn[{%
\renewcommand\twocolumn[1][]{#1}%
\maketitle
\begin{center}
    \centering
    \captionsetup{type=figure}
    \includegraphics[width=\textwidth]{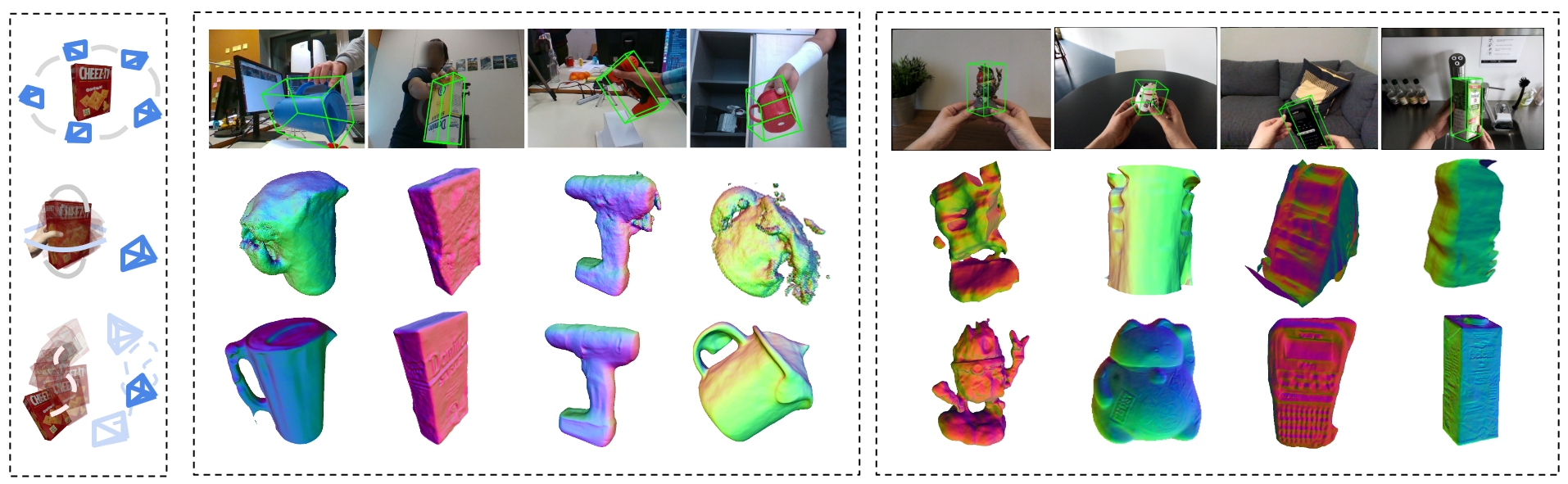}
      \begin{tabular}{m{0.1\textwidth}m{0.45\textwidth}m{0.45\textwidth}}
      \small{(a) Settings} & ~~(b) \small{Hampali~\etal.~\cite{meta-obj} versus ours with fixed camera} & \hspace{-1em}\small{(c) BARF~\cite{barf} versus ours on egocentric sequences} \\
      \end{tabular}
      \vspace{-1.5em}
    \captionof{figure}{
    {\bf Joint object reconstruction and pose estimation from a monocular RGB video.}
    {\bf (a)} {\bf Top}: the standard reconstruction setting with static object captured by a moving camera, which relies on the geometry cue from the whole scene and is inapplicable to dynamic objects. {\bf Middle}: the setting with rotating a hand-held object in front of a fixed camera. {\bf Bottom}: the setting does not assume any prior, which allows the objects to be moved freely with any grasping style. {\bf (b, c)} Our method outperforms state of the art on the HO3D dataset with fixed camera, and produces accurate results on free-moving objects with egocentric views.
    }
    \label{fig:first}
\end{center}%
}]

\maketitle

\footnotetext[1]{Work done as part of Haixin’s Master thesis.}

\input{sec/00_abstract}    
\input{sec/01_introduction}
\input{sec/02_related_work}
\input{sec/03_approach}
\input{sec/04_experiment}
\input{sec/05_conclusion}
{
    \small
    \bibliographystyle{ieeenat_fullname}
    \bibliography{main}
}


\end{document}

%% file: preamble.tex
\usepackage[dvipsnames]{xcolor}

\newif\ifdraft
\draftfalse
\drafttrue

\usepackage{xcolor}
\usepackage{multirow} 
\definecolor{orange}{rgb}{1,0.5,0}
\definecolor{violet}{RGB}{70,0,170}

\newcommand{\HS}[1]{}
\newcommand{\YH}[1]{}
\newcommand{\DK}[1]{}
\newcommand{\JT}[1]{}
\newcommand{\MS}[1]{}
\newcommand{\DF}[1]{}

\newcommand{\bK}{\mathbf{K}}
\newcommand{\bM}{\mathbf{M}}
\newcommand{\bR}{\mathbf{R}}

\newcommand{\bC}{\mathbf{C}}
\newcommand{\bD}{\mathbf{D}}

\newcommand{\bu}{\mathbf{u}}
\newcommand{\bv}{\mathbf{v}}

\newcommand{\bx}{\mathbf{x}}

\newcommand{\bd}{\mathbf{d}}
\newcommand{\bc}{\mathbf{c}}
\newcommand{\bt}{\mathbf{t}}
\newcommand{\br}{\mathbf{r}}

%% file: sec/00_abstract.tex
\begin{abstract}
We propose an approach for reconstructing free-moving object from a monocular RGB video. Most existing methods either assume scene prior, hand pose prior, object category pose prior, or rely on local optimization with multiple sequence segments.
We propose a method that allows free interaction with the object in front of a moving camera without relying on any prior, and optimizes the sequence globally without any segments. We progressively optimize the object shape and pose simultaneously based on an implicit neural representation.
A key aspect of our method is a virtual camera system that reduces the search space of the optimization significantly.  
We evaluate our method on the standard HO3D dataset and a collection of egocentric RGB sequences captured with a head-mounted device. We demonstrate that our approach outperforms most methods significantly, and is on par with recent techniques that assume prior information.



\end{abstract}

%% file: sec/01_introduction.tex
\section{Introduction}
\label{sec:intro}

Understanding 3D objects around us is a fundamental problem in computer vision~\cite{han2021trends}, and also a critical component in many applications, such as augmented reality~(AR)~\cite{Billinghurst2021archallenges} and robot manipulation~\cite{Thalhammer2023ChallengesFM}. This requires an accurate reconstruction and pose estimation of such objects. With a monocular RGB camera, most current work tackle this problem with major simplifications, either by moving the camera around a static object~\cite{unisurf, frodo, yariv2020multiview} or by rotating the object with hands in front of a stationary camera~\cite{rusinkiewicz2002realtime, tzionas2015handobject, weise2008inhandmodelin, weise2009loopclosure}.
In this paper, we investigate a more general setting with the example of an AR device where the object is free-moving in front of a head-mounted camera. We neither assume any object category prior nor any hand prior in this new setting, which allows the objects to be moved in any manner, or freely manipulated with any grasping style if moved by hands, as shown in Fig.~\ref{fig:first}.



\begin{figure}[tb]
  \centering
   \includegraphics[width=\linewidth]{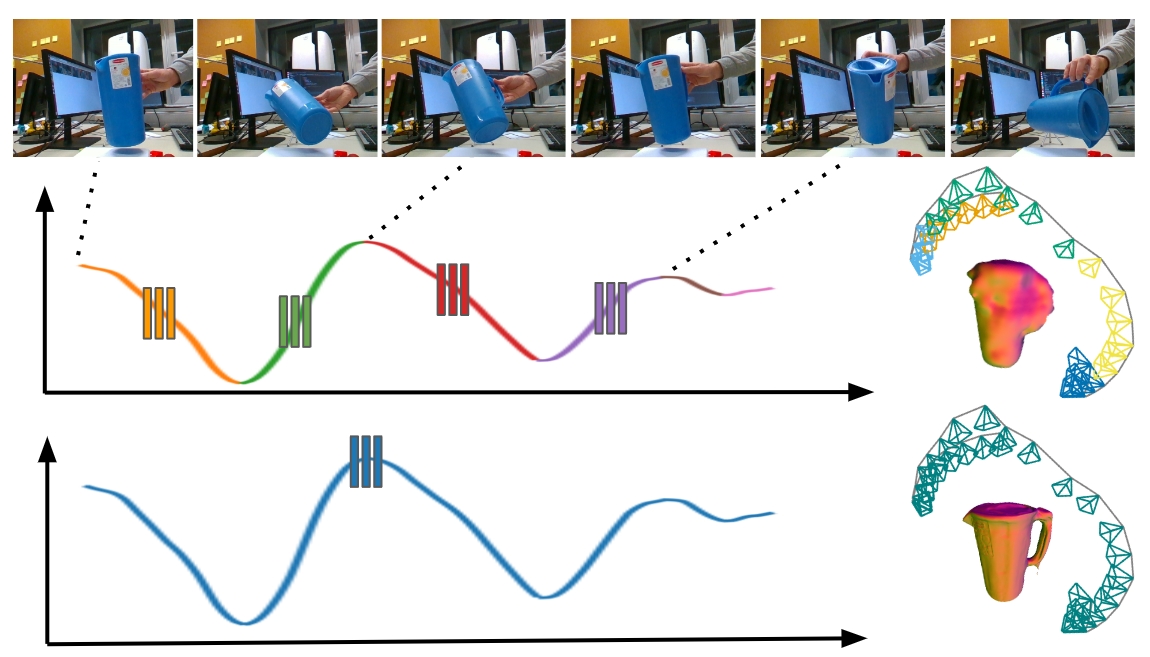}
  \caption{
  {\bf Paradigms of pose-free object reconstruction.} {\bf Top}: The input sequence. {\bf Middle}: Existing method~\cite{meta-obj} relies on segment-wise joint optimization based on multiple easy segments of the sequence, as shown with different colors in the pose trajectories, which tends to be local optimal. {\bf Bottom}: Our method optimizes object shape and pose progressively without any segments, producing globally-consistent shape and pose results.
  }
  \label{fig:compare_paradigm}
\end{figure}


Without pose initialization, some reconstruction methods~\cite{barf, SCNeRF2021, rosinol2022nerf} propose to optimize the shape representation and poses simultaneously. However, most of them either rely on geometry clues of the background which is inapplicable for free-moving objects, or can only handle a restricted range of viewpoints. On the other hand, some recent methods~\cite{meta-obj} use progressive training to solve this problem and rely on segment-wise optimization based on multiple overlapping segments of the sequence. This strategy, however, suffers mainly in two ways. First, the frame selection of each segment is error-prone, since it typically relies on the changes of mask area of the target between consecutive frames, which hardly can be generalized to different object shapes. Second, the segment-wise optimization is inherently local and sub-optimal, as shown in Fig.~\ref{fig:compare_paradigm}.


We propose a method for joint reconstruction and pose estimation of free-moving objects without any segments, which can be globally optimized with a single network.
Our observation is that the unknown pose trajectory of the object can be simplified with a new virtual camera system that always points to the object center with the guidance of 2D object masks, which reduces the search space of the optimization significantly. We first use off-the-shelf 2D segmentation methods to get object masks in each frame, and then optimize the network w.r.t. the virtual camera. To handle the approximation error between virtual camera and real camera, we finally convert the results to real camera coordinate system and refine all the results w.r.t. the real camera. We evaluate our method on both the HO3D dataset~\cite{ho3d} with fixed camera and a collection of data captured using a head-mounted AR device with egocentric views. The experiments show that our method outperforms traditional methods and most baselines significantly.

We summarize our contributions as follows. First, we investigate the problem of existing methods in reconstructing free-moving object from a RGB monocular video. Second, we propose a simple-but-effective strategy with a virtual camera system that simplifies the object trajectory and reduce the search space of the optimization significantly. Finally, we demonstrate the effectiveness of our method on datasets with either a fixed or egocentric camera.

%% file: sec/02_related_work.tex
\begin{figure*}[tb]
  \centering
  \includegraphics[width=0.9\linewidth]{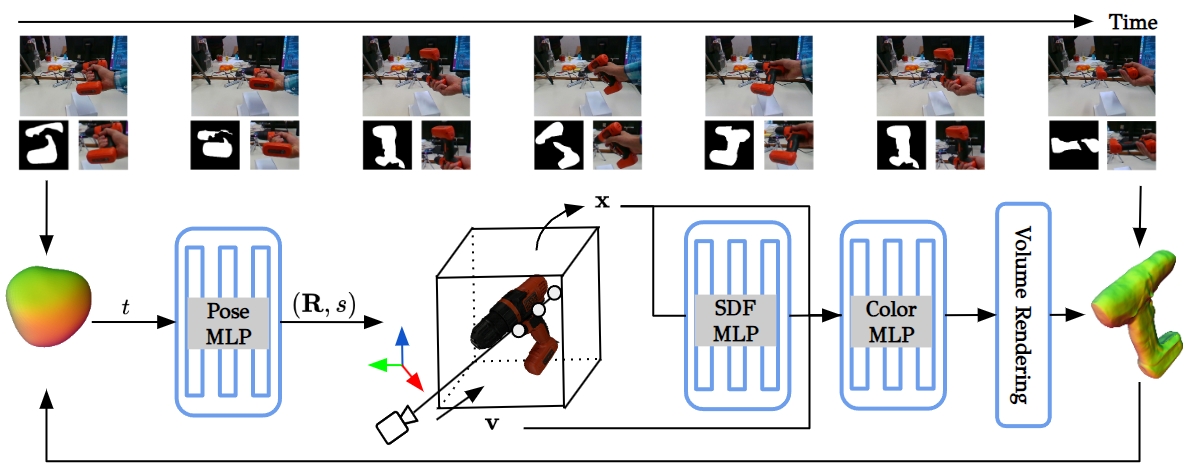}
  \caption{{\bf Overview of our method.}
  We first use off-the-shelf 2D segmentation methods to get object mask in each frame, and then optimize the MLP networks w.r.t. a virtual camera system, with which the camera always points to areas near the object center, as illustrated as colored 3D axis in the figure. We optimize three MLPs with progressively added images. For each frame with time index $t$, we use the Pose MLP to predict the object pose~$(\bR, s)$, which corresponds to the rotation and the distance from the camera center to the object center, summing up to only 4 degrees of freedom. For each 3D point $\bx$ along the view direction $\bv$, we use the SDF MLP and Color MLP to predict its corresponding SDF value and color opacity, respectively. We compare the rendered image with the input and update the MLP networks based on volume rendering. We finally conduct the virtual-to-real conversion and refine all the results w.r.t. the real camera.  
  }
  \label{fig:over_view}
\end{figure*}

\section{Related Work}

\noindent {\bf 3D Reconstruction} is a fundamental problem in computer vision. Traditional methods~\cite{agarwal2011building, mohr1995relative, pollefeys2004visual, snavely2006photo}, typically COLMAP~\cite{colmap}, first estimate camera parameters with 2D matching and then reconstruct the scene with multi-view-stereo~(MVS) techniques~\cite{cheng2021multi}. Most recent methods solve the reconstruction by optimizing a neural implicit representation with rendering techniques~\cite{nerf, d-nerf, neus, neus2}. Although they achieve high-quality reconstruction results, most of them rely on SfM to obtain accurate camera poses for each frame. However, most SfM-based methods assume static rigid scenes and only work when there are enough textures in the scene~\cite{colmap, theia-manual, Bundler}, which is inapplicable in our setting where the object moves independently of the background and the object is often texture-less. Some recent works try to remove the SfM pose initialization by optimizing the neural representation and camera poses simultaneously~\cite{barf, SCNeRF2021, rosinol2022nerf, nope-nerf, nope-neus}. However, they can only work with forward-facing scenarios with a restricted range of view. The recent method~\cite{meta-obj} tries to divide the whole sequence into multiple easy short segments to facilitate the optimization. However, the frame selection of the segmenting procedure is shape dependent and the segment-wise based optimization is local optimal. By contrast, our method is segment-free and can produce globally-consistent results.

\noindent {\bf Object Pose Estimation} aims to produce accurate 3D rotation and 3D translation of the object w.r.t. the camera, which usually serves as pose initialization for dynamic object reconstruction. Most recent object pose methods~\cite{BOP,PFA,WDR,Zebrapose,GDRNet}
first establish 3D-to-2D correspondences via networks and then use a Perspective-n-Points (PnP) solver~\cite{epnp, moreno2007accurate}
to get the pose results. However, most of them rely on the object's 3D mesh, which is one of the goals of reconstruction and is inapplicable in this work. Some recent category-level methods~\cite{category_pose_1, category_pose_2, category_pose_3} do not rely on the target's mesh explicitly by training the network with mixed data from different instances of the same category. However, they are limited to known categories in the training set and can not be generalized. By contrast, we do not rely on any category prior of the object but optimize the shape, color, and pose of the target simultaneously. On the other hand, most hand-held reconstruction methods~\cite{hhor-handold,hold} rely on the hand-object interaction, and most of them assume a firmly-grasping of the object during the whole capture to leverage hand pose estimators for pose initialization. Differently, our method does not make any assumptions about specific grasping styles and is effective for free-moving objects.

%% file: sec/03_approach.tex
\section{Approach}
The goal of our method is to reconstruct a rigid dynamic object from a sequence of RGB images captured with a calibrated camera. We first explain our capturing and data preprocessing setup in Sec.~\ref{sec:method_setup}. We then present the object representation strategy in Sec.~\ref{sec:method_neus}, and the virtual camera system in Sec.~\ref{sec:virtual_camera}, which reduces the search space of optimization significantly. After that, we discuss segment-free progressive training in Sec.~\ref{sec:method_progressive}. We finally discuss the refinement w.r.t. the real camera and the implementation details in Sec.~\ref{sec:method_global} and Sec.~\ref{sec:impmentation_details}, respectively. We show the overview of our method in Fig.~\ref{fig:over_view}.

\subsection{Capture Setup and Data Pre-Processing}
\label{sec:method_setup}

We capture a sequence of RGB images of dynamic objects with either a fixed or egocentric camera, where the relative pose between the camera and the object is unknown.
We do not assume any object prior or any hand pose prerequisites, which allows the users to rotate the object in any grasping style or even switch between different hands during capture. Users only need to ensure the object is within the field of view of the camera, and all sides of the object are covered during capture to have a full reconstruction.

Our method relies on 2D object masks to separate the object from the background. We obtain object mask of the first frame with some simple clicks based on an interactive segmentation method~\cite{sam}, and then obtain all the object masks in the following frames with a segmentation tracking method~\cite{xmem}.

\subsection{Learning Object Representation}
\label{sec:method_neus}

We use the Signed Distance Function~(SDF)~\cite{neus, volsdf, unisurf} as an implicit representation for the object surface, where the surface of the object is given by the zero-level set of its SDF. We learn the SDF from an image sequence, and compare the input images with volume-rendered images~\cite{neus} after converting the SDF to a radiance field. The final surface mesh is extracted by Marching Cubes~\cite{marching-cubes}.

We follow NeuS~\cite{neus} and use MLP networks to learn the object surface and appearance: 
\begin{equation}
(d(\bx),\bc(\bx,\bv)) = F_\theta(\bx,\bv),
\end{equation}
where $F_\theta$ is the MLP networks. For the 3D location $\bx$ and viewing direction $\bv$, $F_\theta$ predicts their SDF value $d(\bx)$ and RGB colors $\bc(\bx,\bv)$. In practice, we use two different MLPs to predict the surface and color field~\cite{neus}, and apply positional encoding~\cite{nerf, InstantNGP} to $\bx$ and $\bv$ to capture high-frequency signals.

We use volume rendering~\cite{volume_rendering} to optimize the implicit representation. The rendered color of each pixel is an integration of colors along the camera ray $\br$ passing through the pixel, which is usually numerically approximated in practice using quadrature~\cite{nerf}:
\begin{align}
\hat{\bC}(\br) = \sum_{k=1}^{N}w_k\bc_k
\label{eq:blending}
\end{align}
with the alpha-blending coefficient $w_k = \text{exp}(-\sum_{i=1}^{k-1}\delta_i\sigma_i)(1-\text{exp}(-\delta_k\sigma_k))
$, where $N$ is the number of sampled points along the ray $\br$, $\delta_k$ is the 3D distance between adjacent sampled points $\bx_{k}$ and $\bx_{k+1}$, and $\sigma_k$ is the volume density of $\bx_k$ after a transformation of its signed distance $d_k$~\cite{neus}. During training, the network parameters are learned using multi-view images with photometric loss:
\begin{align}
\mathcal{L}_{color} = \sum_{\br \in \mathcal{R}}\|\hat{\bC}(\br)-{\bC}(\br)\|
\label{eq:color_loss}
\end{align}
to measure the difference between the rendered color $\hat{\bC}(\br)$ and the observed color ${\bC}(\br)$ of a pixel intersected by the ray $\br$, where $\mathcal{R}$ is the set of camera rays going through sampled pixels. We use the same Eikonal loss and mask loss as in NeuS for network regularization.

\begin{figure}[tb]
  \centering
    \begingroup
    \begin{tabular}{ccc}

   \includegraphics[width=0.30\linewidth]{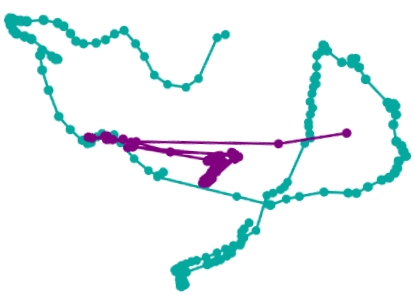} & \includegraphics[width=0.30\linewidth]{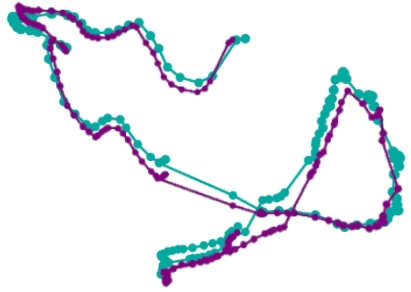} & \includegraphics[width=0.30\linewidth]{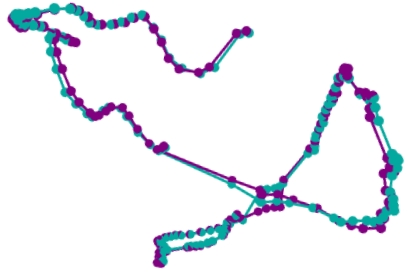} \\

   \includegraphics[width=0.25\linewidth]{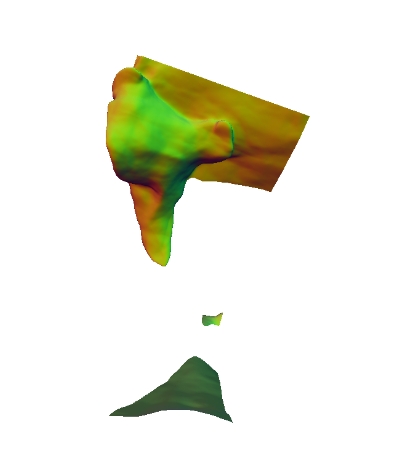} & \includegraphics[width=0.25\linewidth]{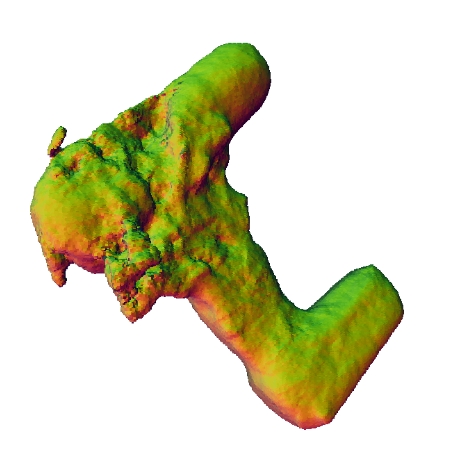} & \includegraphics[width=0.25\linewidth]{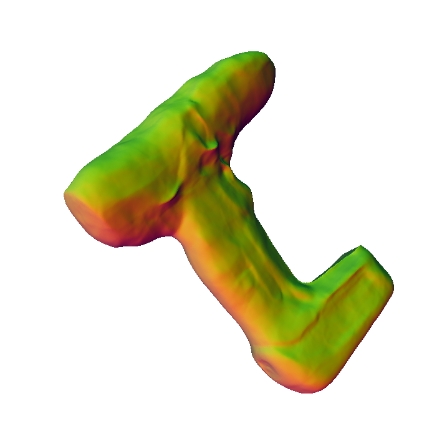} \\
    \small{(a) BARF} & \small{(b) Hampali~\etal.} & \small{(c) Ours}
    \end{tabular}
    \endgroup
  \caption{
  {\bf Different methods for joint pose and shape optimization.} {\bf (a)} BARF~\cite{barf} struggles in handling 360-degree sequences. {\bf (b)} The segment-wise optimization of Hampali~\etal.~\cite{meta-obj} is local optimal and suffers in this scenario with large pose changes. {\bf (c)} Our method produces globally-consistent results. We visualize the ground truth pose and the predicted pose in \textcolor[RGB]{3,168,158}{cyan} and \textcolor[RGB]{128,0,128}{purple}, respectively.
  }
  \label{fig:compare_barf_and_ours}
\end{figure}

\begin{figure}[tb]

  \centering
    \begingroup
    \setlength{\tabcolsep}{1pt}
    \begin{tabular}{cc}
\includegraphics[width=0.45\linewidth]{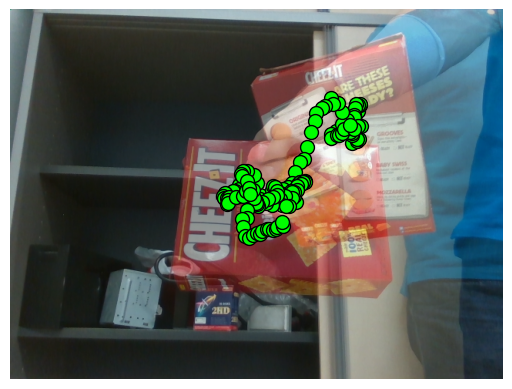} &
\includegraphics[width=0.45\linewidth]{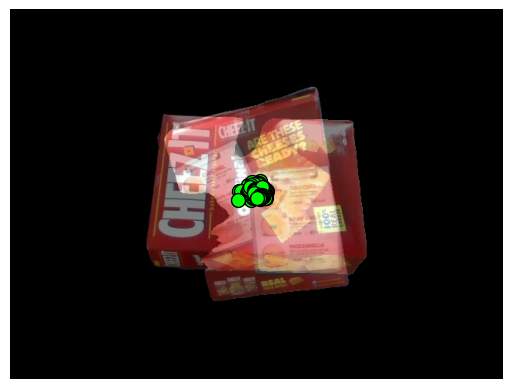} \\
\includegraphics[width=0.45\linewidth]{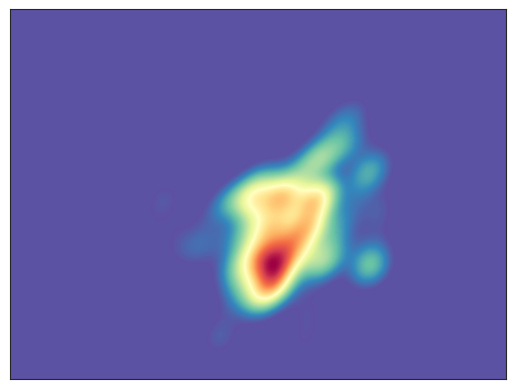} &
\includegraphics[width=0.45\linewidth]{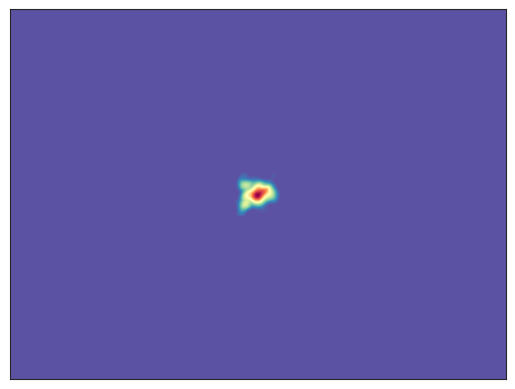} \\
    \small{Real camera} & \small{Virtual camera} \\
    \end{tabular}
    \endgroup
  \caption{\textbf{Effect of the virtual camera.}
The top row shows the trajectory of the object w.r.t. the real camera and the virtual camera, respectively. The bottom row shows the heatmap of 2D reprojections of the 3D object center across the whole HO3D dataset w.r.t different camera systems. The poses w.r.t the virtual camera do not have significant magnitude in both horizontal and vertical directions, which allows the poses to be approximately captured by only 4 degrees of freedom~(3 for rotation and 1 for distance).
  }
  \label{fig:vc_effect_short_demo}
\end{figure}

\subsection{Optimization with Guided Virtual Camera}
\label{sec:virtual_camera}

To optimize the object representation discussed in the above section, it is essential to specify both the origin and direction of camera rays or camera/object poses.
We do not assume the existence of such poses in our setting. Similar to~\cite{barf, meta-obj, nope-nerf, nope-neus}, we optimize the camera poses and the object representation simultaneously. Since camera rays are functions of camera parameters, we condition the camera rays in Eq.~\ref{eq:color_loss} on learnable camera poses, as illustrated as the Pose MLP in Fig.~\ref{fig:over_view}. We assume the camera intrinsics and lens distortions of the camera are known and only optimize the camera pose in this work.

Typically, joint optimization of camera pose and radiance field can only handle forward-facing scenarios and fails to converge if the images cover a larger range of viewpoints or there are some large pose changes between consecutive images, as shown in Fig.~\ref{fig:compare_barf_and_ours}.

To handle this problem, we propose to solve the optimization problem w.r.t. a new virtual camera system with the guidance of 2D object masks.
Specially, given a 3D point $\bx$ and the camera intrinsic matrix $\bK$, we have:
\begin{equation}
    \begin{aligned}
    \bu = \bK(\bR{\bf x}+{\bf t}),
    \end{aligned}
    \label{eq:perspective}
\end{equation}
where $\bu$ is the reprojected pixel location on the image, and $\bR$ and $\bt$ are the 3D rotation and 3D translation respectively. On the other hand, given the object mask obtained in Sec.~\ref{sec:method_setup}, we crop the object with a transformation matrix $\bM$, and with a virtual camera whose intrinsic matrix is $\bK_v$, we have:
\begin{equation}
    \begin{aligned}
    \bM\bu & = \bM\bK(\bR{\bf x}+{\bf t}) \\
    & = \bK_v\bK_v^{-1}\bM\bK(\bR{\bf x}+{\bf t}) \\
    & = \bK_v(\bK_v^{-1}\bM\bK\bR{\bf x}+\bK_v^{-1}\bM\bK{\bf t}), \\
    \end{aligned}
    \label{eq:virtual_camera_geo}
\end{equation}
where $(\bK_v^{-1}\bM\bK\bR, \bK_v^{-1}\bM\bK{\bf t}) \rightarrow (\bR_v, \bt_v)$ are the 3D rotation and 3D translation w.r.t. the new virtual camera. While note that $(\bR_v, \bt_v)$ is not physically-compliant and can only be approximately estimated. Instead of optimizing $(\bR, \bt)$ directly, we optimize $(\bR_v, \bt_v)$ and the object representation simultaneously w.r.t. the virtual camera.
Since the translation of the target w.r.t. the new virtual camera does not have significant magnitude in both horizontal and vertical directions, as illustrated in Fig.~\ref{fig:vc_effect_short_demo}, we only predict the rotation and the distance from the camera center to the object center for object poses, which has only 4 degrees of freedom. This simplification of pose formulation reduces the search space of network optimization, and our experiments will show it increases the performance significantly.

\subsection{Segment-Free Progressive Training}
\label{sec:method_progressive}

Although the virtual camera system reduces the search space of optimization, it still can not handle 360-degree sequences. We use progressive training to process the images in the sequence to leverage the temporal information between consecutive images~\cite{meta-obj, 3dgs}.


\begin{figure}[tb]
  
    \begingroup
    \setlength{\tabcolsep}{0.3pt}
    \begin{tabular}{ccc}
    \includegraphics[width=0.33\linewidth]
    {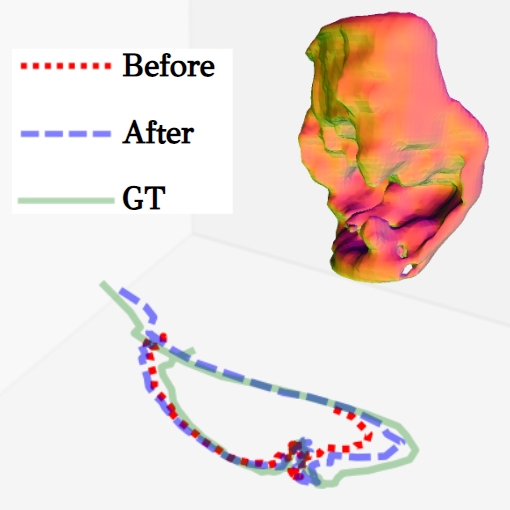} &
    \includegraphics[width=0.33\linewidth]{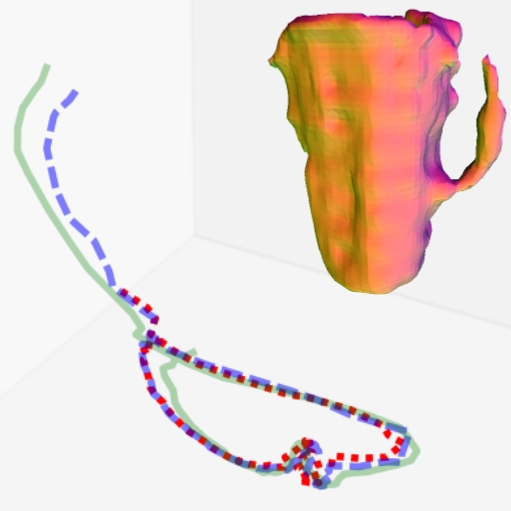} &
    \includegraphics[width=0.33\linewidth]{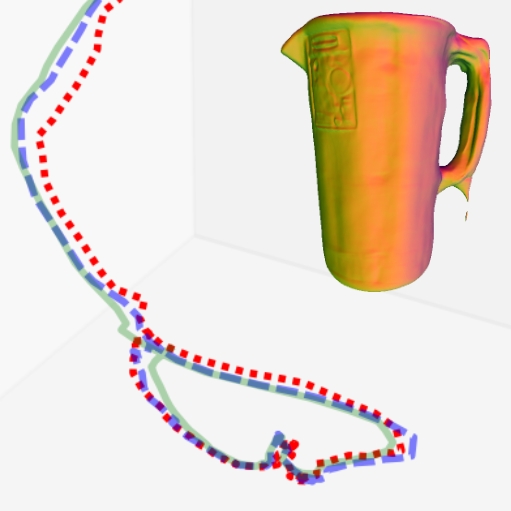}\\
    \small{$t_{1}\rightarrow t_{1}+$10} & \small{$t_{2}\rightarrow t_{2}+$10} & \small{with refinement} \\
    \end{tabular}
    \endgroup
  \caption{
  \textbf{Progressive training and global refinement.}
  The first two figures show the pose and shape results of two examples during progressive training. The result improves with more images involved. The last figure shows the result with global refinement, which improves the performance further.
  }
  \label{fig:progressive_pose}
\end{figure}

We use 2D matches between different images to facilitate the optimization:
\begin{equation}
\begin{aligned}
\mathcal{L}_{match}(\br_a, \br_b) = \sum_{k=1}^N (w_{ak}\cdot\|g_b(\bx_{ak}) - \bu_b\|_1) \\ + \sum_{k=1}^N (w_{bk}\cdot\|g_a(\bx_{bk}) - \bu_a\|_1),
\end{aligned}
\label{eq:match_loss}
\end{equation}
where $\br_a$ and $\br_b$ are two camera rays passing through the matched pixel coordinates $\bu_a$ and $\bu_b$ between two different images $a$ and $b$, $\{\bx_{ak}\}$ and $\{\bx_{bk}\}$ are the sampled points along the ray $\br_a$ and $\br_b$, $g_a(\cdot)$ and $g_b(\cdot)$ are the 2D reprojection functions according to the current predicted poses of images $a$ and $b$, $w_{ak}$ and $w_{bk}$ are the predicted weights for the points on the ray.
We use LoFTR~\cite{loftr} to generate sparse 2D matches for images within a few frame intervals~(typically less than 10). We randomly sample available 2D matches between different frames during training.


Another challenge of pose-free reconstruction is the large pose changes where we observe that the newly appeared surface degrades the previously reconstructed shape. To help the networks learn the pose of newly added images, we reset the weights of the shape network when the relative pose of the current frame exceeds a rotation threshold $\tau$ from the last reset, and only keep the weights of the Pose MLP to preserve the learned pose information of previous images. We typically set $\tau$ as 60$\degree$ in our experiments.



\subsection{Refinement with Real Camera}
\label{sec:method_global}

The optimization target $\{(\bR_v, \bt_v)\}$ of previous sections is based on the virtual camera and not physically-compliant.
It is usually not accurately aligned with the original optimization target $\{(\bR, \bt)\}$.
To address this problem, we use a PnP solver to transform the predicted pose from the virtual camera system back to the real camera system and refine the results w.r.t. the real camera.

Specifically, we sample a set of 3D points from the reconstructed mesh in previous sections, and re-project them to the virtual image plane with the predicted $(\hat{\bR}_v, \hat{\bt}_v)$.
After transforming the reprojected 2D position with $\bM^{-1}$ in Eq.~\ref{eq:virtual_camera_geo}, we establish a set of 3D-to-2D correspondences between 3D points and 2D image locations on the raw image, we use RANSAC EPnP~\cite{epnp} to compute $(\hat{\bR}, \hat{\bt})$ which is the estimated object pose w.r.t. the real camera. We then conduct a global optimization with all available images starting from $\{(\hat{\bR}, \hat{\bt})\}$. We do not use the match loss in this procedure, and Fig.~\ref{fig:progressive_pose} shows some results.

\subsection{Implementation Details}
\label{sec:impmentation_details}
\noindent {\bf Networks.} Besides the standard SDF MLP and Color MLP used in standard NeuS~\cite{neus}, we use another small Pose MLP to estimate the object pose of the target in every frame.
Similar to~\cite{NeRFtrinsic_Four}, we map each frame ID to a 256-dim feature vector based on Gaussian Fourier features and then use 3 layers of MLP with GELU activation functions to output the poses.
We use a single ADAM optimizer~\cite{adam} for SDP MLP and Color MLP. During training, the learning rate warms up linearly from 0 to 5e-4 during the first 5k iteration and then follows a cosine decay schedule with alpha=0.05. For Pose MLP, we use another ADAM optimizer with a cosine decay schedule of alpha=0.5.

\noindent {\bf Sampling.} We use an unit sphere for the initialization of the SDF network, and define the sampling range~(i.e., near and far) within the unit sphere.
For each training step, we randomly sample 512 rays from the input image batch. During the optimization with guided virtual camera, we only sample 32 points along each ray for efficiency.
We progressively train our model with $B$ consecutive images as a group. For every group, we train the networks with a fixed number of training steps (typically 1K). We sample 20\% of the rays from images within previously-converged groups and 80\% from the images within the newly added group.

In the phase of refinement with the real camera, we use importance-based hierarchy sampling strategies~\cite{neus} to uniformly sample 64 points and then sample another 64 points based on the current predicted SDF values.
We train the networks for 150K training steps for refinement. On a typical NVIDIA V100 GPU, the training of a 100-frame sequence takes about 10 hours including both the initialization and refinement.

%% file: sec/04_experiment.tex
\section{Experiments}

\begin{table*}
  \centering
  \begin{tabular}{lcccccccccc}
    \toprule
    Methods & cracker & sugar & mustard & bleach & meatcan & driller & pitcher & mug & banana & Average \\
    \midrule
    COLMAP~\cite{colmap}  & 4.08 & 6.66 & 4.43 & 14.11 & 10.21 & 11.06 & 43.38 & - & - & 13.41\\
    Ye~\etal.~\cite{ihoi}& 10.21 & 6.19 & \textbf{2.61} & \textbf{4.18}  & 3.43 & 15.15 & 8.87 & - & \textbf{3.47} & 6.76\\
    Hampali~\etal.~\cite{meta-obj} & 2.91 & 3.01 & 4.44 & 5.63 & 1.95 & 5.48 & 9.21 & 4.53 & 4.60 & 4.64\\
    \textbf{Ours} & \textbf{1.71}& \textbf{1.84} & 3.49 & 5.38 & \textbf{1.80} & \textbf{3.82} & \textbf{2.84} & \textbf{2.78} & 4.54 & \textbf{3.14}\\
    \midrule
     UNISURF~\cite{unisurf} & 3.40 & 3.49 & 4.34 & 3.41 & 1.54 & 5.33 & 4.63 & - & 3.98 & 3.76 \\
    NeuS~\cite{neus} & \textbf{1.75} & \textbf{1.69} & 2.34 & 3.35 & \textbf{1.17} & \textbf{3.13} & 3.48 & \textbf{2.19} &\textbf{2.08} &  \textbf{2.35}\\
    Patten~\etal.~\cite{rgb-d-baseline} & 3.54 & 3.34 & 3.28 & \textbf{2.43} & 3.26 &3.77 & 4.73 & 4.22 & 2.44 & 3.45\\
  \bottomrule
  \end{tabular}
    \caption{\textbf{Mesh evaluation in $HD_{RMSE}\downarrow$.}
    Our method outperforms most pose-free methods~(top group), and is on par with methods trained based on ground truth poses~(bottom group).
}
  \label{tab:recon_comp_ho3d}
\end{table*}

\begin{table*}
  \centering
  \begin{tabular}{lcccccccccc}
    \toprule
    Methods & cracker & sugar & mustard & bleach & meatcan & driller & pitcher & mug & banana & Average \\
    \midrule
    COLMAP~\cite{colmap}  & 7.4 & 7.4 & 3.5 &  1.5 & 0.1 & 2.8 & 4.1 & 2.4 & 0.0 & 2.9\\
    Hampali~\etal.~\cite{meta-obj} & \textbf{7.6} & 6.8 & \textbf{5.2}  & 4.7 & \textbf{6.8} & 6.4 & 4.6 & 2.2 & \textbf{0.6} & 4.5\\
    \textbf{Ours} & \textbf{7.6} & \textbf{7.6} & 4.2 & \textbf{5.3} & 3.1  & \textbf{8.5} & \textbf{8.7} & \textbf{8.1} & 0.3 & \textbf{5.9}\\
  \bottomrule
  \end{tabular}
    \caption{\textbf{Pose evaluation in $AUC_{ATE}\uparrow$.} Our method produces more accurate pose result on most objects.
    }
  \label{tab:pose_comp_ho3d}
\end{table*}

We evaluate our method systematically in this section. We first introduce the datasets and metrics used in our experiments and then demonstrate the effectiveness of our method by comparing with state-of-the-art methods. We then conduct extensive ablation study to validate different components of our method.

\noindent {\bf Datasets.} We first evaluate our method on the standard HO3D dataset ~\cite{ho3d}, which includes video captures of daily objects with a fixed camera. Since most objects in HO3D are manipulated by one hand, and there exists a fixed relative pose between the hand and the object in most sequences, we collect a more general dataset of free-moving objects with a head-mounted device with egocentric views, where the objects are manipulated by both hands with a free manipulation style.


\noindent {\bf Metrics.} For object reconstruction, we first align the predicted mesh to the GT mesh based on scale-aware ICP~\cite{meta-obj}, then calculate the root mean square of the Hausdorff distance~(mm)~\cite{meta-obj, rgb-d-baseline} between the predicted mesh and the GT mesh ($HD_{RMSE}$). For object pose estimation, we first align predicted poses to GT poses based on similarity transforms~\cite{meta-obj, nope-nerf, 3dgs}, and then compute the area under curve~(AUC) with a threshold of 10cm in Absolute Trajectory Error (ATE)~($AUC_{ATE}$)~\cite{meta-obj}.
We also report the pose accuracy in Relative Pose Error~(RPE)~\cite{nope-nerf, 3dgs}, including $RPE_r$~(degree) and $RPE_t$~(cm), corresponding to the rotation error and the translation error, respectively.


\subsection{Evaluation on HO3D}
We report results on the 9 sequences of HO3D as in~\cite{meta-obj, ihoi}, and compare our method with pose-free methods including COLMAP~\cite{colmap}, Ye~\etal.~\cite{ihoi}, and Hampali~\etal.~\cite{meta-obj}, where Ye~\etal. relies on prior hand pose information.
To further validate our method, we compare our reconstructed mesh with the results of UNISURF~\cite{unisurf} and NeuS~\cite{neus} which are trained with ground truth poses. We also compare with another method Patten~\etal.~\cite{rgb-d-baseline} that rely on additional depth images.

\begin{figure}[!ht]
  \centering
    \begingroup
  \setlength{\tabcolsep}{1pt}
    \begin{tabular}{cccc}
    \includegraphics[width=0.24\linewidth]{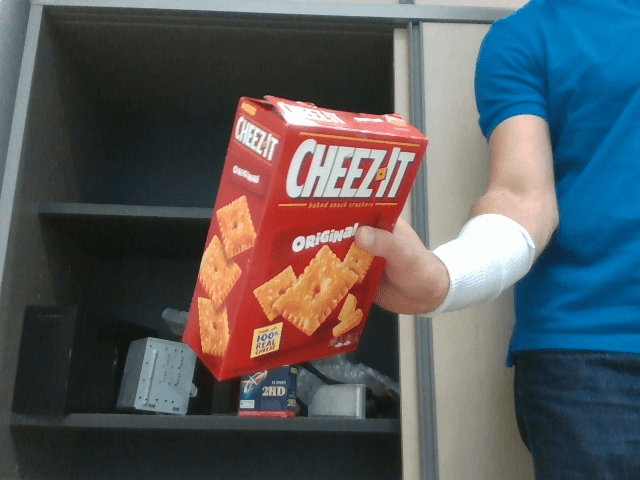} & 
    \includegraphics[width=0.24\linewidth,trim={50, 0, 50, 0}]{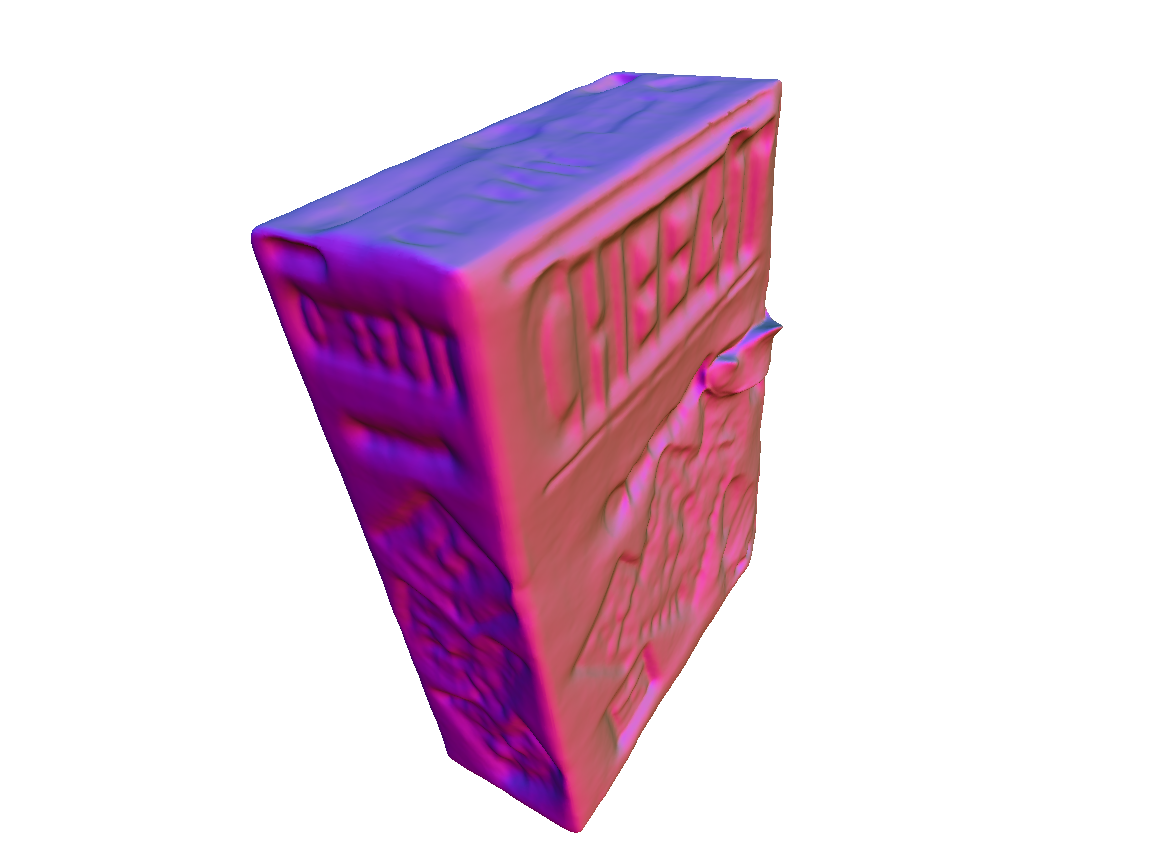} &
    \includegraphics[width=0.24\linewidth,trim={50, 0, 50, 0}]{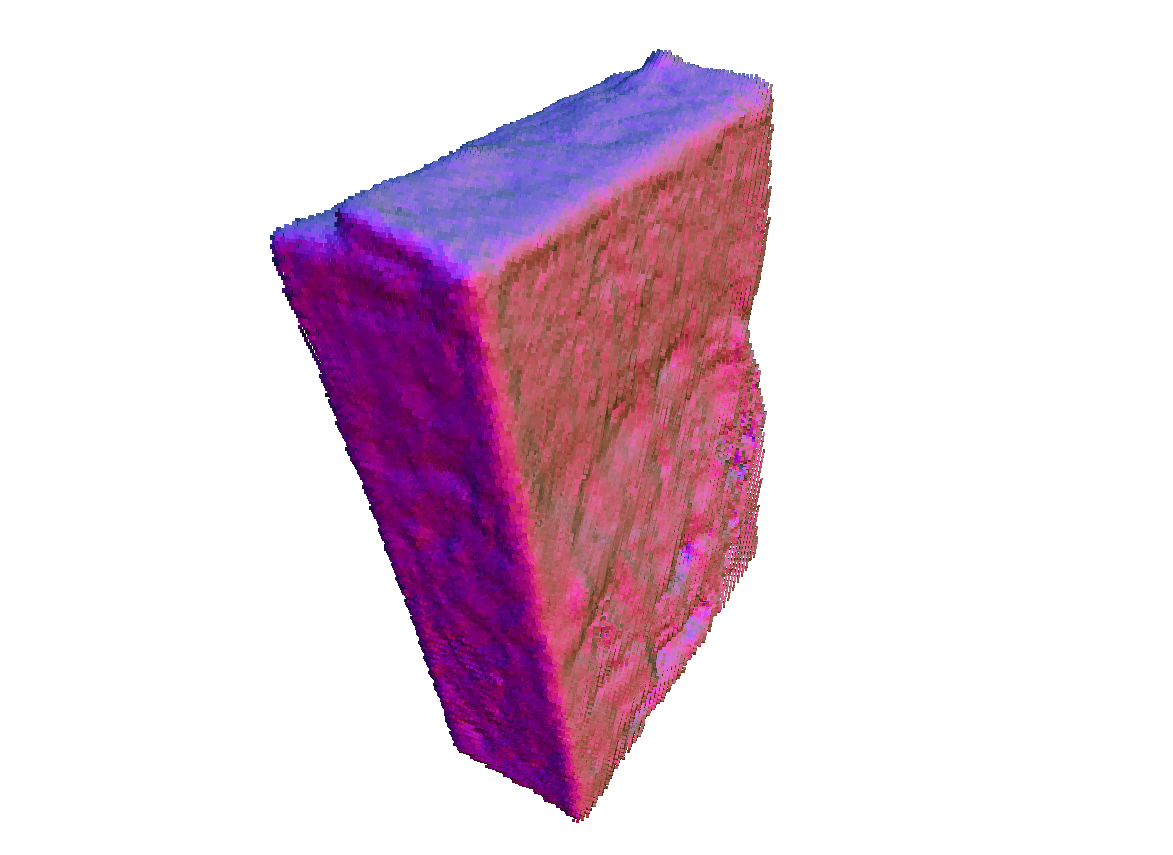} &
    \includegraphics[width=0.24\linewidth,trim={50, 0, 50, 0}]{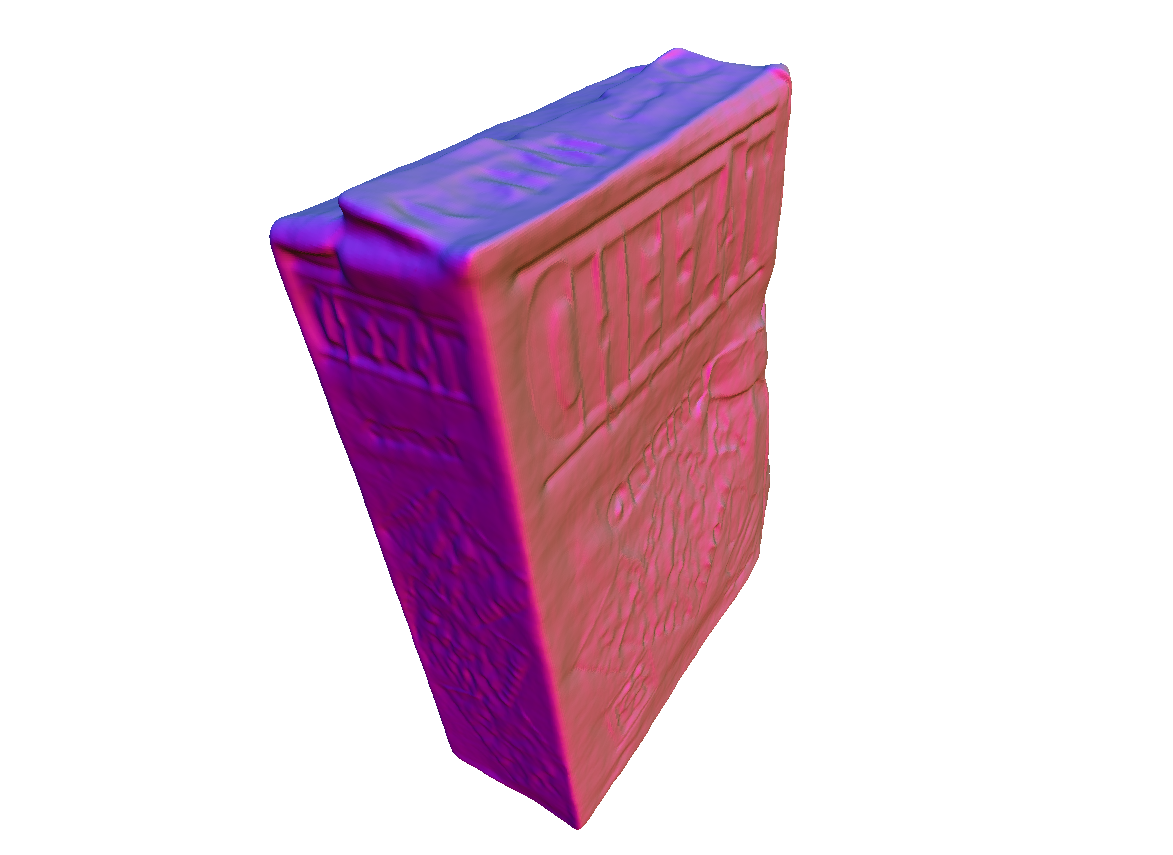} 
    \\
    \includegraphics[width=0.24\linewidth]{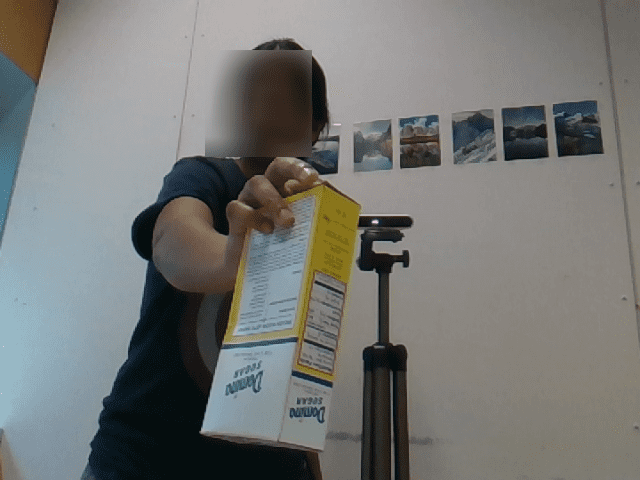} & 
    \includegraphics[width=0.24\linewidth,trim={50, 0, 50, 0},clip]{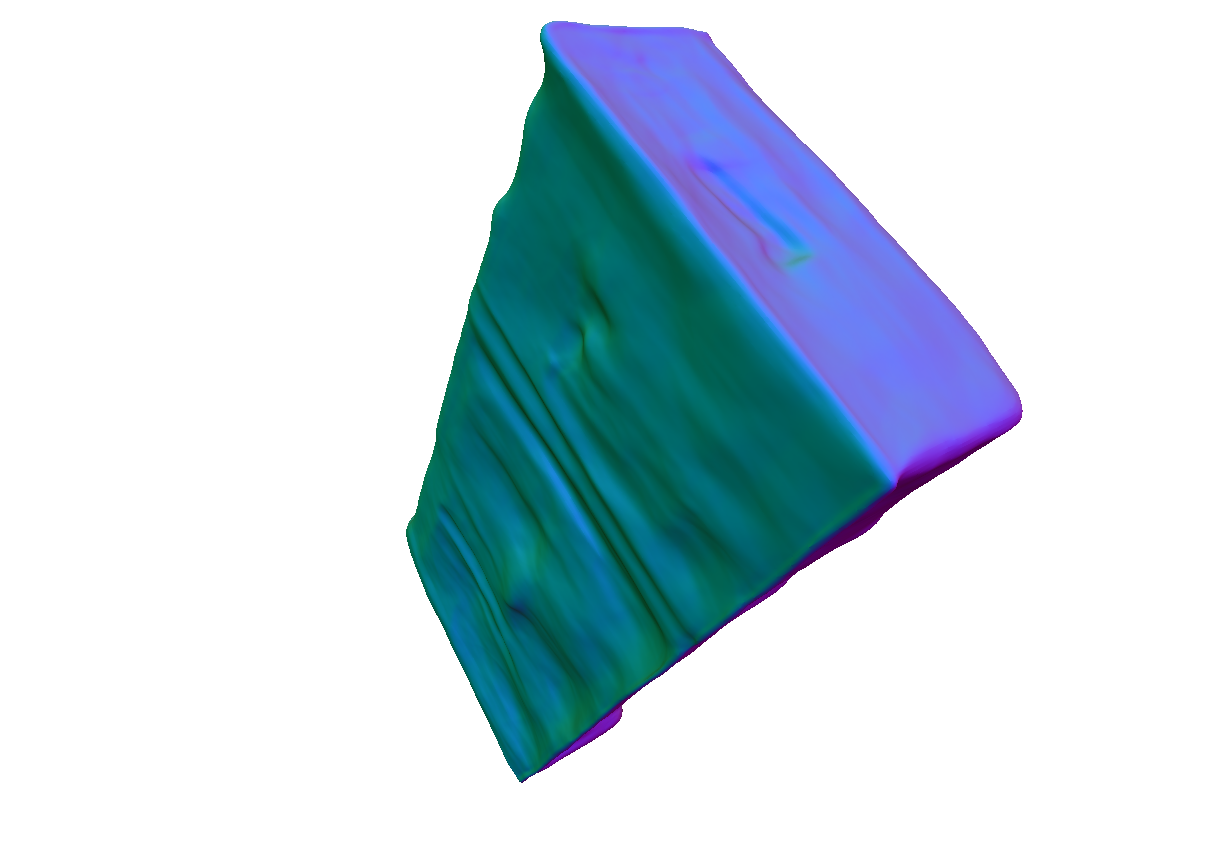} &
    \includegraphics[width=0.24\linewidth,trim={50, 0, 50, 0},clip]{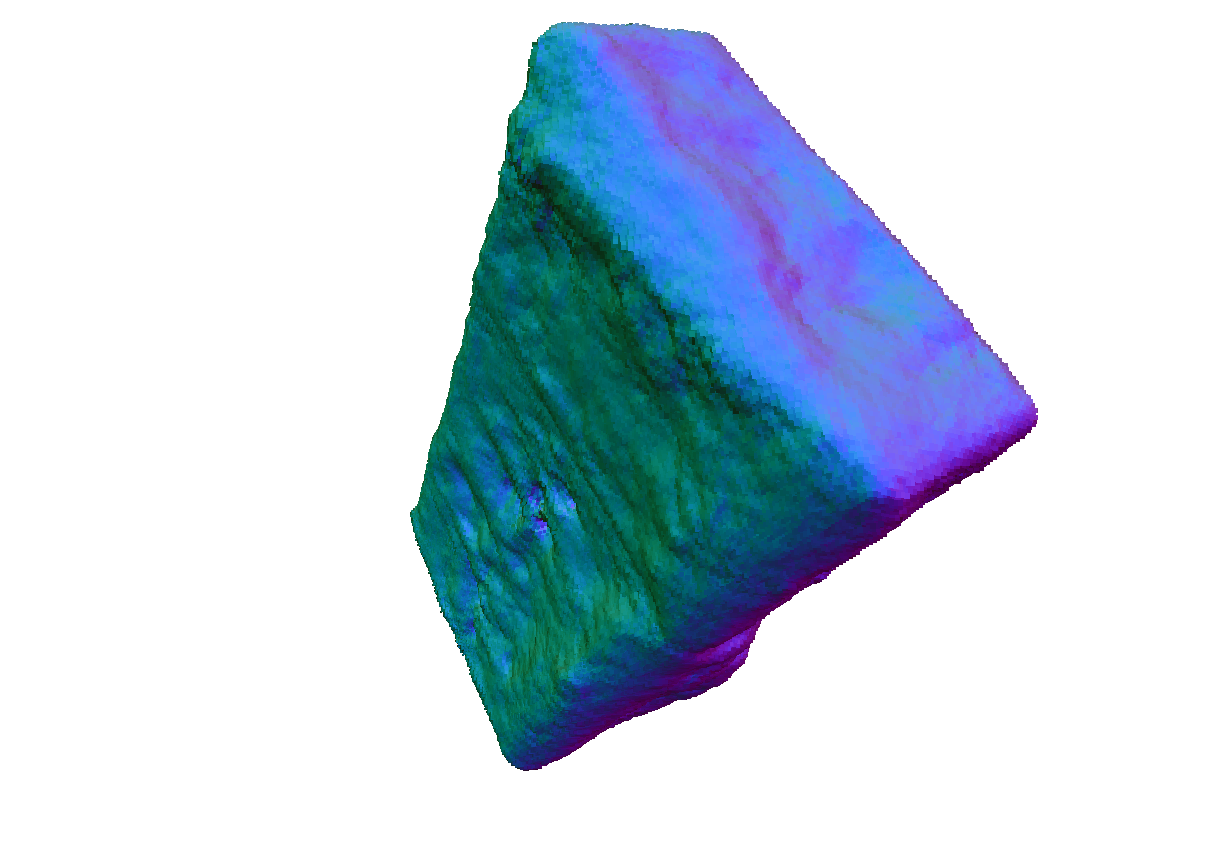} &
    \includegraphics[width=0.24\linewidth,trim={50, 0, 50, 0},clip]{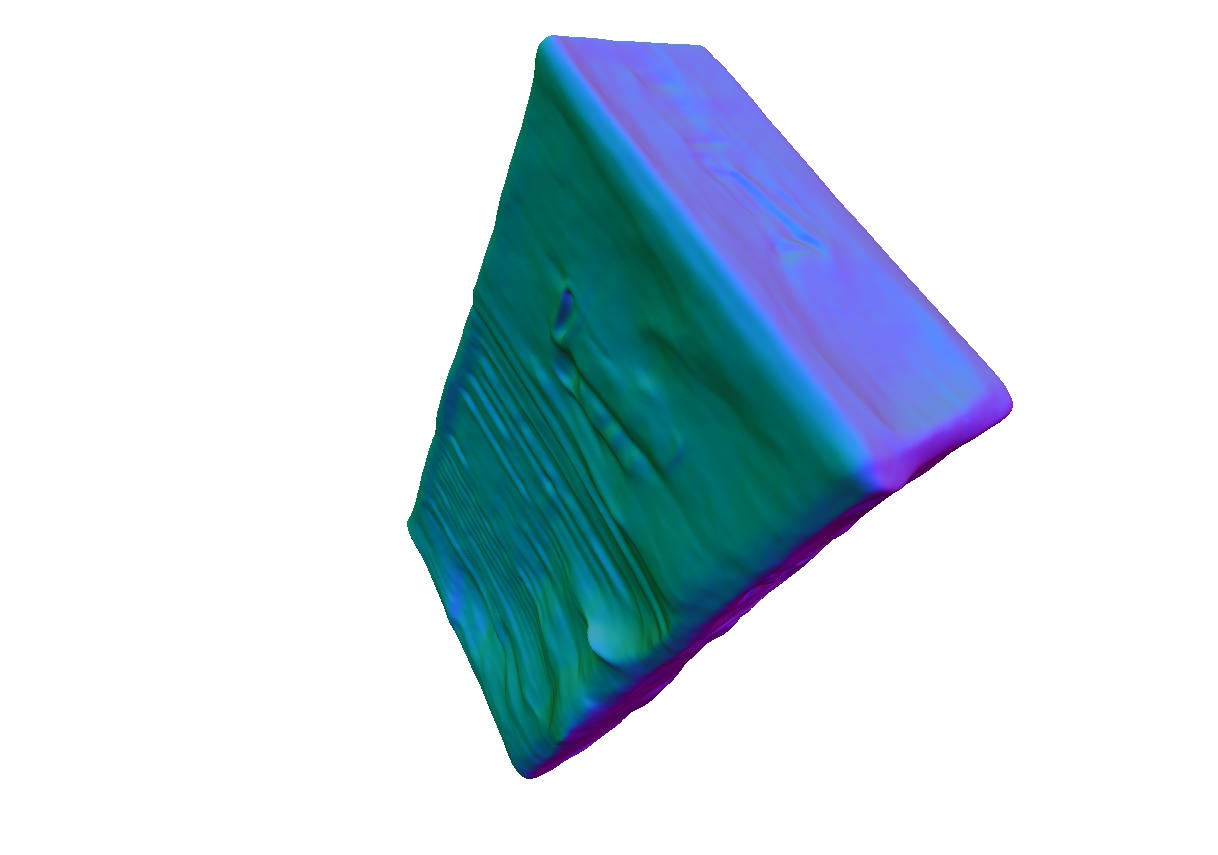} 
    \\
    \includegraphics[width=0.24\linewidth]{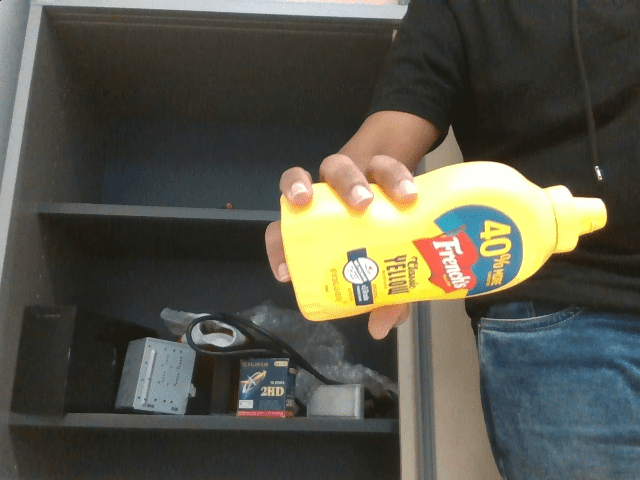} & 
    \includegraphics[width=0.24\linewidth,trim={60, 0, 60, 0},clip]{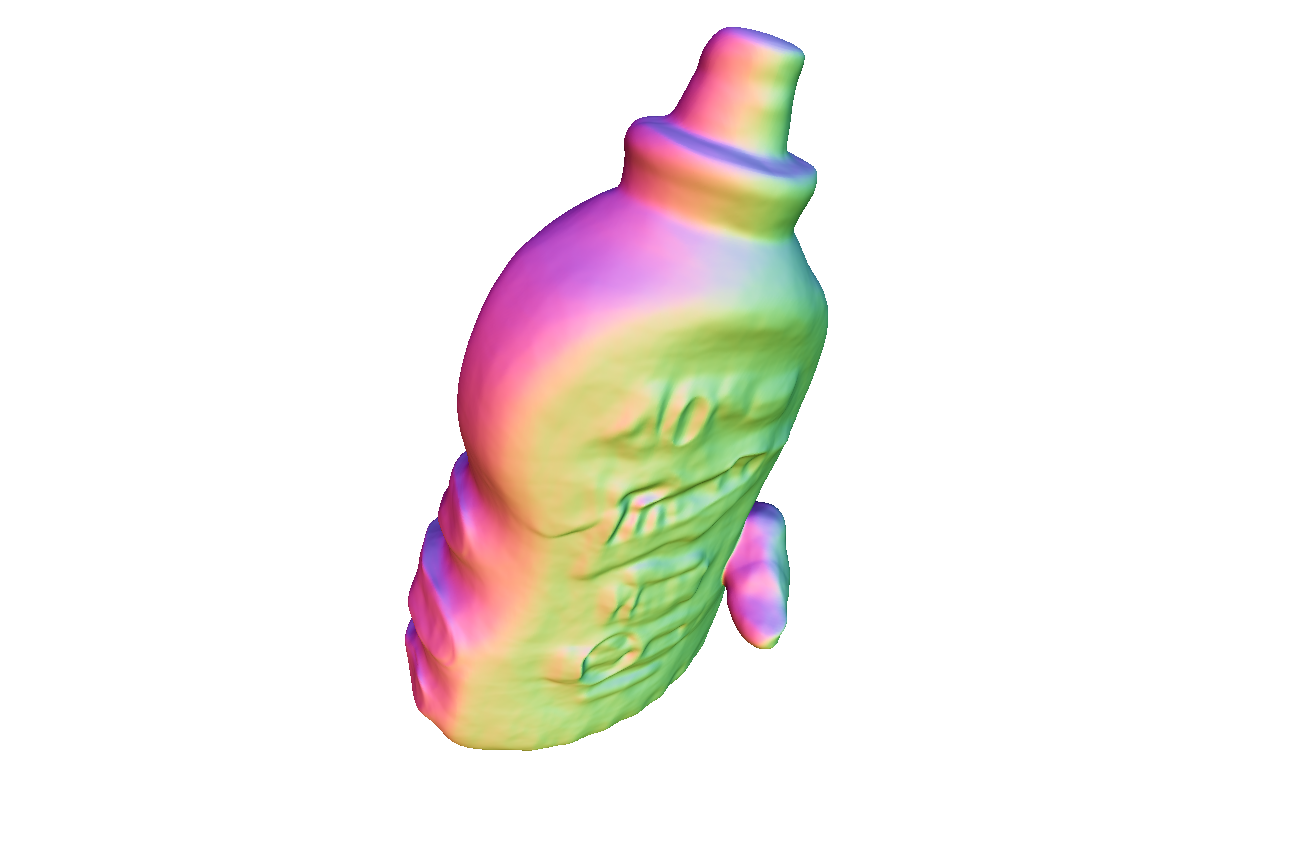} &
    \includegraphics[width=0.24\linewidth,trim={60, 0, 60, 0},clip]{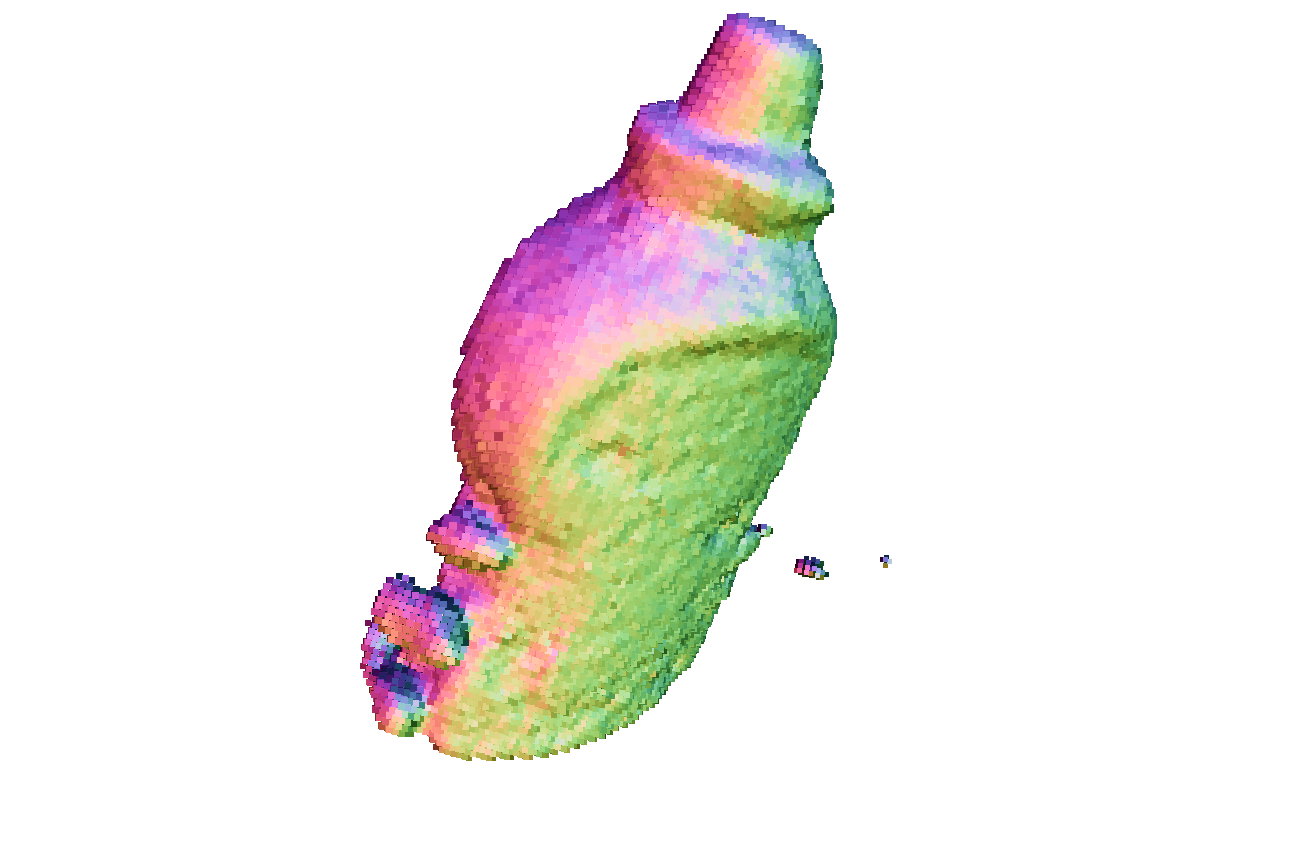} &
    \includegraphics[width=0.24\linewidth,trim={60, 0, 60, 0},clip]{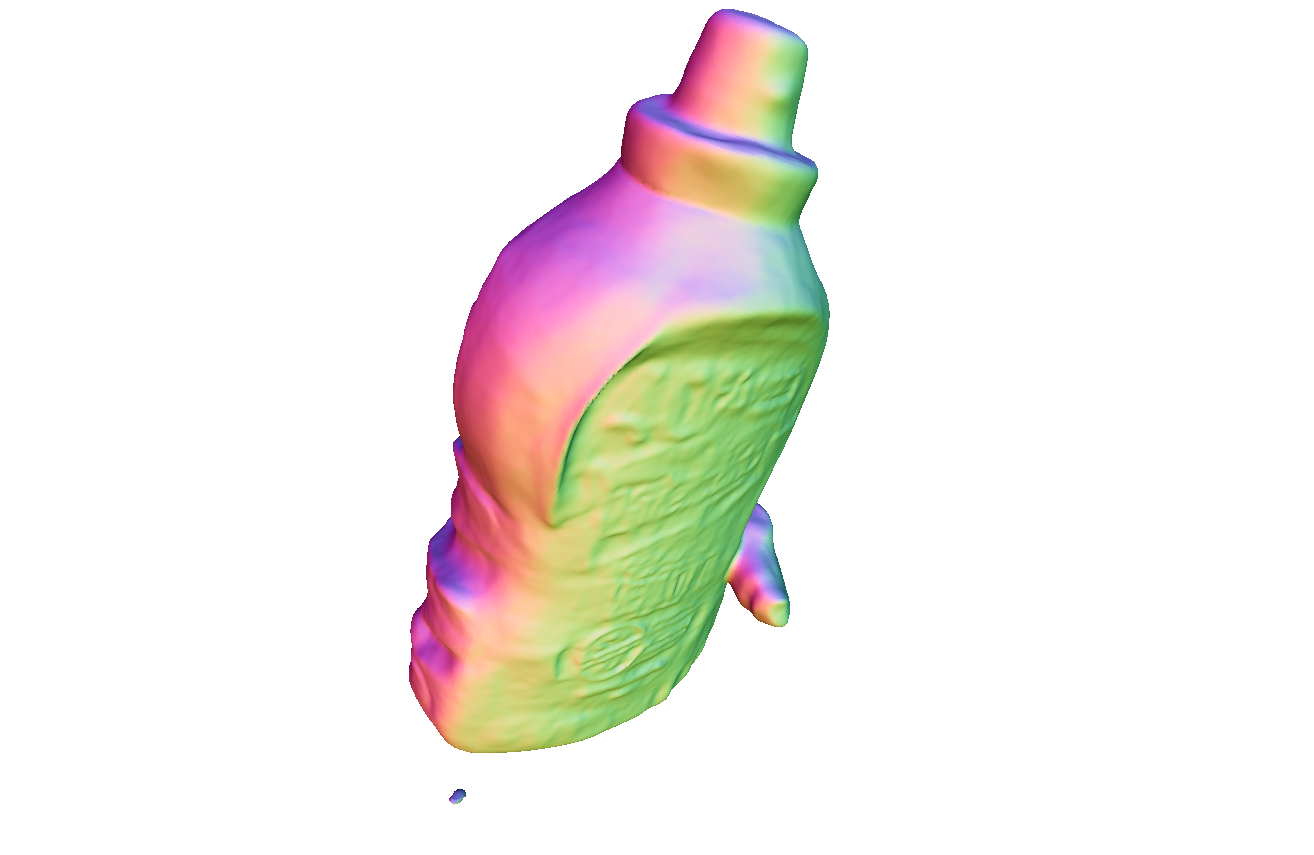}  
    \\
    \includegraphics[width=0.24\linewidth]{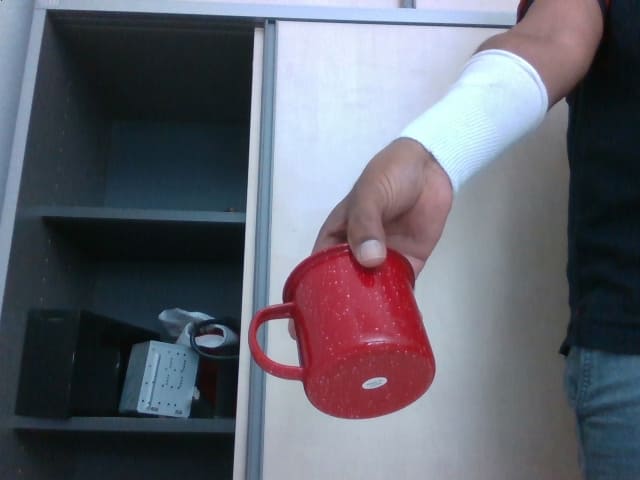} & 
    \includegraphics[width=0.24\linewidth,trim={30, 0, 30, 0}]{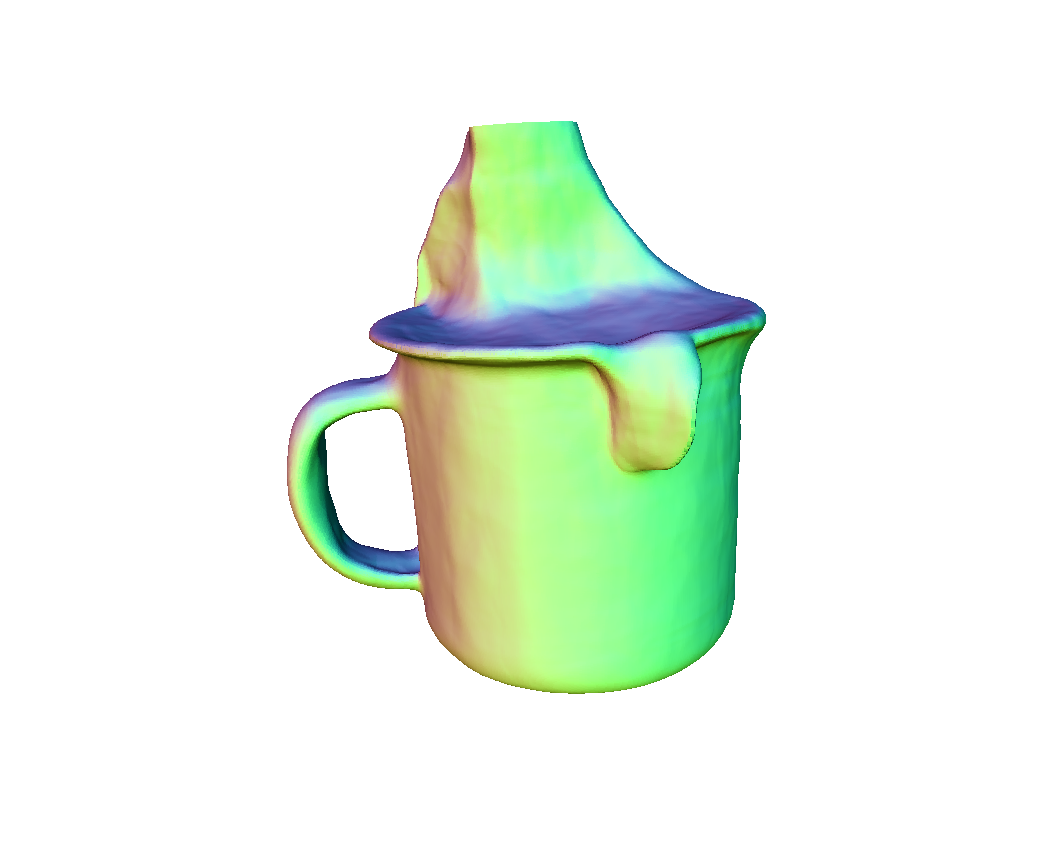} &
        \includegraphics[width=0.24\linewidth,trim={30, 0, 30, 0}]{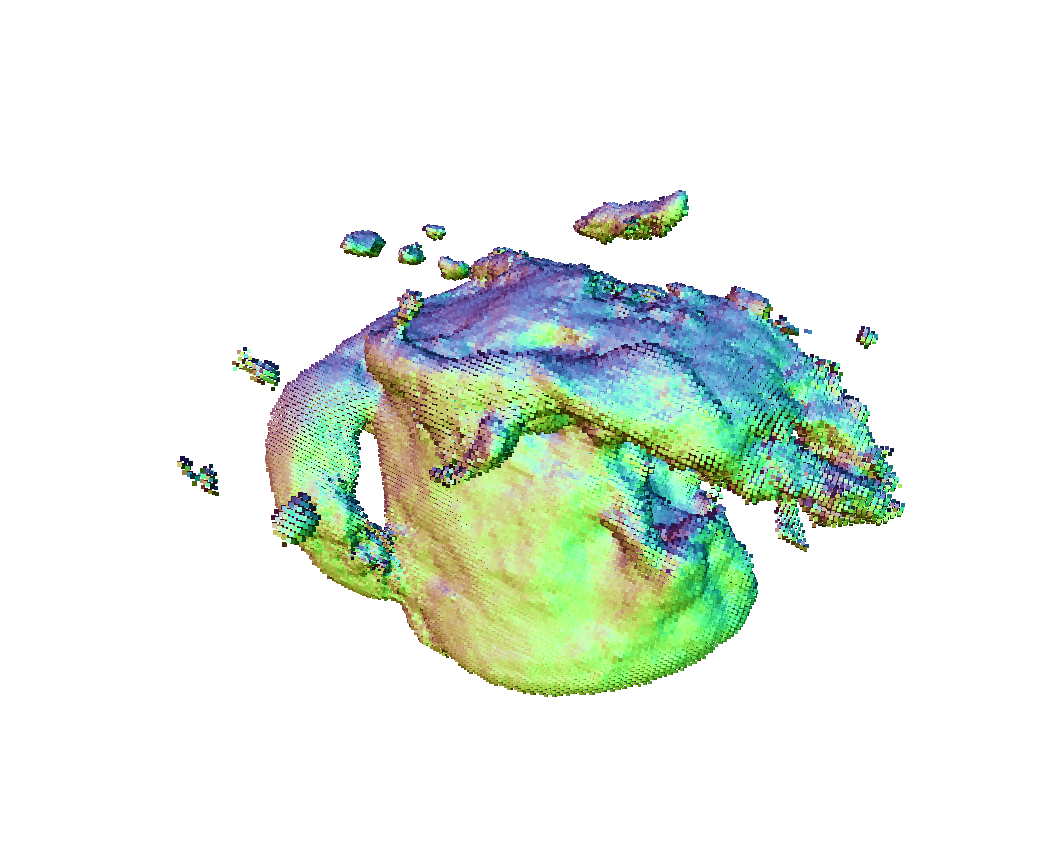} &
    \includegraphics[width=0.24\linewidth,trim={30, 0, 30, 0}]{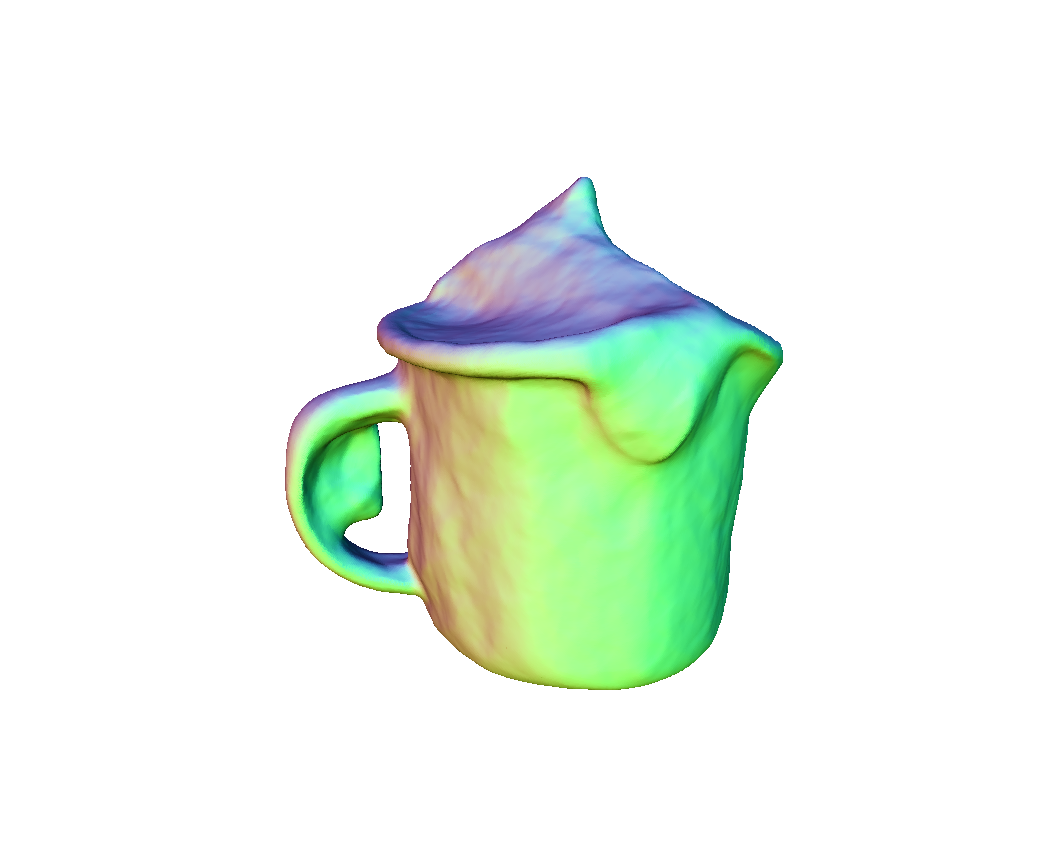}
    \\
    \includegraphics[width=0.24\linewidth]{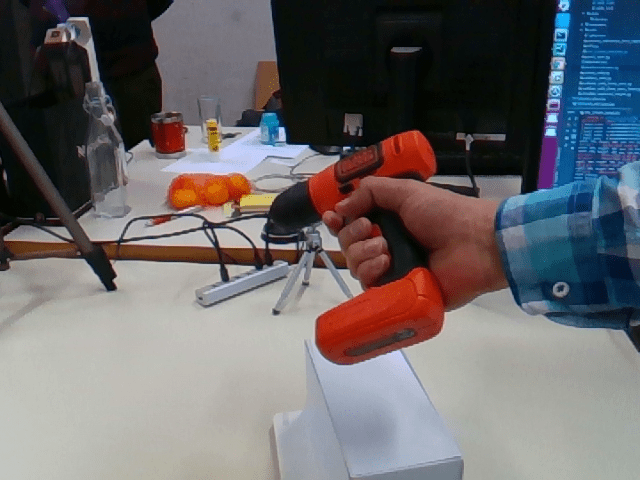} & 
    \includegraphics[width=0.24\linewidth,trim={60, 0, 60, 0},clip]{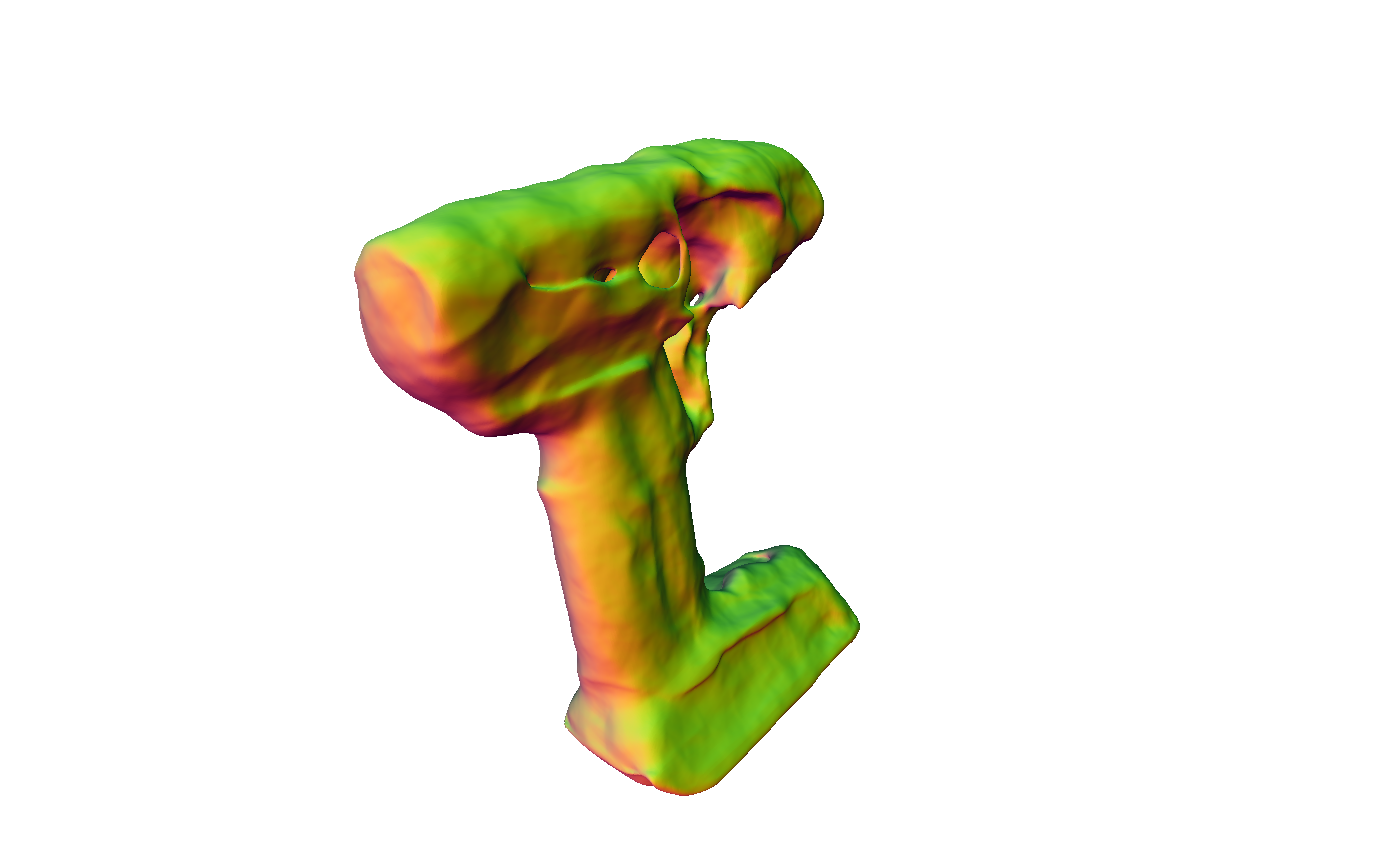} &
    \includegraphics[width=0.24\linewidth,trim={60, 0, 60, 0},clip]{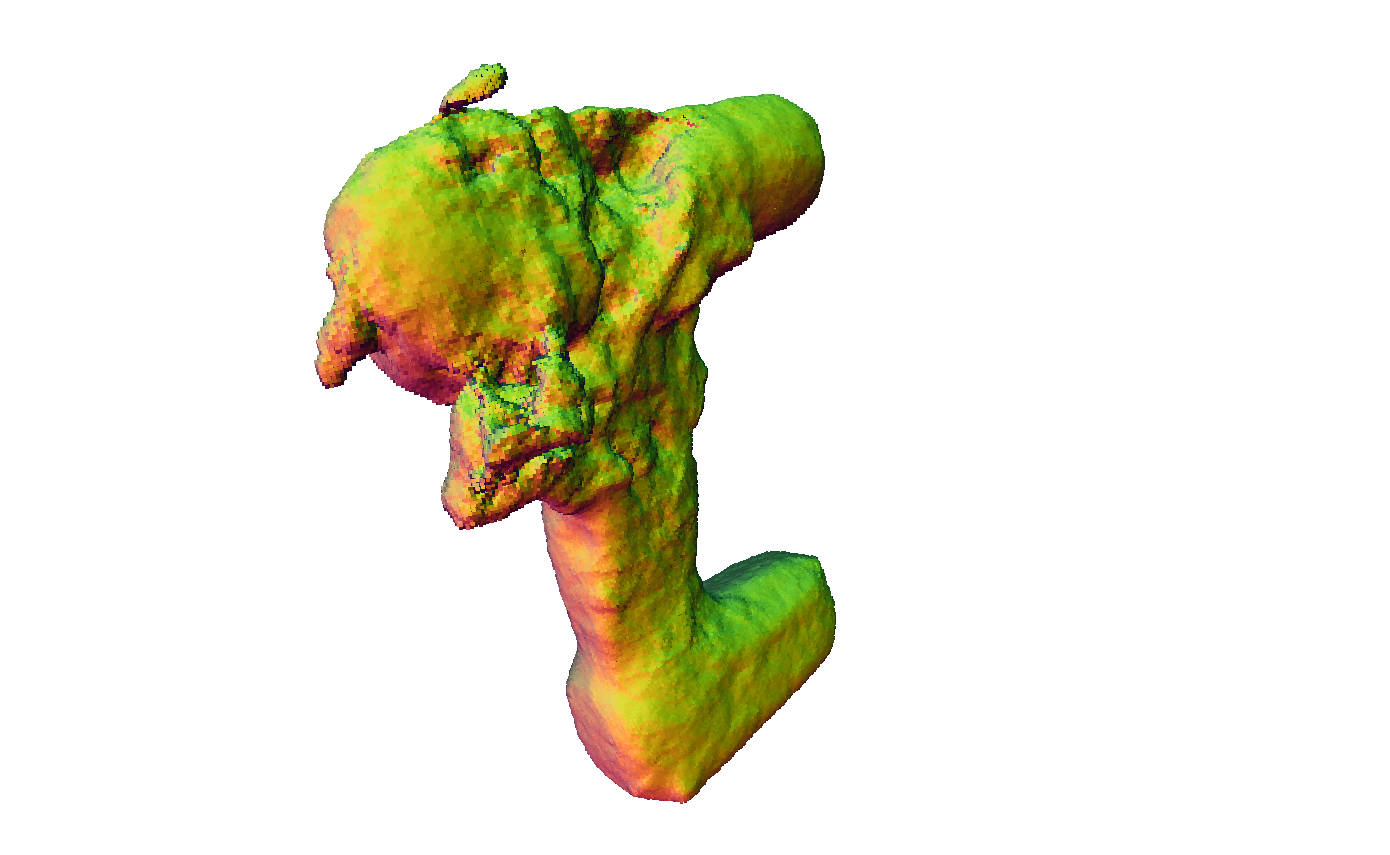} &
    \includegraphics[width=0.24\linewidth,trim={100, 30, 100, 0},clip]{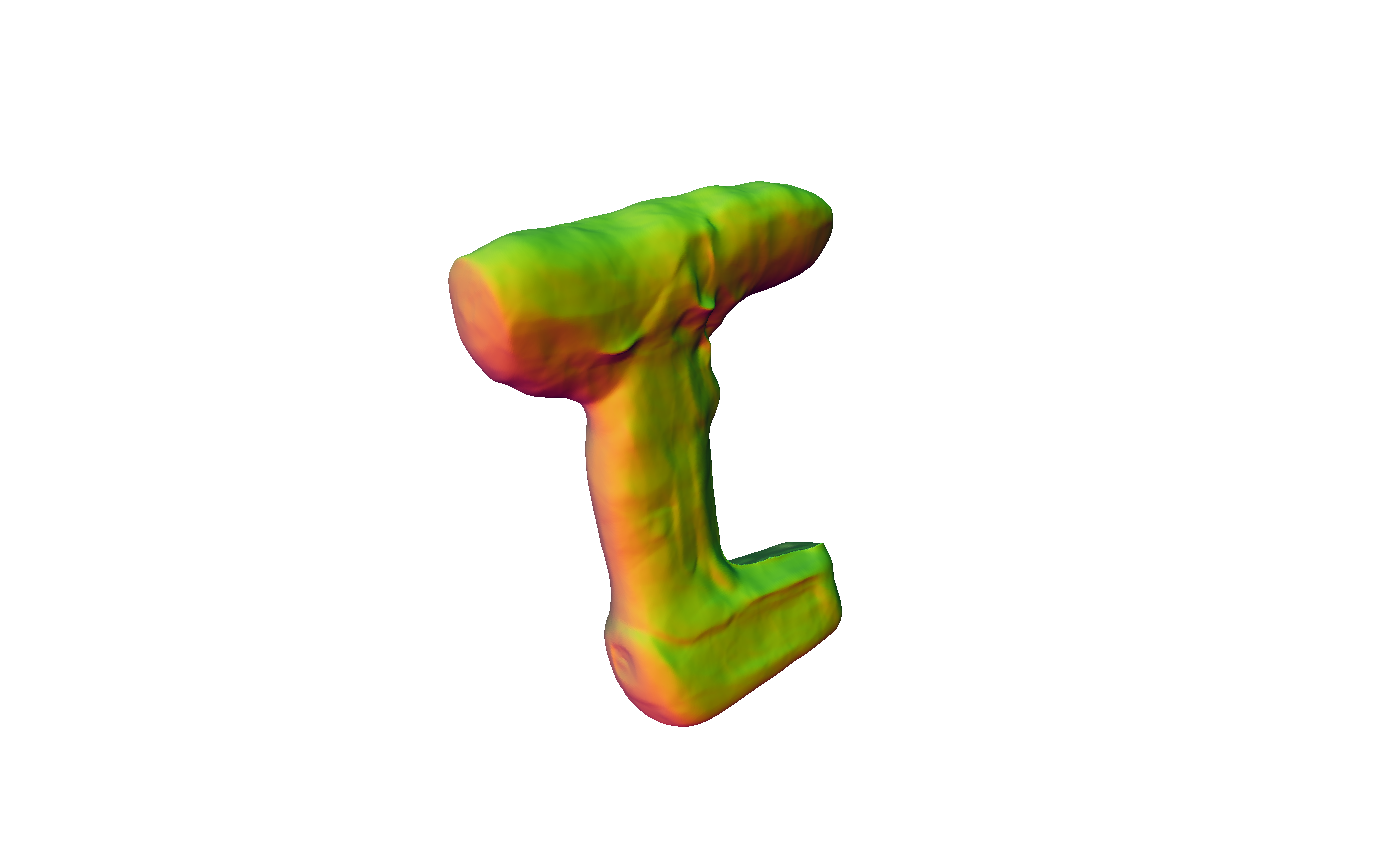}
    \\
    \includegraphics[width=0.24\linewidth]{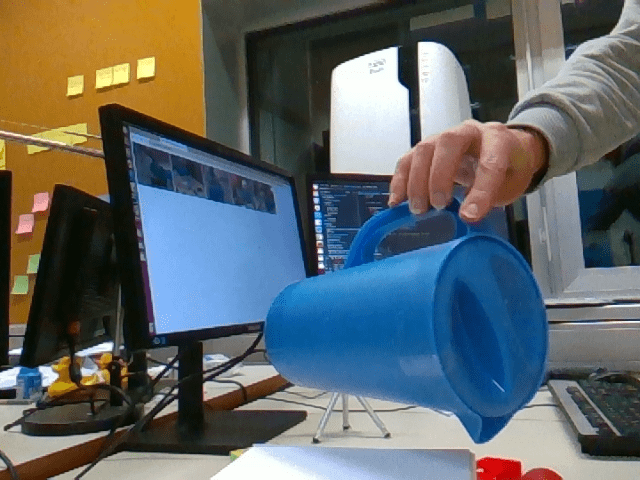} & 
    \includegraphics[width=0.24\linewidth,trim={60, 0, 60, 0},clip]{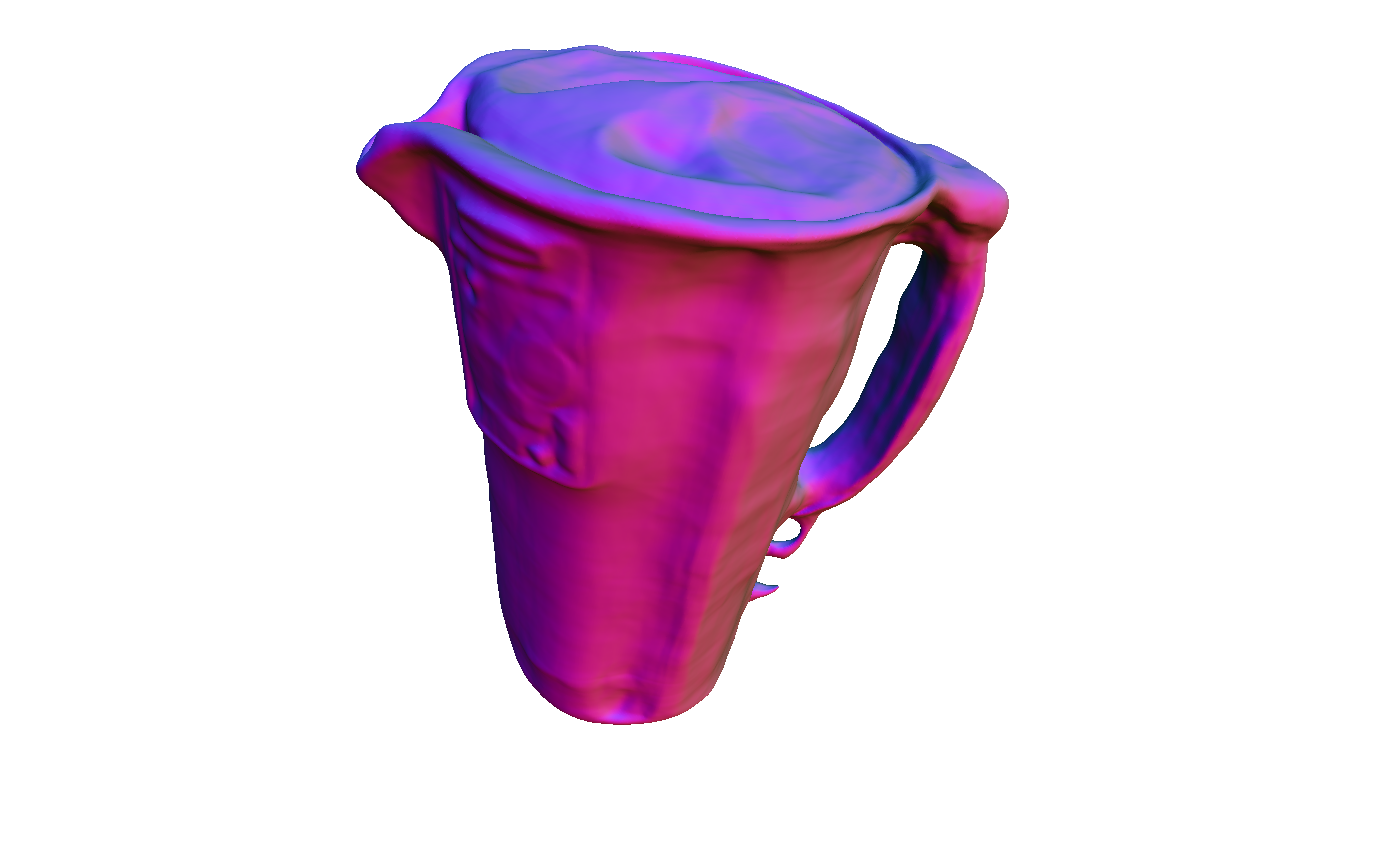} &
    \includegraphics[width=0.24\linewidth,trim={60, 0, 60, 0},clip]{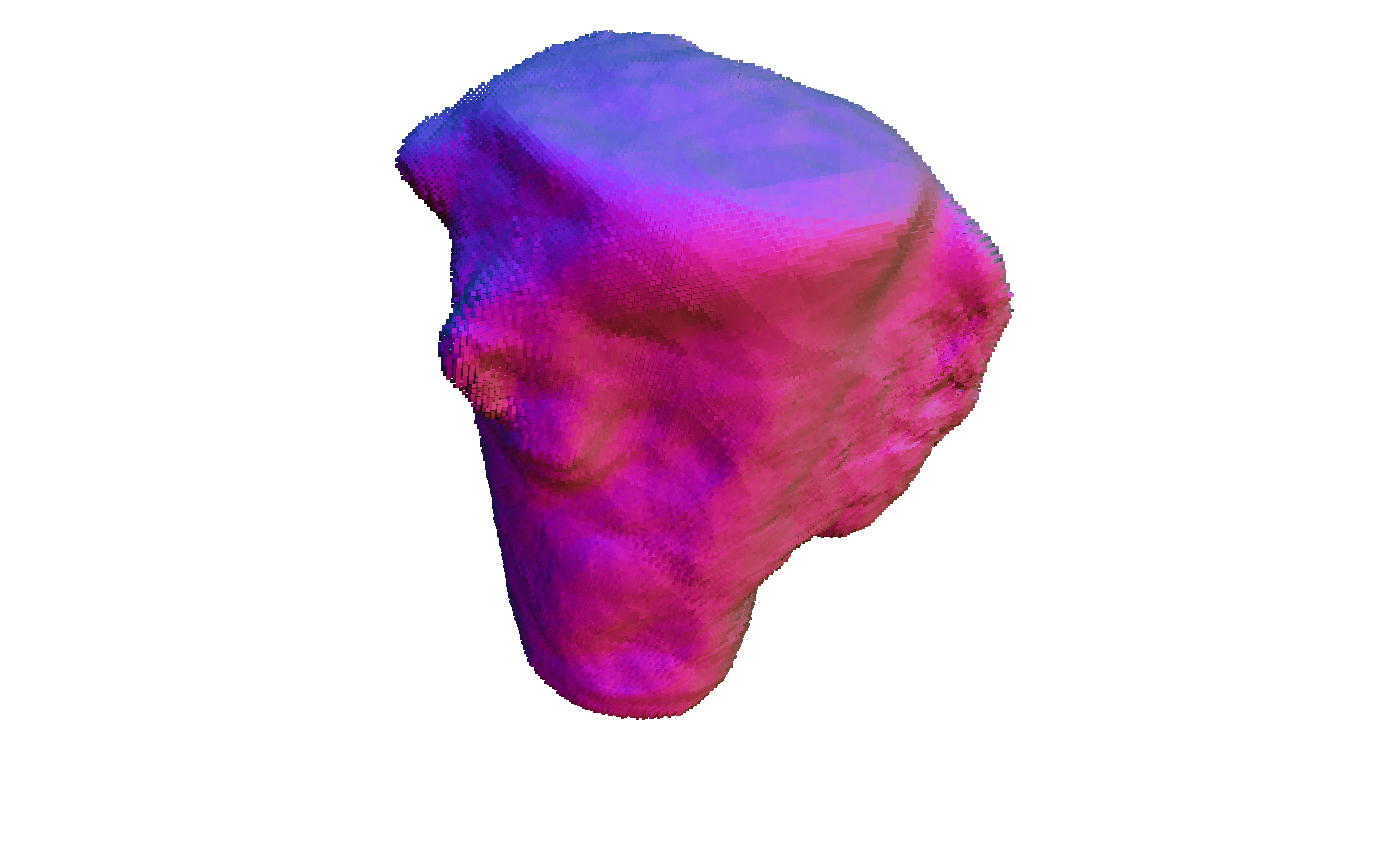} &
    \includegraphics[width=0.24\linewidth,trim={60, 0, 60, 0},clip]{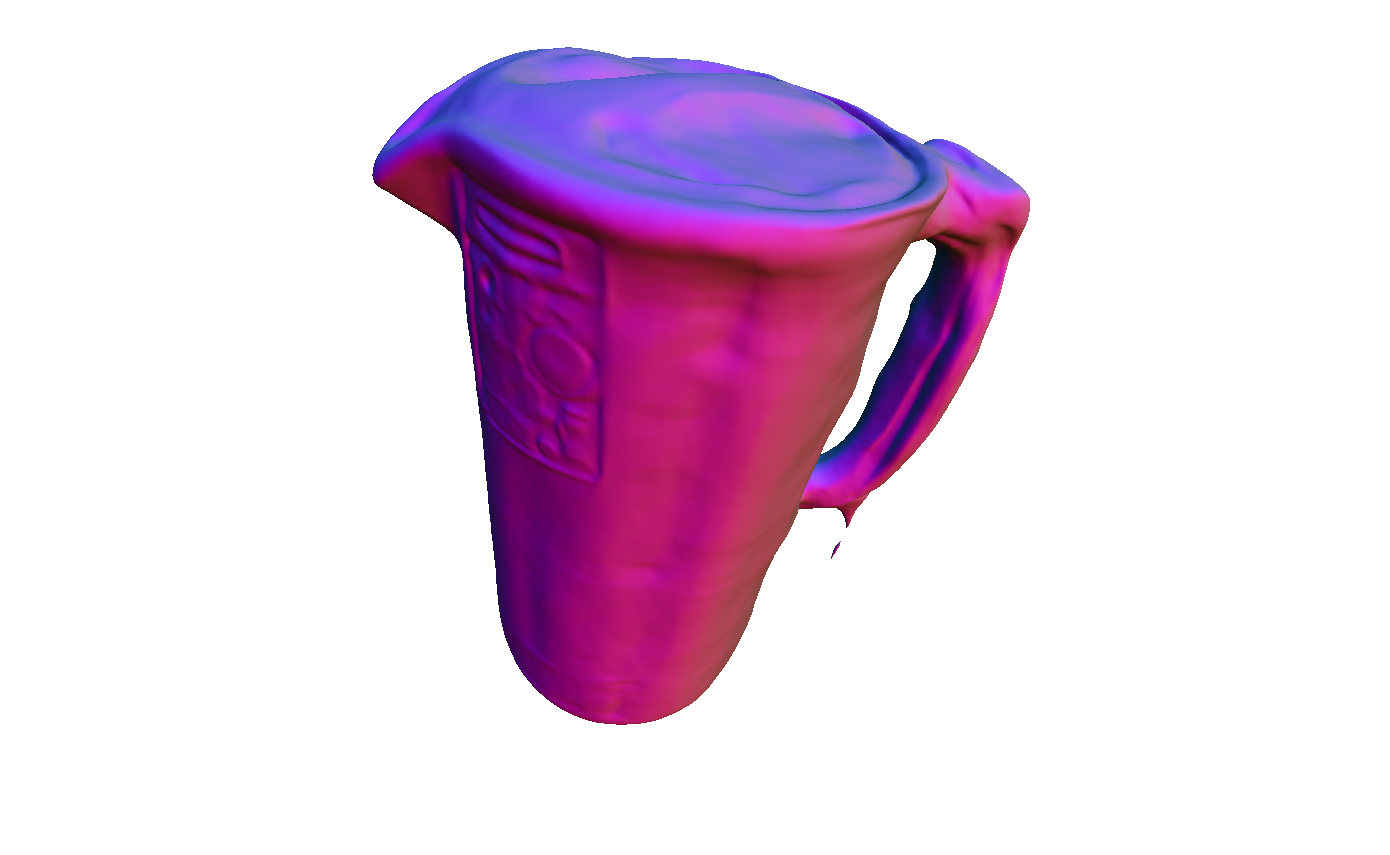} 
    \\
     \small{Reference} & \small{NeuS} & \small{Hampali \etal.} & \small{Ours}\\
    \end{tabular}
    \endgroup
  \caption{\textbf{Reconstruction result on HO3D.}
  Our method produces significantly better mesh results compared to Hampali~\etal.'s method~\cite{meta-obj}, and is on par with NeuS~\cite{neus} which is trained with ground truth poses.
  }
  \label{fig:recon_comp}
\end{figure}

We summarize the reconstruction results in Table~\ref{tab:recon_comp_ho3d}.
COLMAP can not produce accurate results on most objects, mainly caused by the lacking of enough textures, especially for objects like bleach and pitcher in the table.
With prior hand pose information, Ye’s method produces better results than COLMAP. With carefully selected multiple easy segments, Hampali’s method outperforms Ye's method. Nevertheless, our method produces better results than most of them, and is even on par with methods relying on ground truth poses (UNISURF and NeuS) or depth images (Patten's method). Fig.~\ref{fig:recon_comp} shows some visualization results of the reconstruction meshes.

Table~\ref{tab:pose_comp_ho3d} shows the evaluation of pose results. Our method outperforms COLMAP and Hampali’s method significantly.

\begin{table}%
  \centering
  \begin{tabular}{lcccc}
    \toprule
    \multirow{2}{*}{Strategy}&  \multicolumn{3}{c}{Pose} & \multicolumn{1}{c}{Mesh} \\
     & $ \!\!\!\!\!\!\!AUC_{ATE}\uparrow\!\!\!$ & $RPE_{t}\downarrow\!\!\!$ & $RPE_{r}\downarrow\!\!\!$ & $HD_{RMSE}\downarrow\!\!\!$\\
    \midrule
    RC & 4.33& 4.14& 7.19& 4.15 \\
    VC & 5.42& 1.94& 2.65& 3.90 \\
    VC + RC & \textbf{5.93}& \textbf{1.57}& \textbf{2.20}& \textbf{3.14}\\
  \bottomrule
  \end{tabular}
    \caption{\textbf{Effect of virtual cameras.}
    Progressively training w.r.t. the real camera~(RC) typically fails. The proposed virtual camera system~(VC) reduces the search space and improves the results significantly. The global refinement w.r.t. the real camera~(``+RC'') improves the performance further.
    }
    \label{table:virtual_abalation}
\end{table}

\begin{table}%
  \centering
  \begin{tabular}{lcccccc}
    \toprule
    \multirow{2}{*}{Settings} & \multicolumn{3}{c}{Pose} & \multicolumn{1}{c}{Mesh} \\
     & $ \!\!\!\!\!\!\!AUC_{ATE}\uparrow\!\!\!$ & $RPE_{t}\downarrow\!\!\!$ & $RPE_{r}\downarrow\!\!\!$ & $HD_{RMSE}\downarrow\!\!\!$\\
    \midrule
    w/o match & 5.21& 1.87& 2.38& 4.10 \\
    w/o reset & 5.58& 1.64& 2.26& 3.28 \\
    w/o 4D & 5.68& 2.26& 3.49& 3.27\\
    Full &\textbf{ 5.93}& \textbf{1.57}& \textbf{2.20}& \textbf{3.14}\\
  \bottomrule
  \end{tabular}
    \caption{\textbf{Ablation study.}
    We evaluate the effectiveness of the proposed components, including reducing the 6 degrees of freedom of poses to only 4~(``w/o 4D''), the periodical reset of shape networks~(``w/o reset''), and the match loss~(``w/o match'').
    }
    \label{table:ablation_study_1}
\end{table}

\begin{figure}[tb]
  \centering
  \begingroup
  \setlength{\tabcolsep}{1pt}
    \begin{tabular}{cccc}
   \includegraphics[width=0.24\linewidth]{Experiment_Figure/ho3d_ref/AP13.png} &
   \includegraphics[width=0.24\linewidth,trim={80, 0, 80, 0},clip]{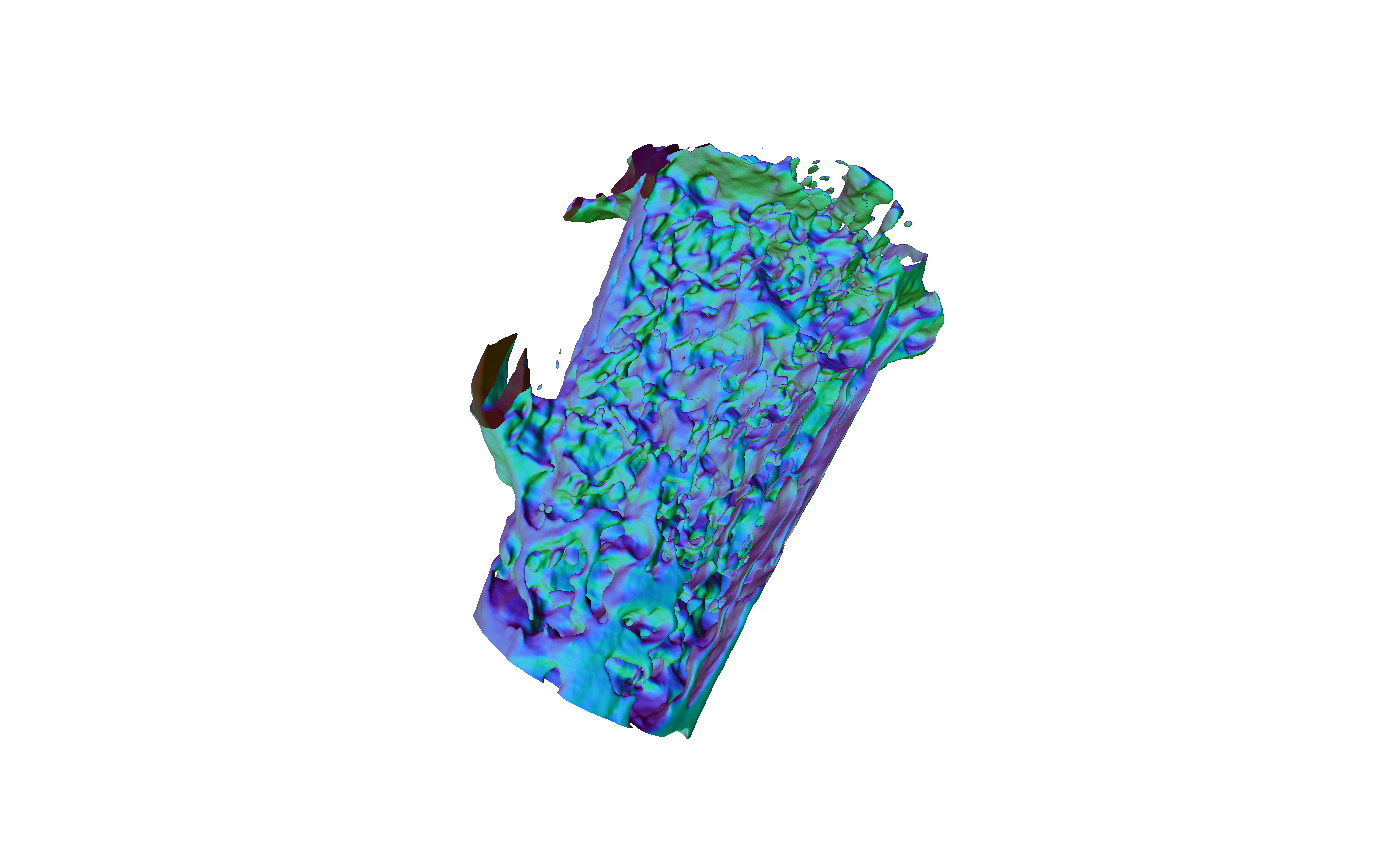} &
   \includegraphics[width=0.24\linewidth,trim={80, 0, 80, 0},clip]{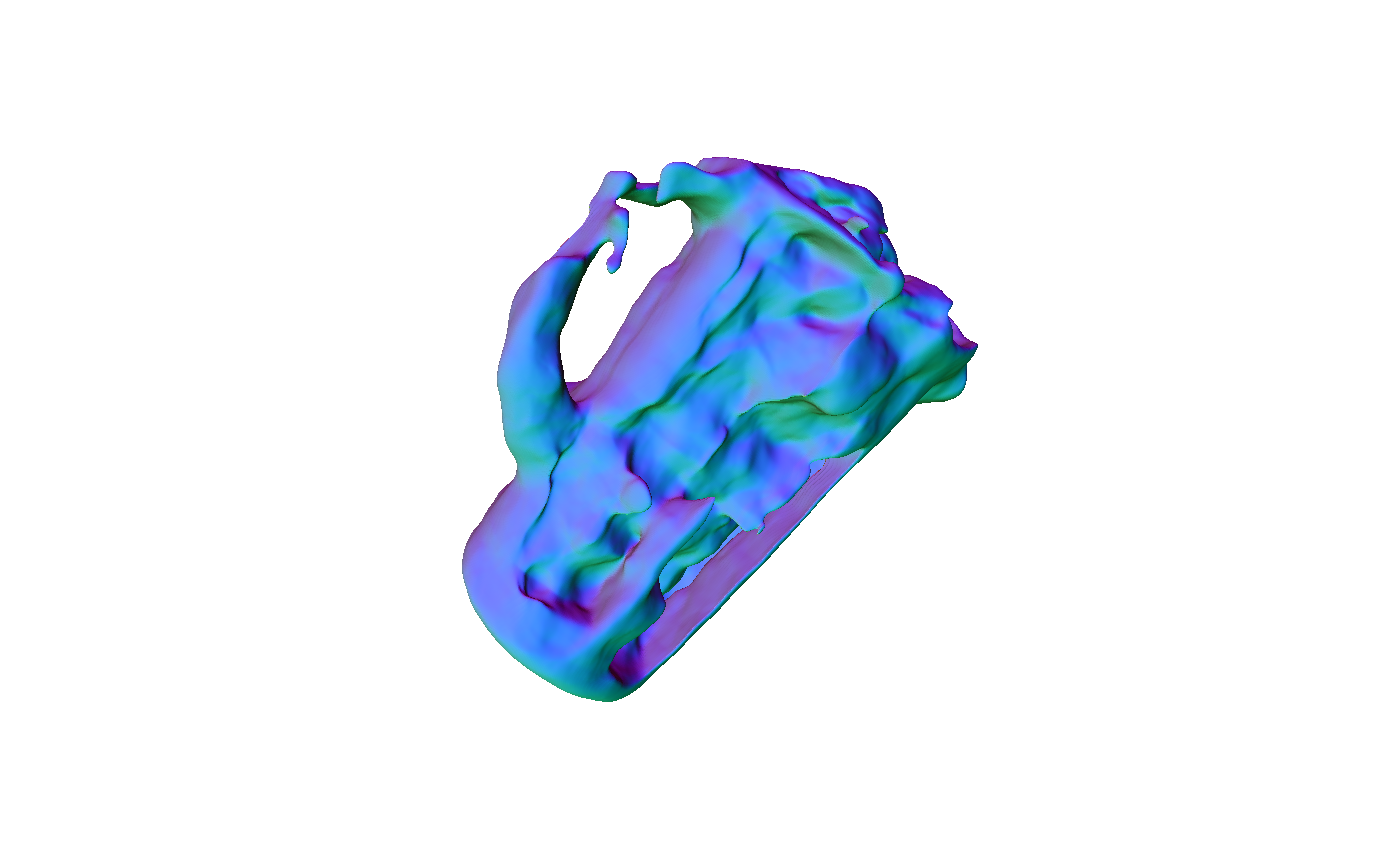} &
   \includegraphics[width=0.24\linewidth,trim={80, 0, 80, 0},clip]{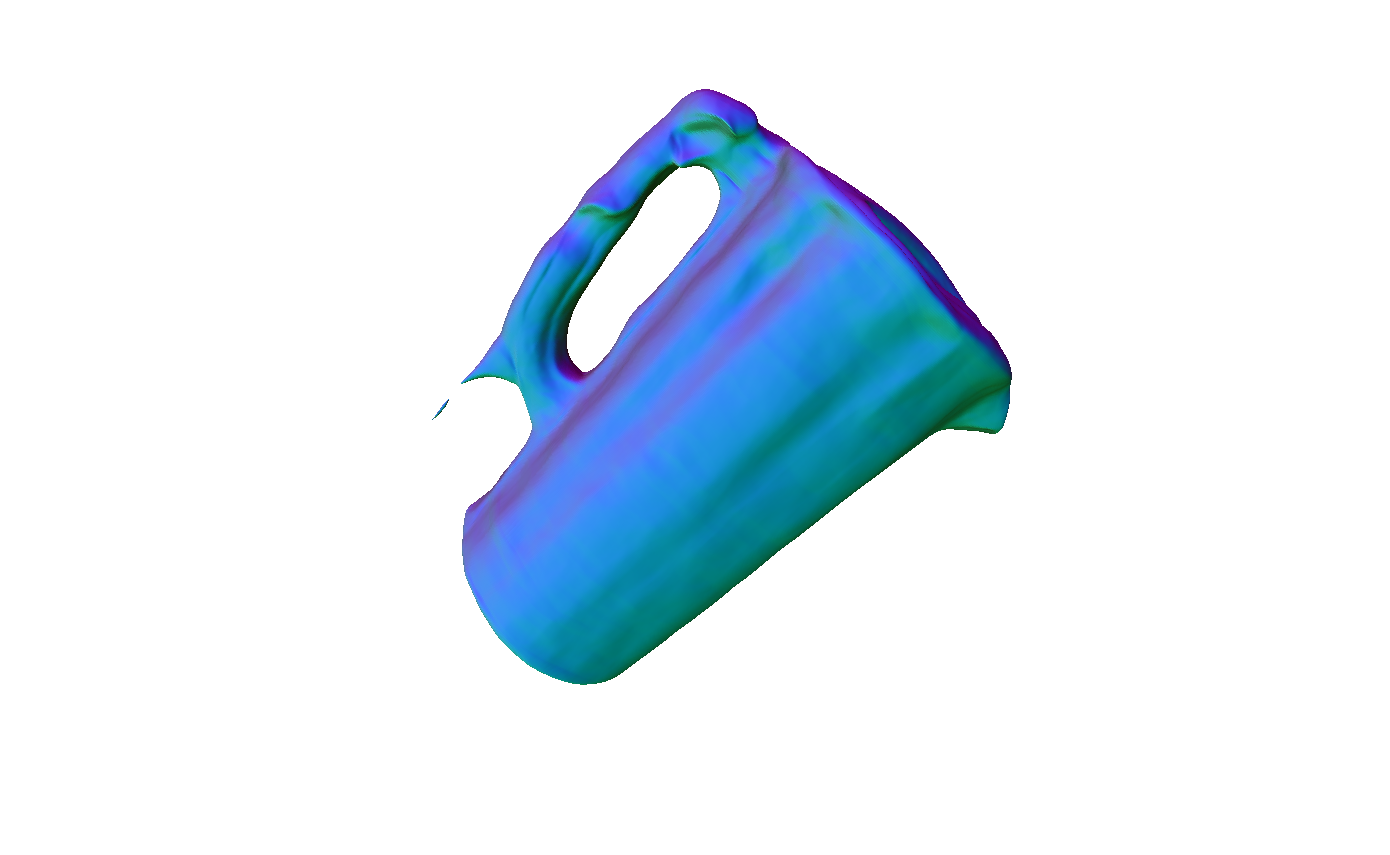} \\

\includegraphics[width=0.24\linewidth]{Experiment_Figure/ho3d_ref/MDf14} &
   \includegraphics[width=0.24\linewidth]{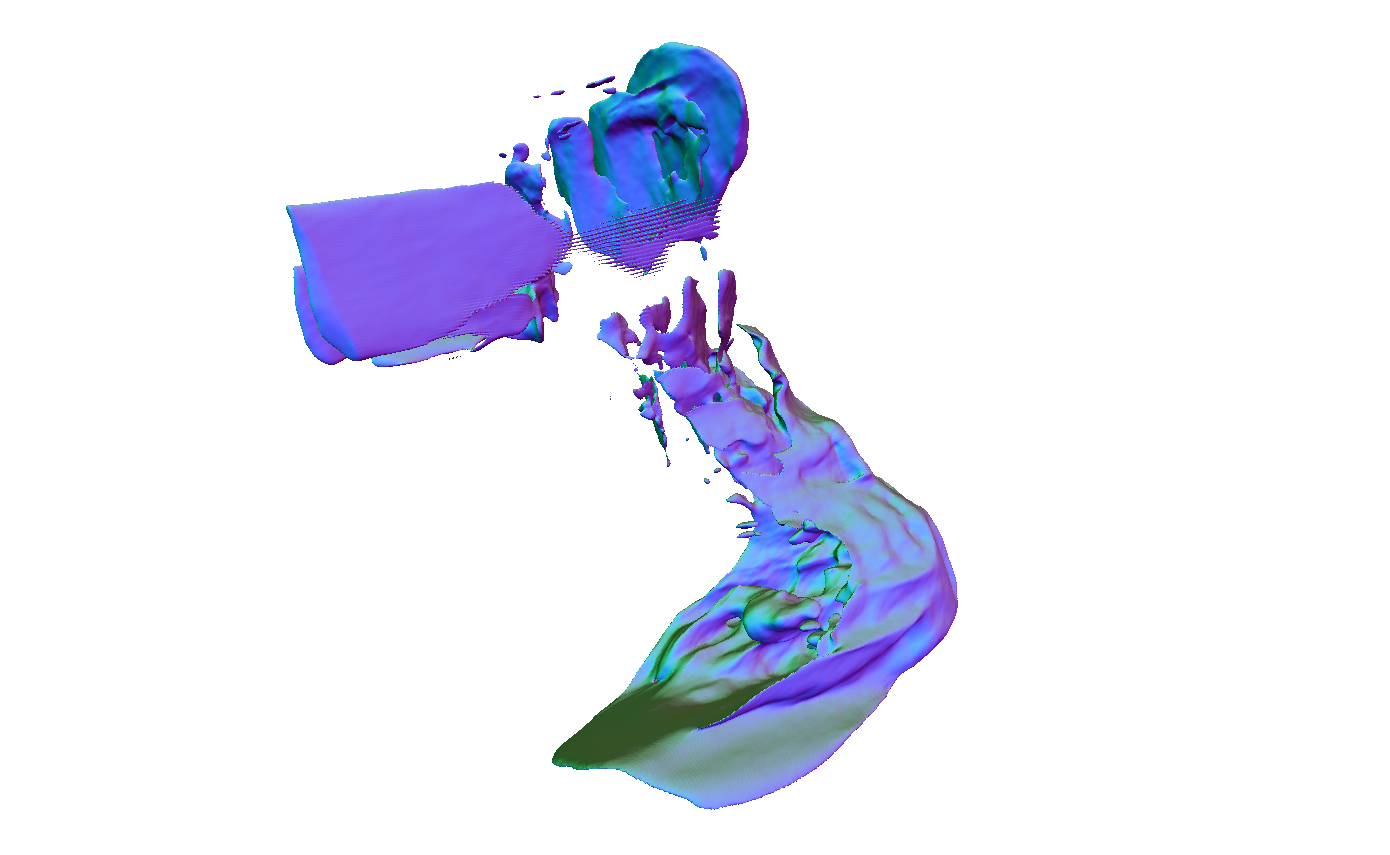} &
   \includegraphics[width=0.24\linewidth]{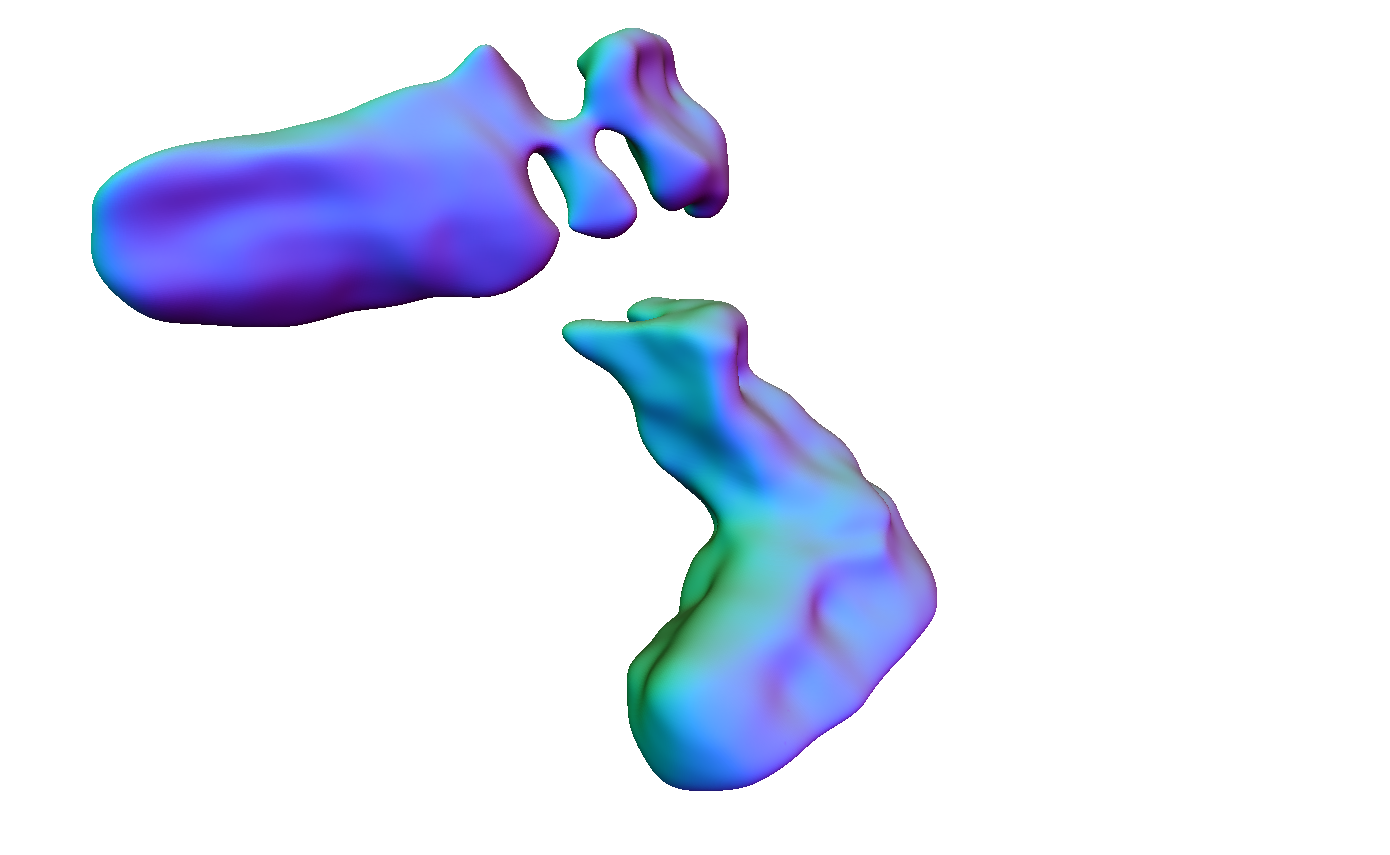} &
   \includegraphics[width=0.24\linewidth]{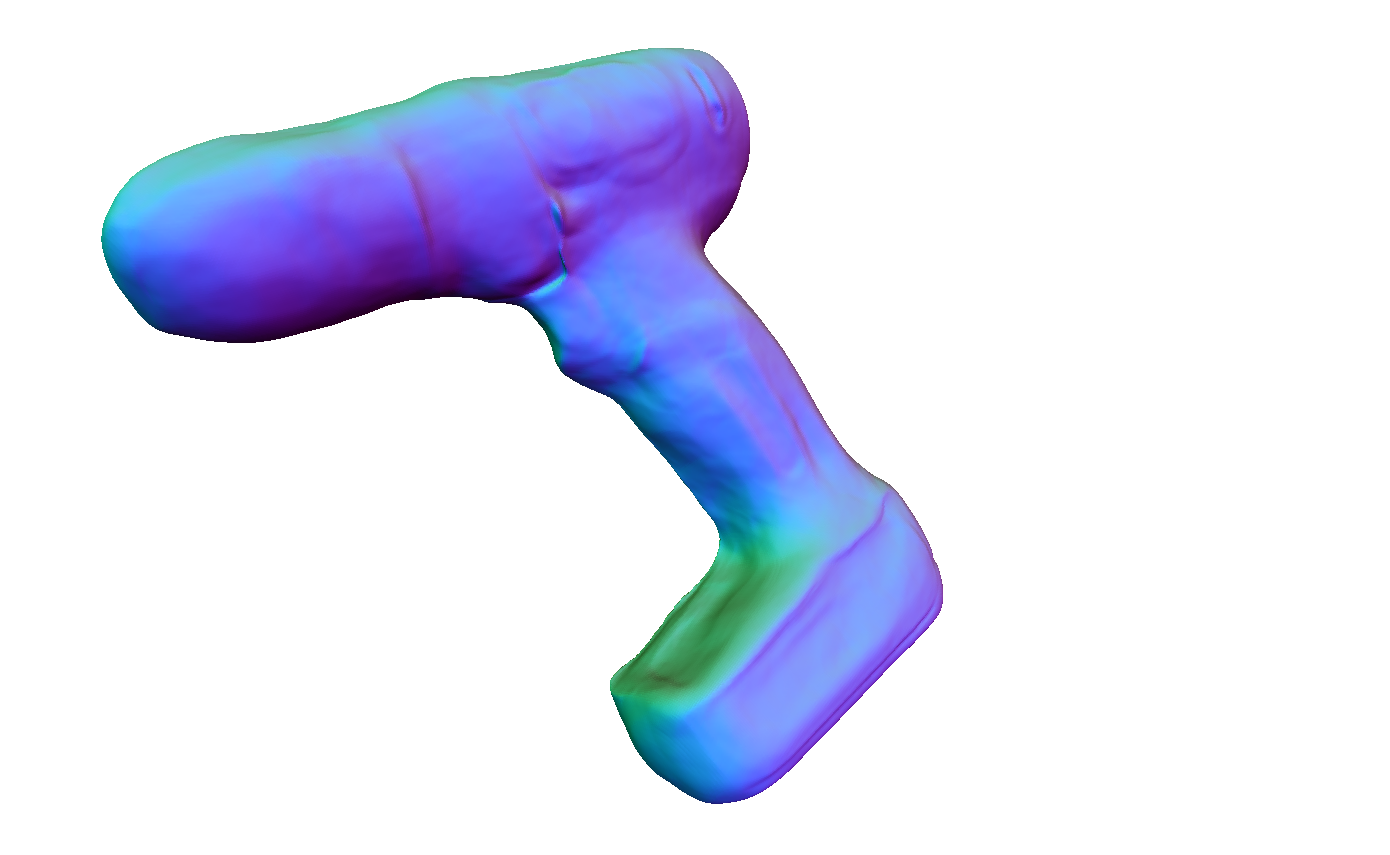} \\

   \small{Reference} & \small{RC} & \small{VC} & \small{VC + RC} \\
   \end{tabular}
   \endgroup
  \caption{
  {\bf Effect of virtual cameras.}
      Progressively training w.r.t. the real camera~(RC) is challenging and typically fails. The proposed virtual camera system~(VC) reduces the search space of optimization and improves the results significantly after refinement w.r.t. the real camera~(``+RC'').
  }
  \label{fig:effect_of_vc}
\end{figure}


\subsection{Ablation Study}
\begin{figure}
  \centering
  \begingroup
  
  \setlength{\tabcolsep}{1pt}
  \centering
    \begin{tabular}{cccc}
   \includegraphics[width=0.24\linewidth,trim={120, 30, 120, 0},clip]{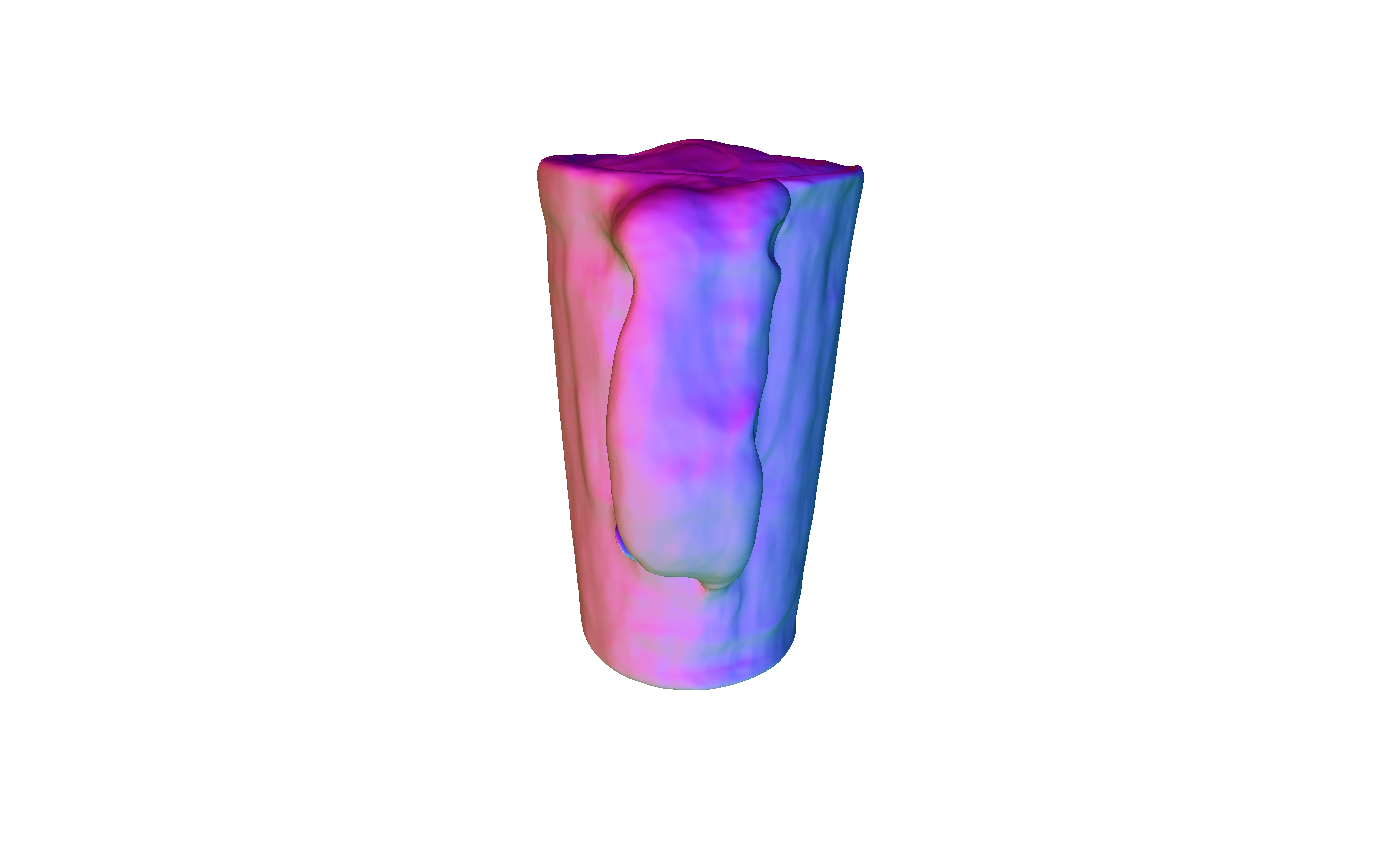} &
    \includegraphics[width=0.24\linewidth,trim={120, 30, 120, 0},clip]{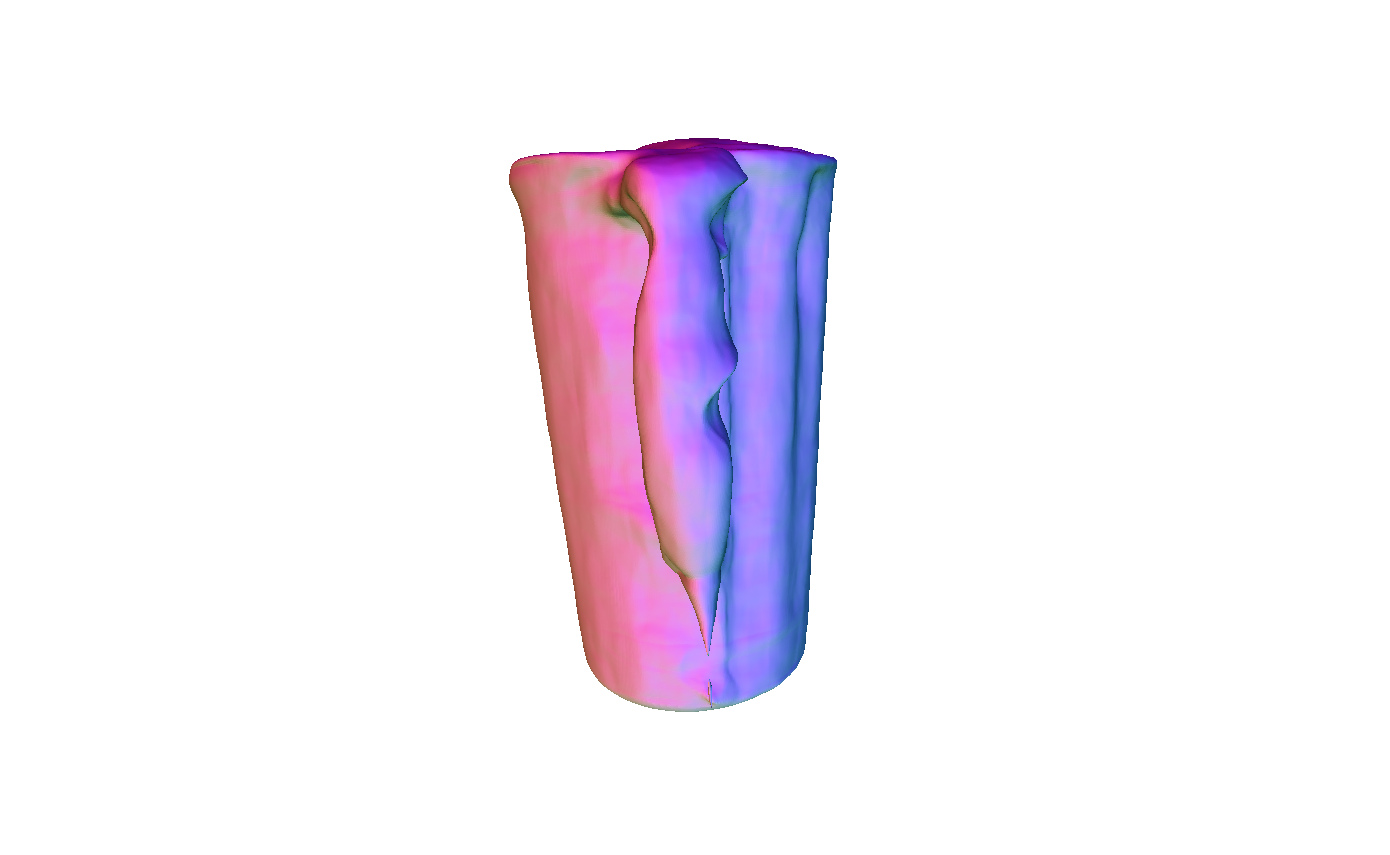} &
       \includegraphics[width=0.24\linewidth,trim={120, 20, 120, 0},clip]{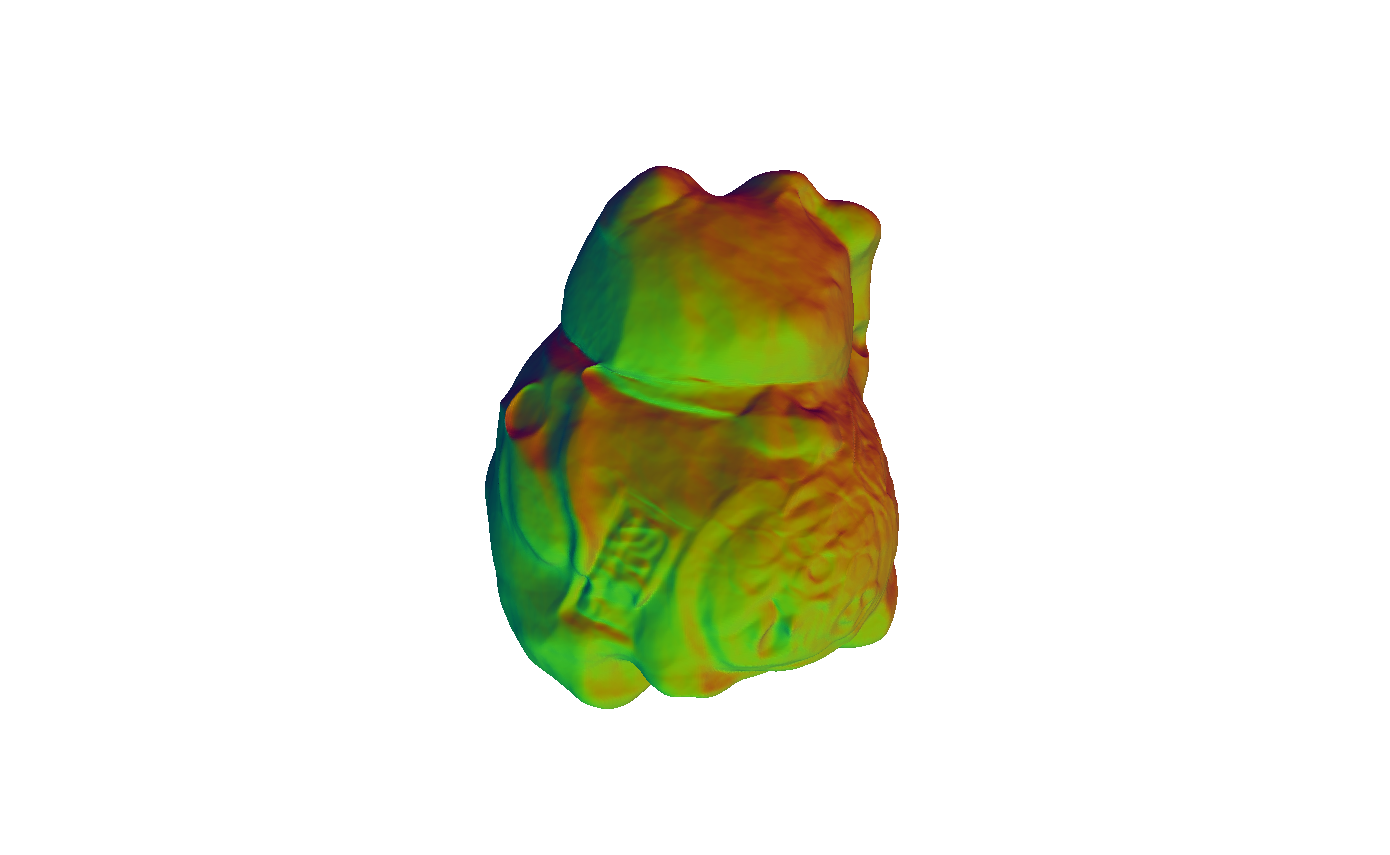} &
    \includegraphics[width=0.24\linewidth,trim={120, 20, 120, 0},clip]{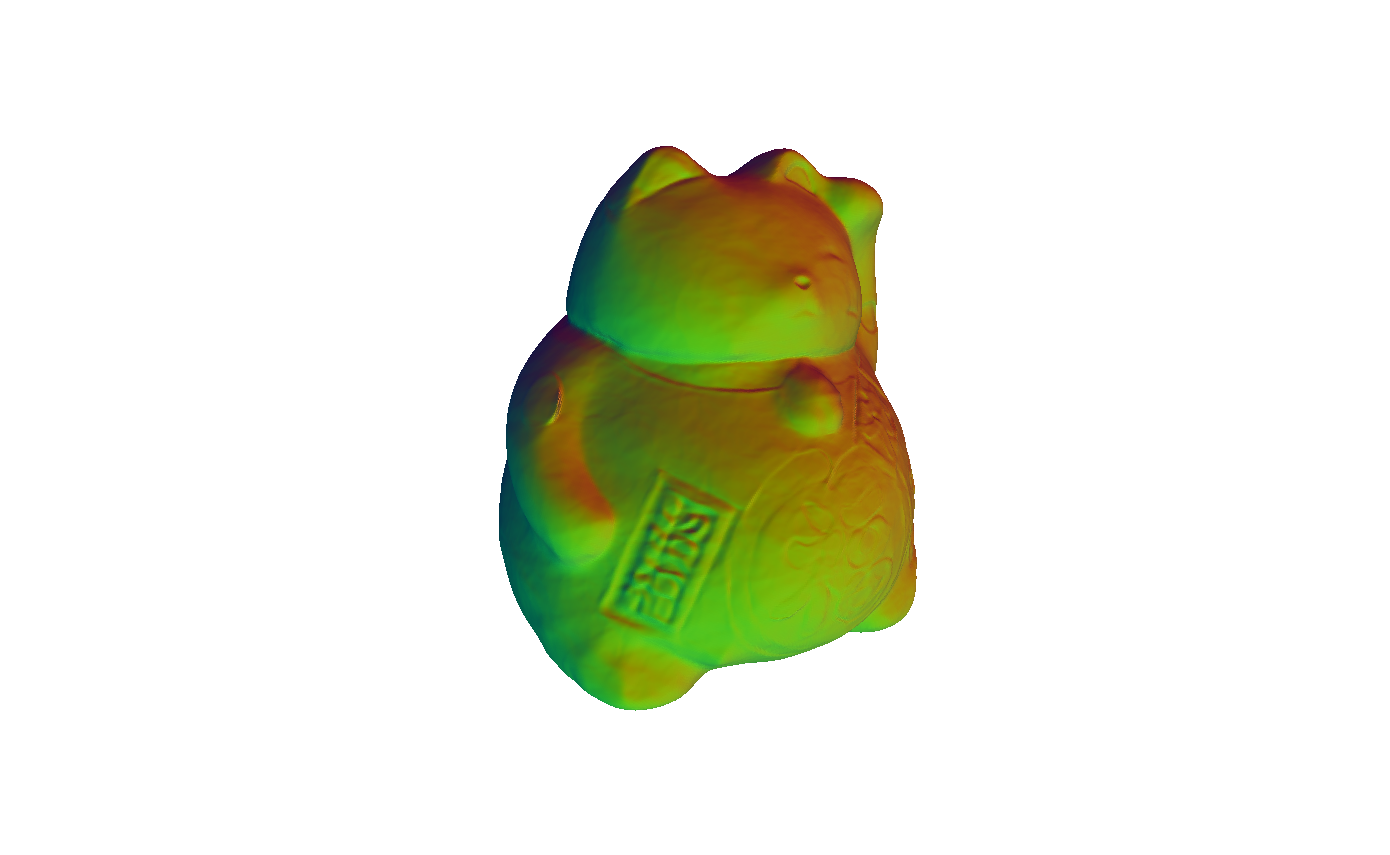}
    \\
   \includegraphics[width=0.24\linewidth,trim={120, 30, 120, 0},clip]{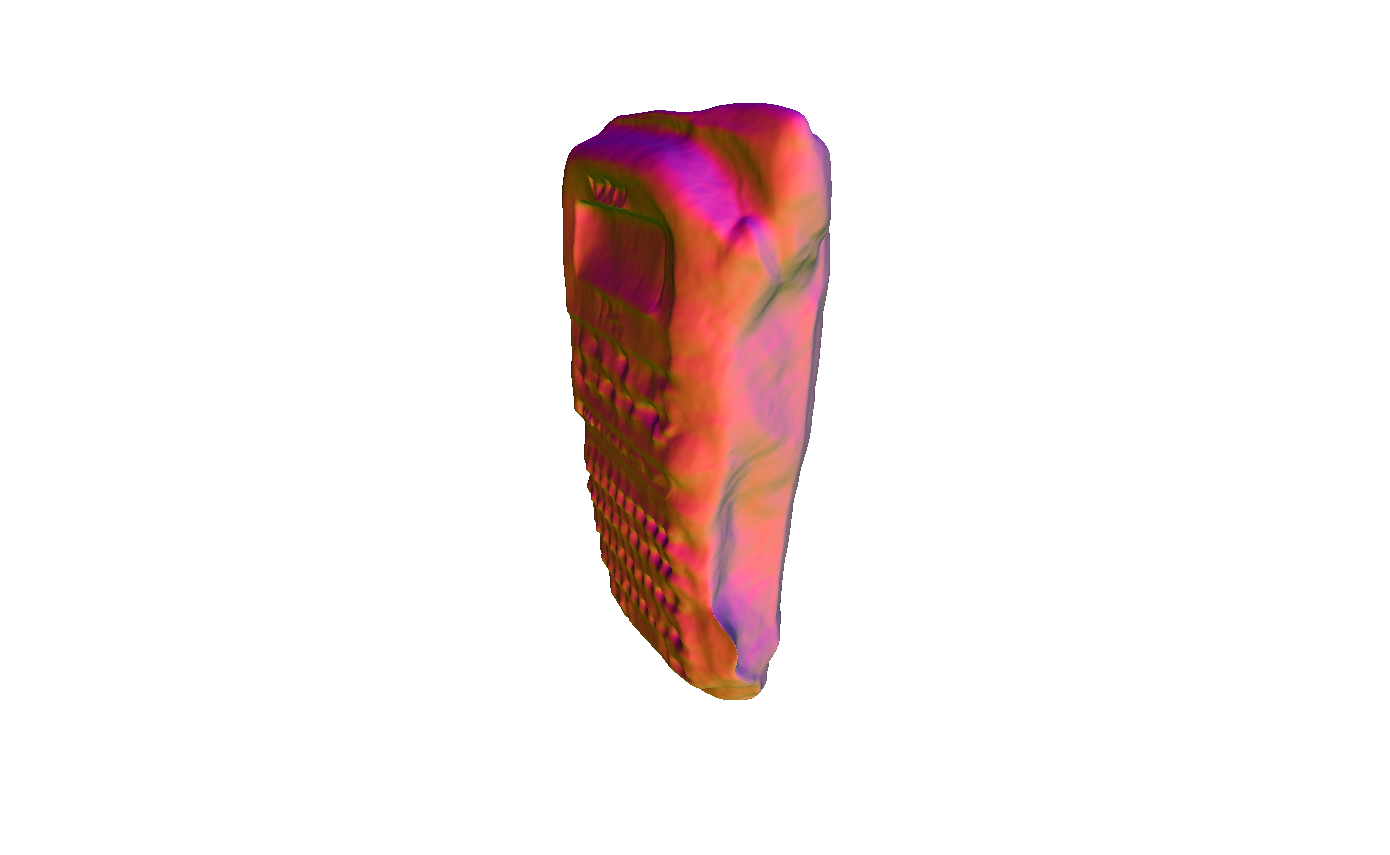} &
    \includegraphics[width=0.24\linewidth,trim={120, 30, 120, 0},clip]{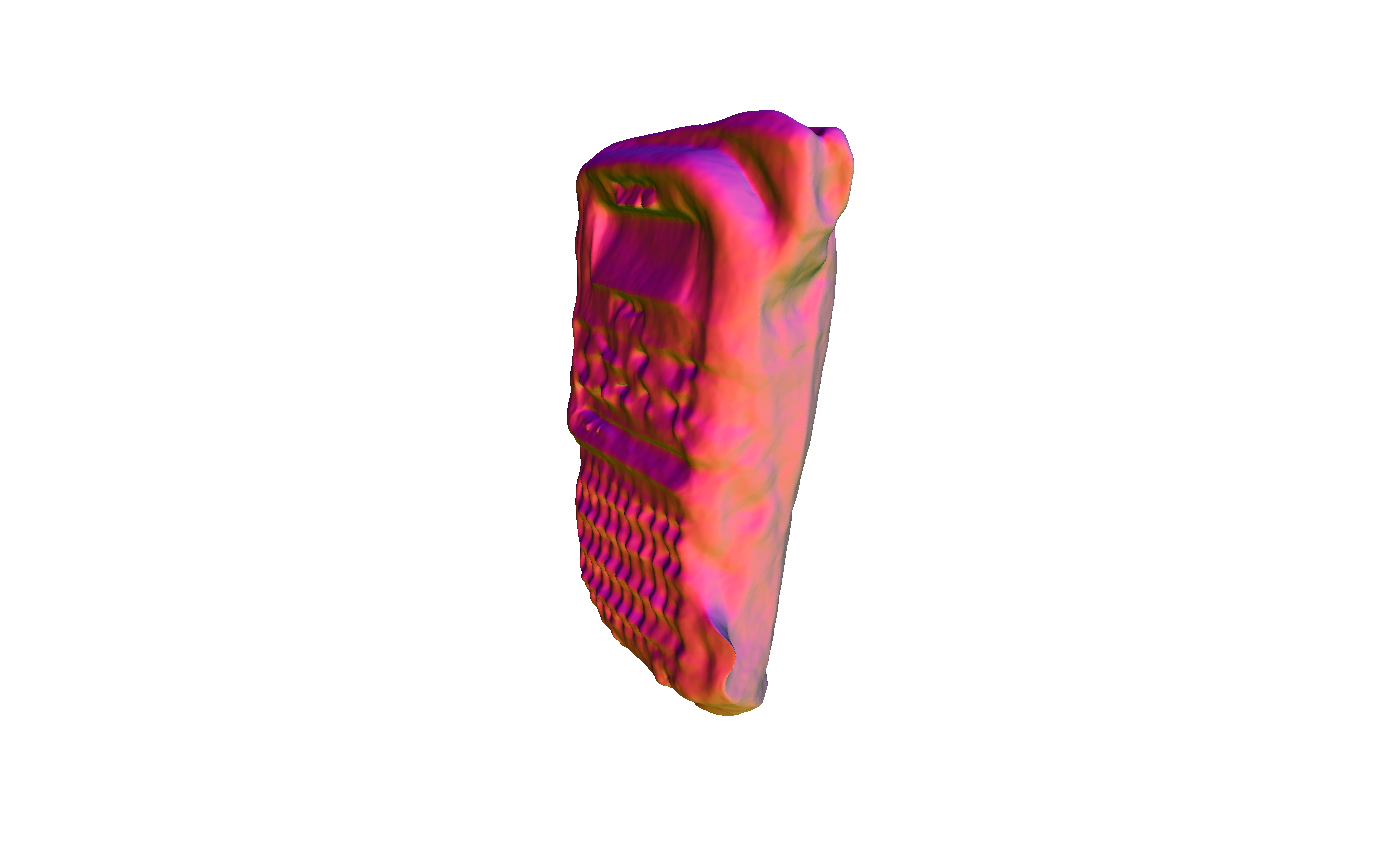} &
   \includegraphics[width=0.24\linewidth,trim={150, 80, 150, 0},clip]{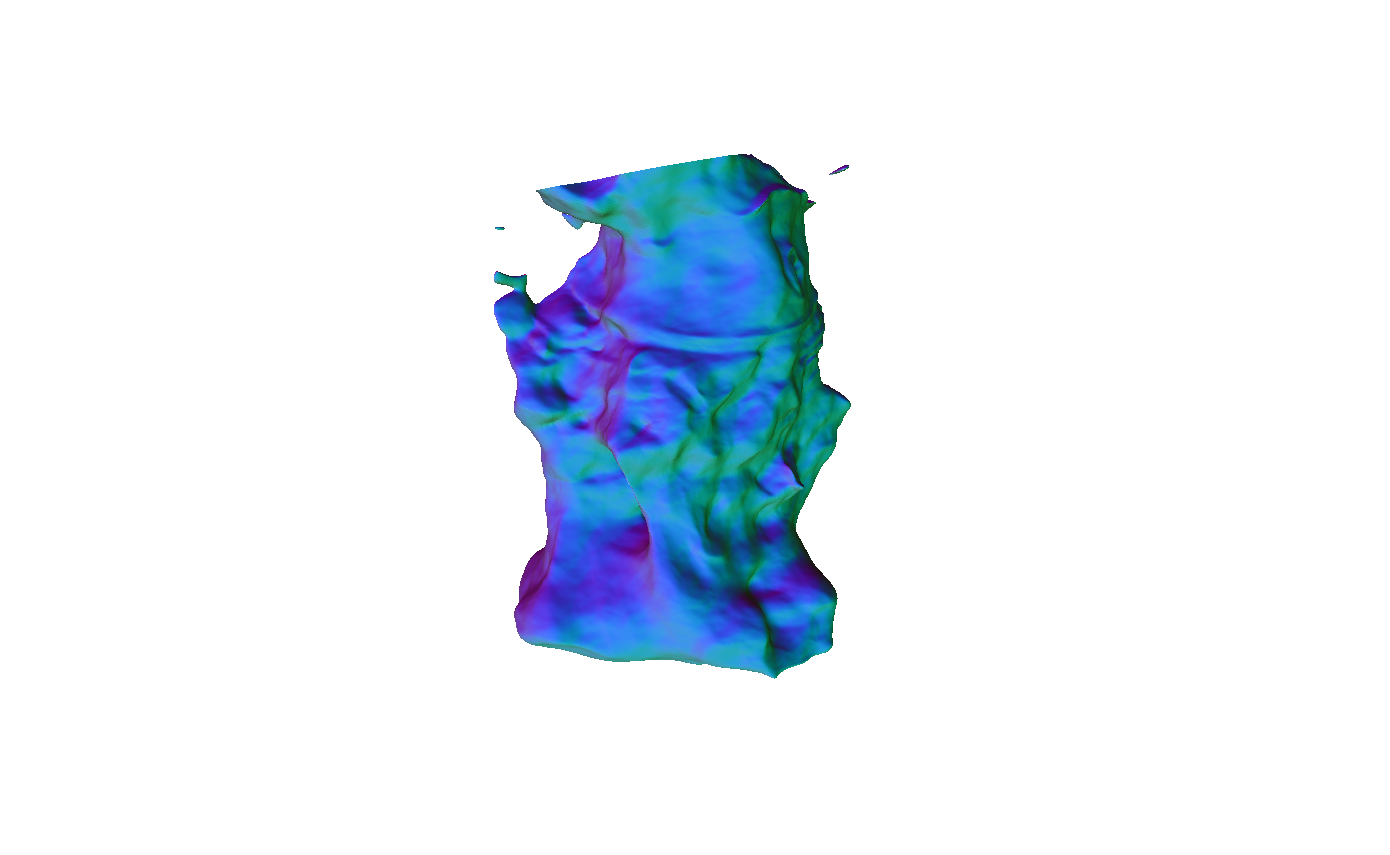} &
    \includegraphics[width=0.24\linewidth,trim={150, 80, 150, 0},clip]{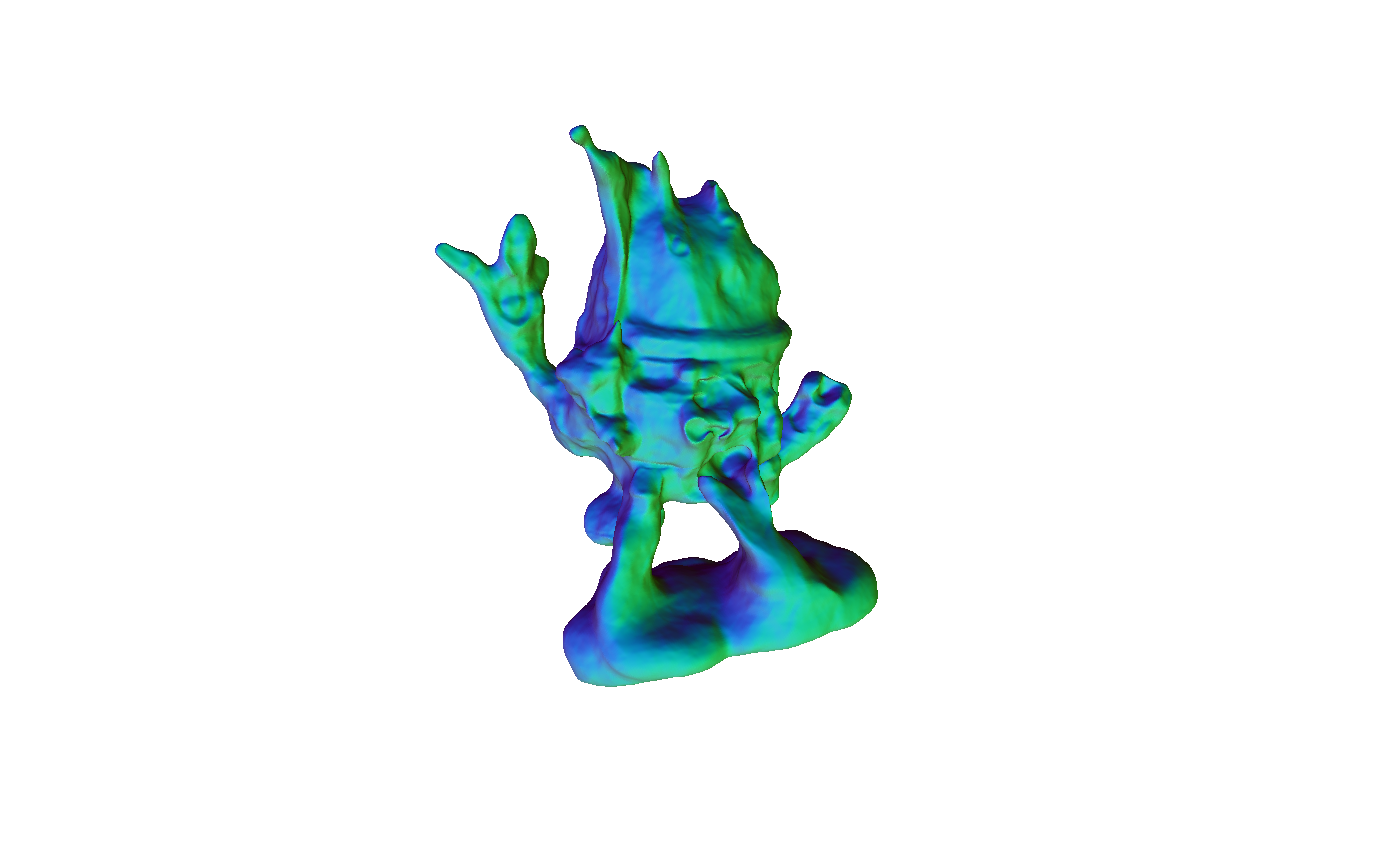}
    \\
\multicolumn{1}{c}{{\cellcolor{red!10} \small{w/o match}}} & \multicolumn{1}{c}{{\cellcolor{green!10} \small{w/ match}}} & \multicolumn{1}{c}{{\cellcolor{red!10} \small{w/o reset}}} & \multicolumn{1}{c}{{\cellcolor{green!10} \small{w/ reset}}} \\
   \end{tabular}
   \endgroup
  \caption{
  {\bf Visualization result of ablation study.}
  The network benefits from the match loss~(``w/ match'') for recovering detailed shapes, while the periodic reset of shape networks~(``w/ reset'') aids optimization in handling local minima.
  }
  \label{fig:mini_ablations_visual}
\end{figure}

We evaluate the design of our method systematically in this section. We report the results averaging across all the 9 sequences in HO3D, if not explicitly mentioned.

\noindent {\bf Effectiveness of guided virtual camera.}
We study the effect of the proposed virtual camera in Table~\ref{table:virtual_abalation}.
In principle, one could leverage the temporal consistency between consecutive frames simply by introducing the progressive training, similar to~\cite{nope-nerf,3dgs}. We compare this strategy with our method. For a fair comparison, we use the same network and the same segmentation mask and 2D matches as those used in our method. The major problem of this strategy is that the pose and shape are optimized w.r.t. the raw camera, which is challenging due to the target’s free movement and the wide range it covers in front of the camera. We denote this method as ``RC'' in the table. By contrast, the proposed virtual camera reduces the search space and improves the results significantly~(``VC''). The result improves further with the refinement w.r.t the real camera~(``+RC''). Fig.~\ref{fig:effect_of_vc} shows some visualization results.

\noindent {\bf Ablation study of different components.} We study the effect of different components of our method in Table~\ref{table:ablation_study_1}.
Without the match loss or the periodic reset of the shape network, the performance has a significant drop~(denoted as ``w/o match'' and ``w/o reset'' respectively). On the other hand, using the standard 6D pose representation~(denoted as ``w/o 4D'') also suffers in performance.
We use the same refinement procedure for all the results in this table. Fig.~\ref{fig:mini_ablations_visual} shows some visualization results.

\begin{figure}
  \centering
  \begingroup

    \setlength{\tabcolsep}{0pt}
    {\includegraphics[width=\linewidth]{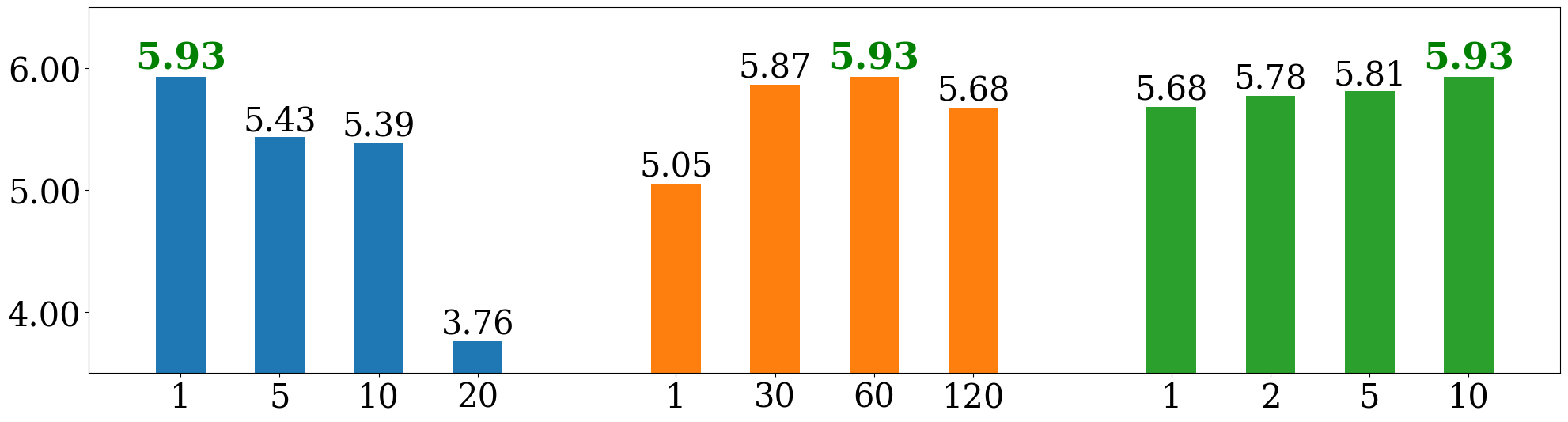}} \\
    \setlength{\tabcolsep}{0pt}
    \begin{tabular}{>{\centering\arraybackslash}p{0.06\linewidth}>{\centering\arraybackslash}p{0.30\linewidth}>
    {\centering\arraybackslash}p{0.02\linewidth}>
    {\centering\arraybackslash}p{0.30\linewidth}>
    {\centering\arraybackslash}p{0.02\linewidth}>{\centering\arraybackslash}p{0.30\linewidth}}
            &\small{Group size} && \small{Reset degree} & &\small{Match interval}
    \end{tabular}
  \endgroup
  \caption{\textbf{Ablation study of hyper-parameters in $AUC_{ATE}\uparrow$.}
We report the average results of all HO3D sequences. It shows that processing one frame per group in progressive training, resetting shape networks after every 60 degrees, and relying on 2D matches across 10 consecutive frames is the best setting in general.
  }
  \label{fig:ablation_study_2}
\end{figure}

\noindent {\bf Ablation study of different hyper-parameters.}
We show ablation results of our method with different hyper-parameters in Table~\ref{fig:ablation_study_2}, including the number of frames $B$ that is processed as a group during progressing training, the degree threshold $\tau$ for periodic reset of the shape network, and the the maximum frame internal $n$ for 2D matching across consecutive images.
As we can see, $B$=1 gives the best results, and as the number of images in each group increases, the performance deteriorates, which we believe is caused by the difficulties introduced with the increased data varieties. On the other hand, $\tau$=60$\degree$ and $n$=10 produce the best results in most of our experiments.

\begin{figure}[tb!]
  \centering
      \begingroup
  \setlength{\tabcolsep}{1pt}
    \begin{tabular}{cccc}
    \includegraphics[width=0.24\linewidth]{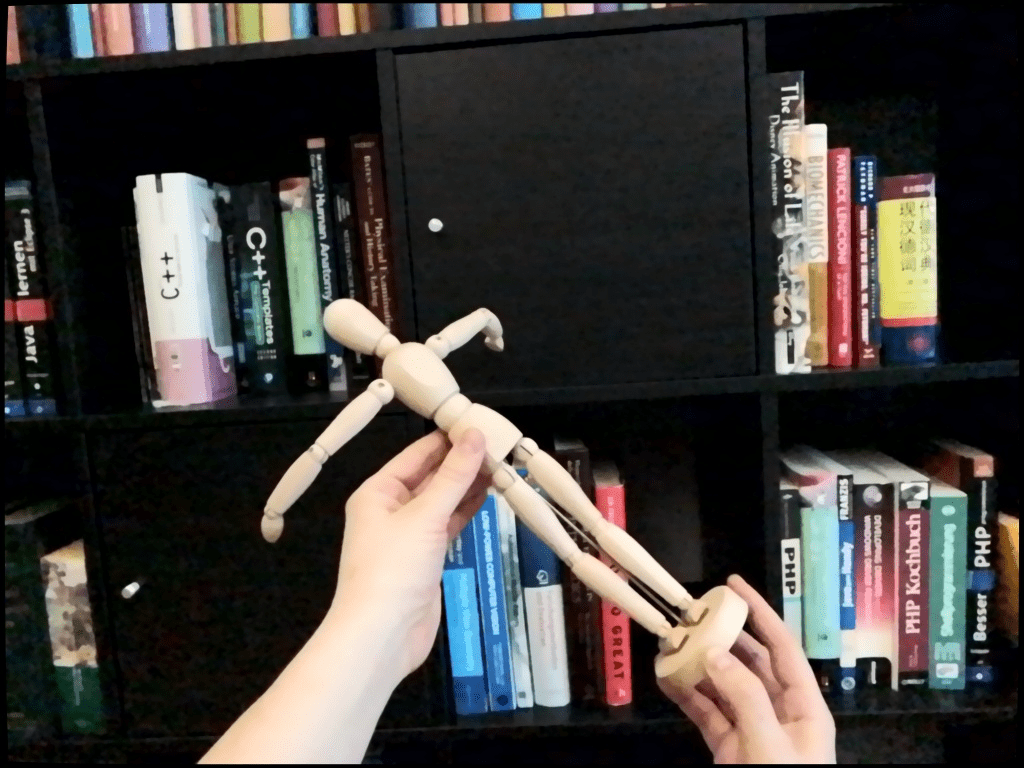} &
    \includegraphics[width=0.24\linewidth,trim={0, 0, 80, 0},clip]{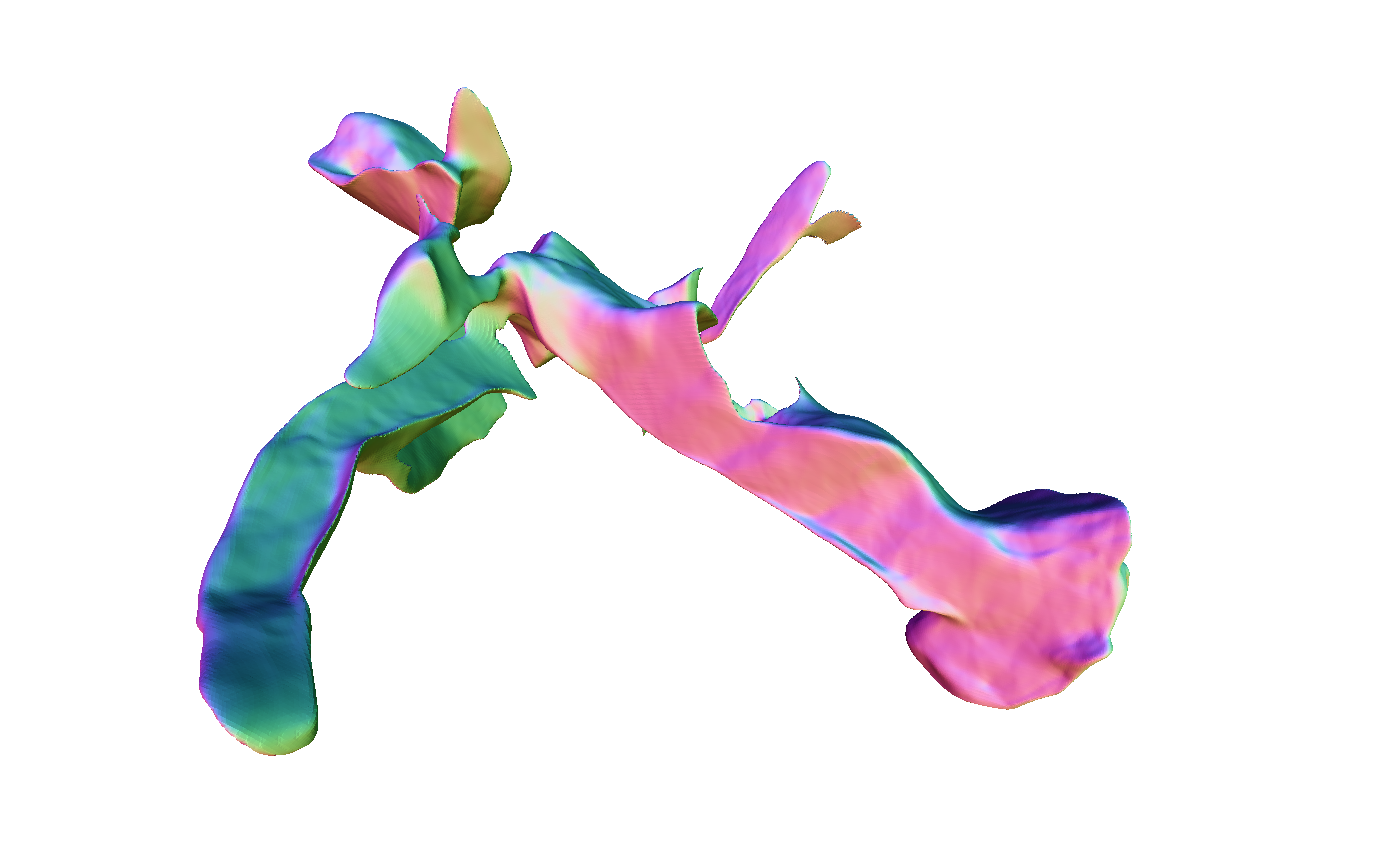} &
    \includegraphics[width=0.24\linewidth,trim={30, 0, 30, 0},clip] {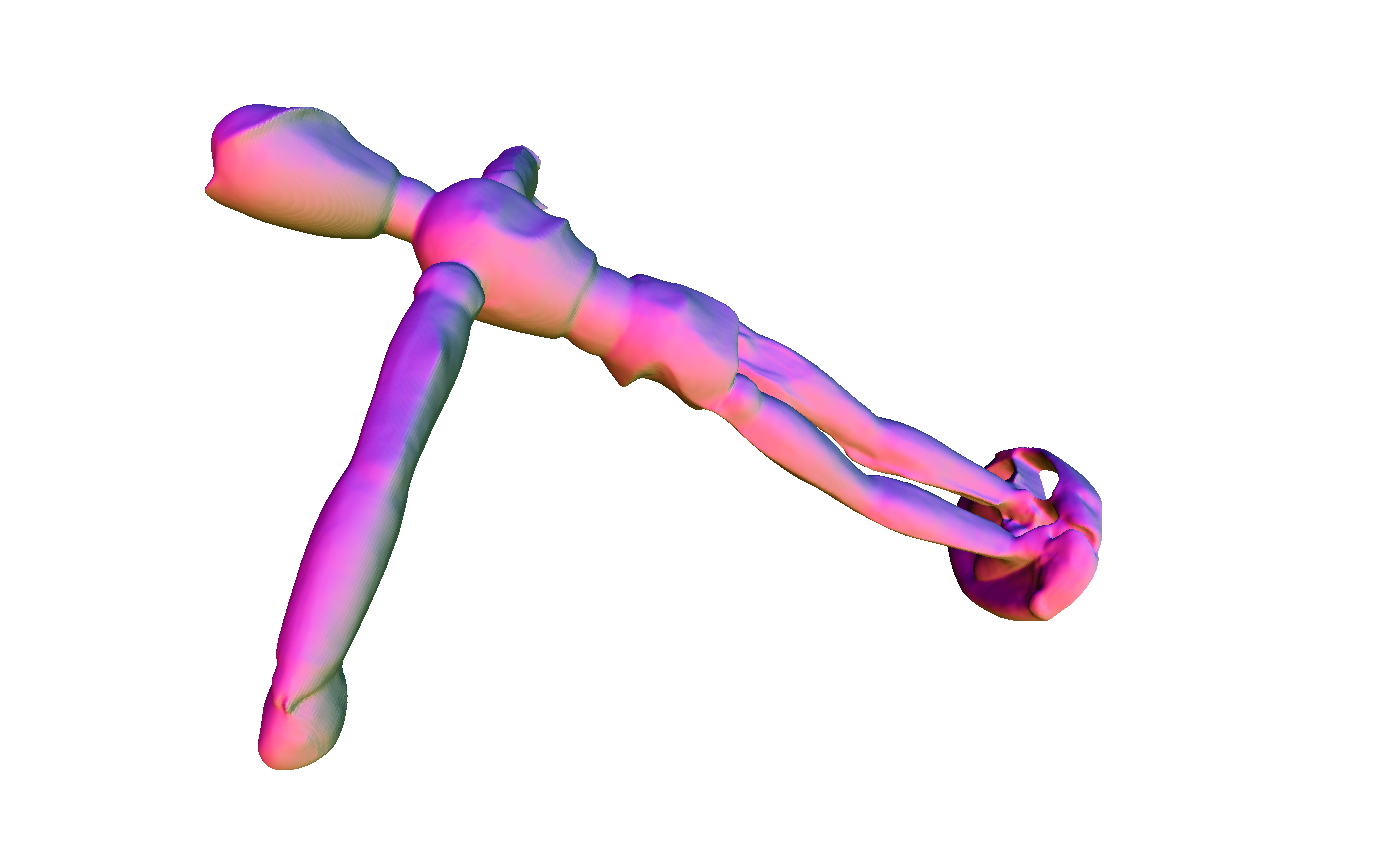}&
        \includegraphics[width=0.20\linewidth]{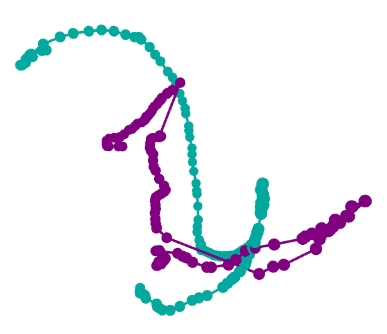} \\
    
    \includegraphics[width=0.24\linewidth]{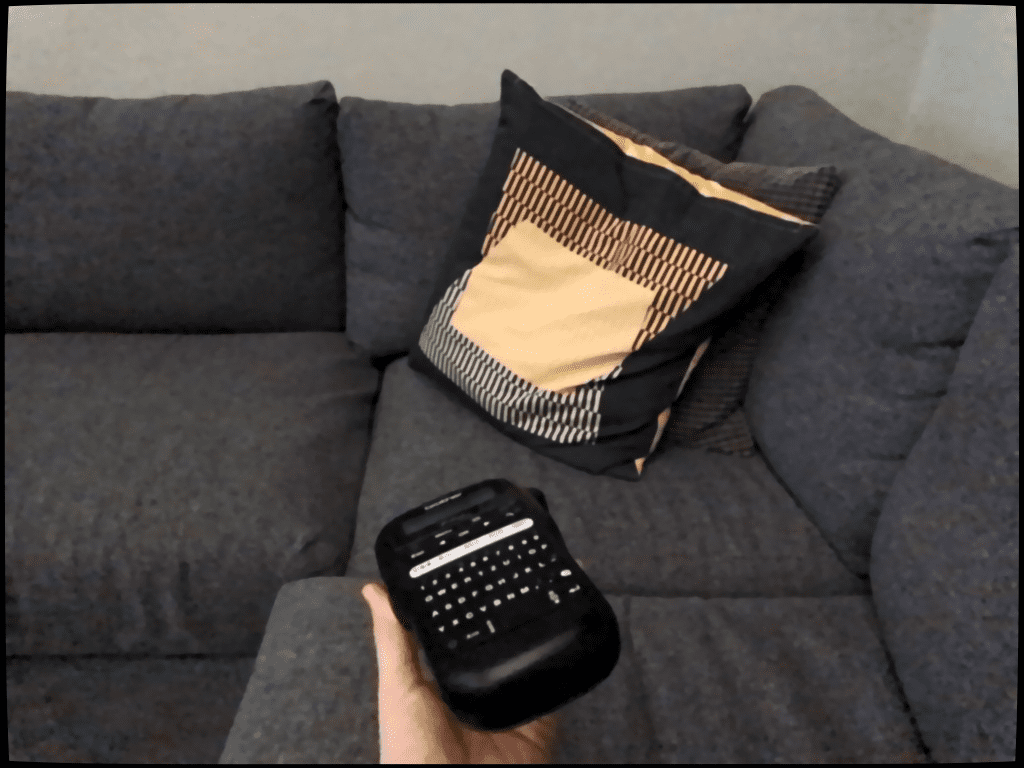} &
    \includegraphics[width=0.24\linewidth]{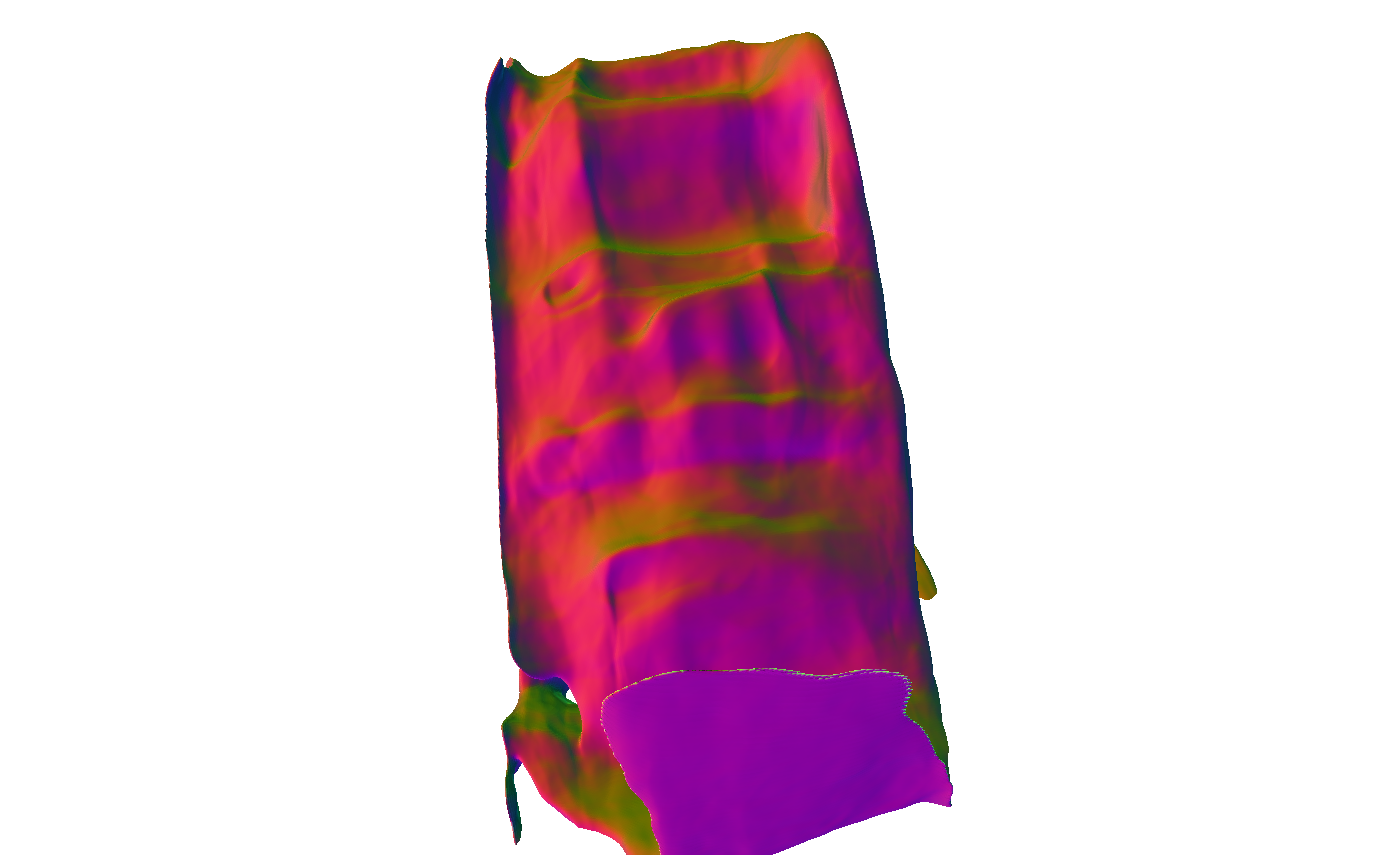} &
    \includegraphics[width=0.24\linewidth,trim={100, 40, 100, 0},clip]{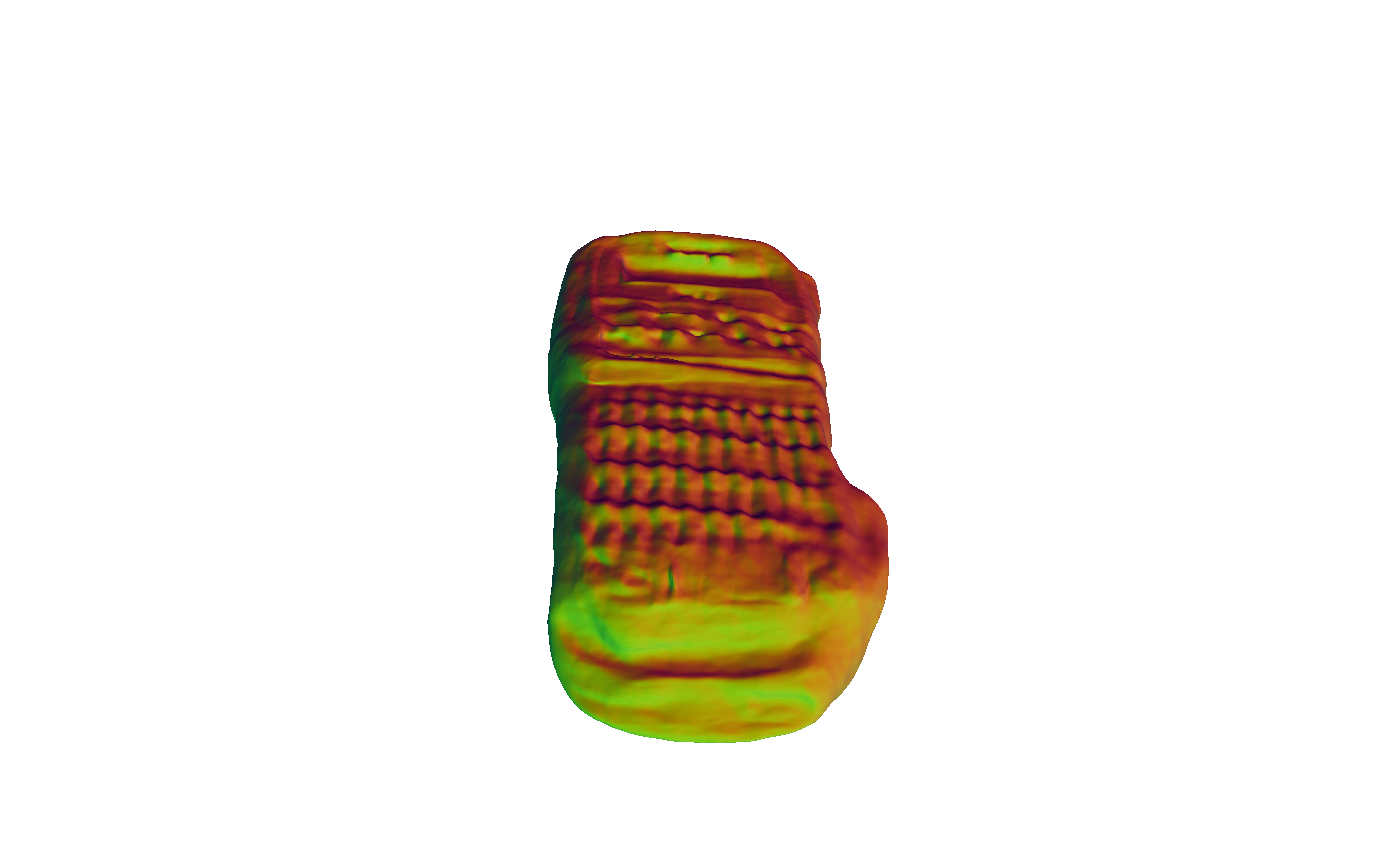}&
        \includegraphics[width=0.18\linewidth]{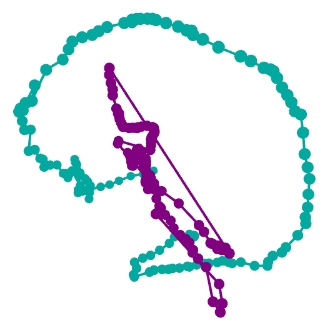} \\
    
    \includegraphics[width=0.24\linewidth]{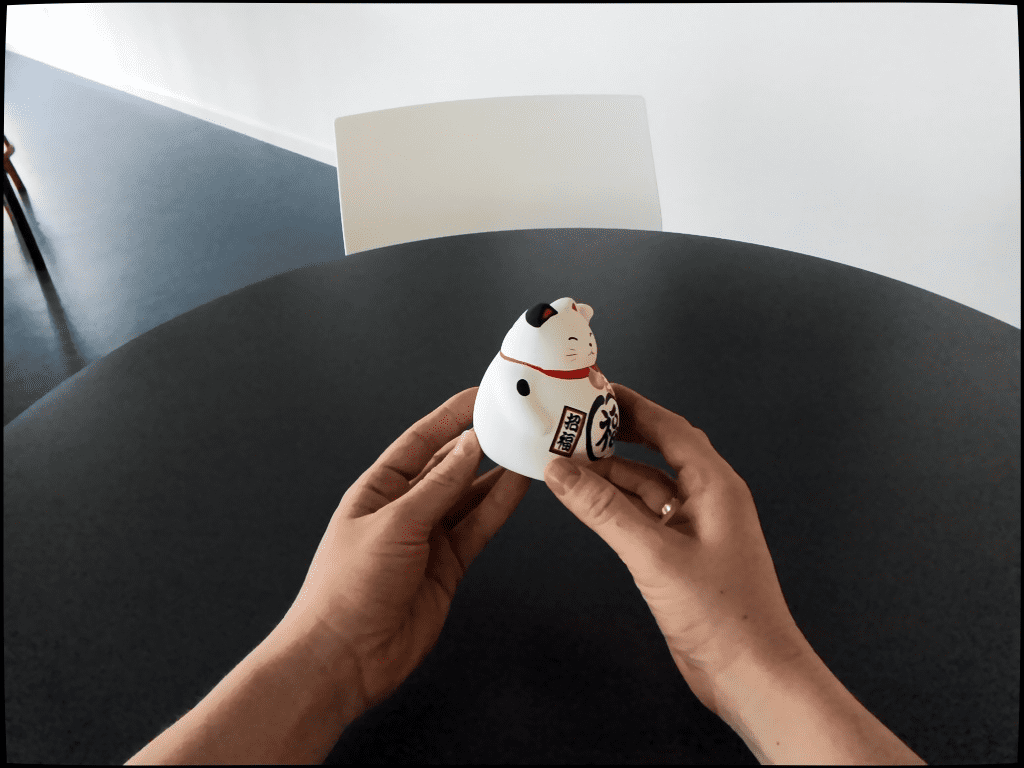} &
    \includegraphics[width=0.24\linewidth,trim={0, 0, 0, 30},clip]{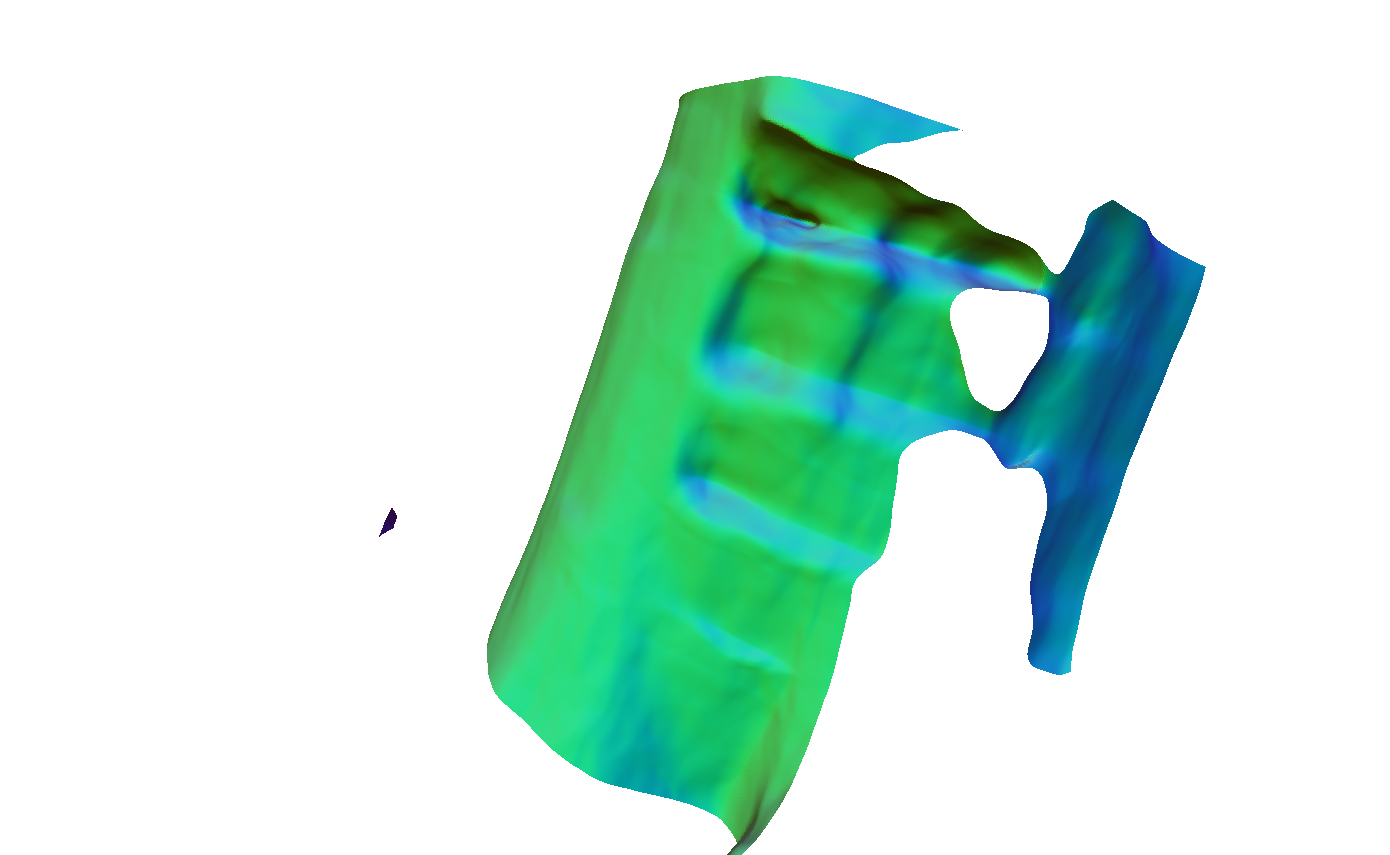} &
    \includegraphics[width=0.24\linewidth,trim={100, 30, 100, 0},clip]{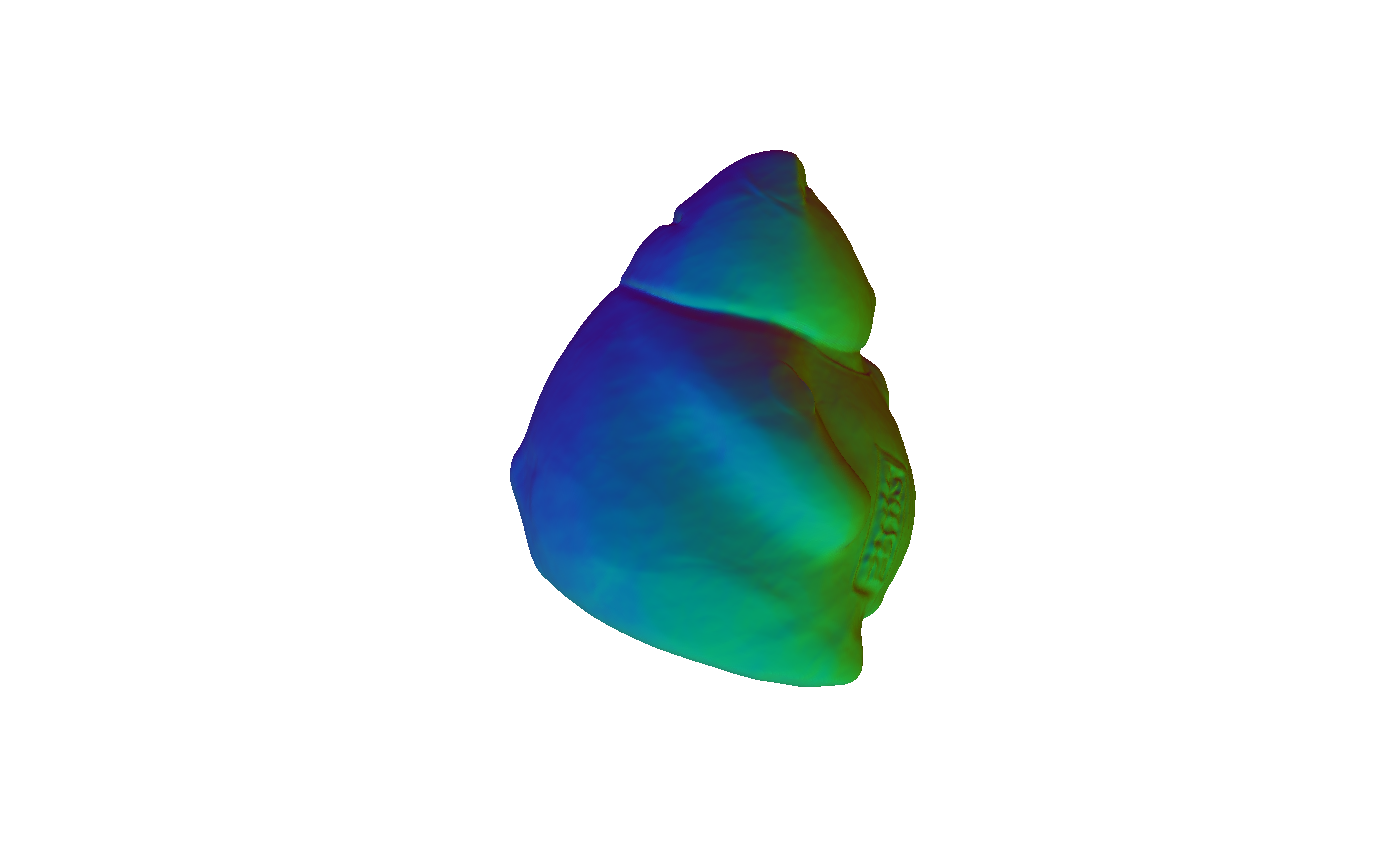} &
        \includegraphics[width=0.20\linewidth]{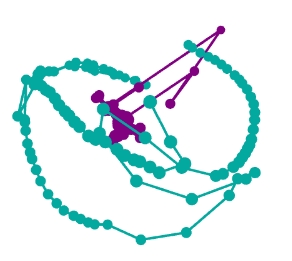} \\
    
    \includegraphics[width=0.24\linewidth]{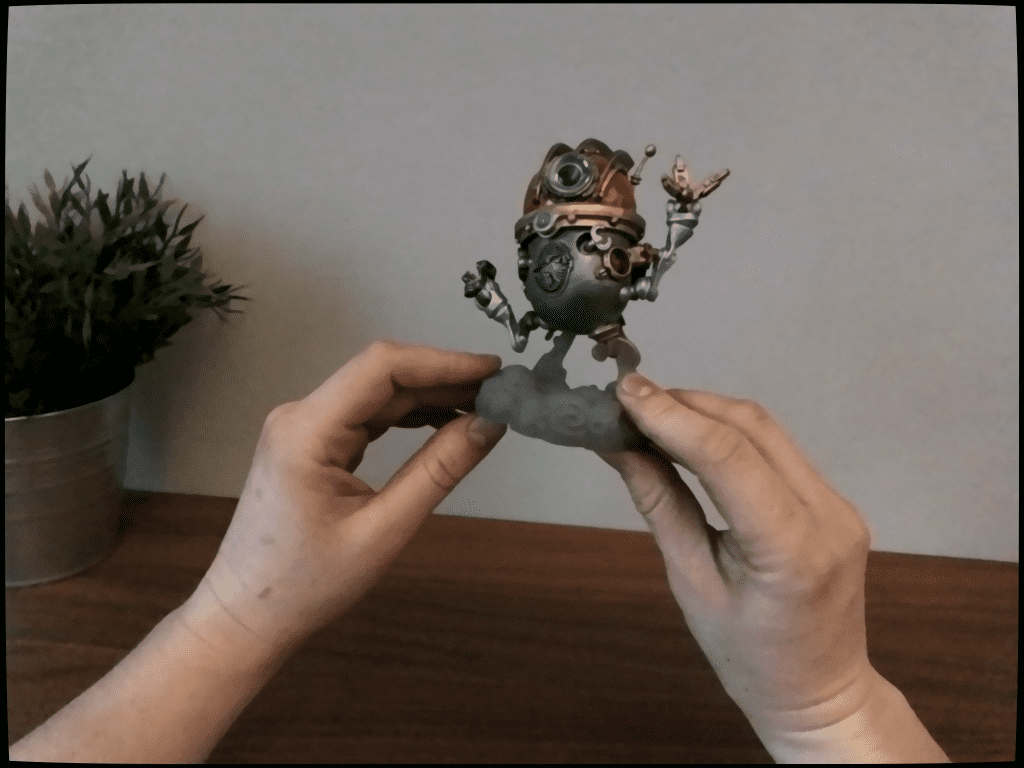} &
    \includegraphics[width=0.24\linewidth,trim={0, 0, 0, 20},clip]{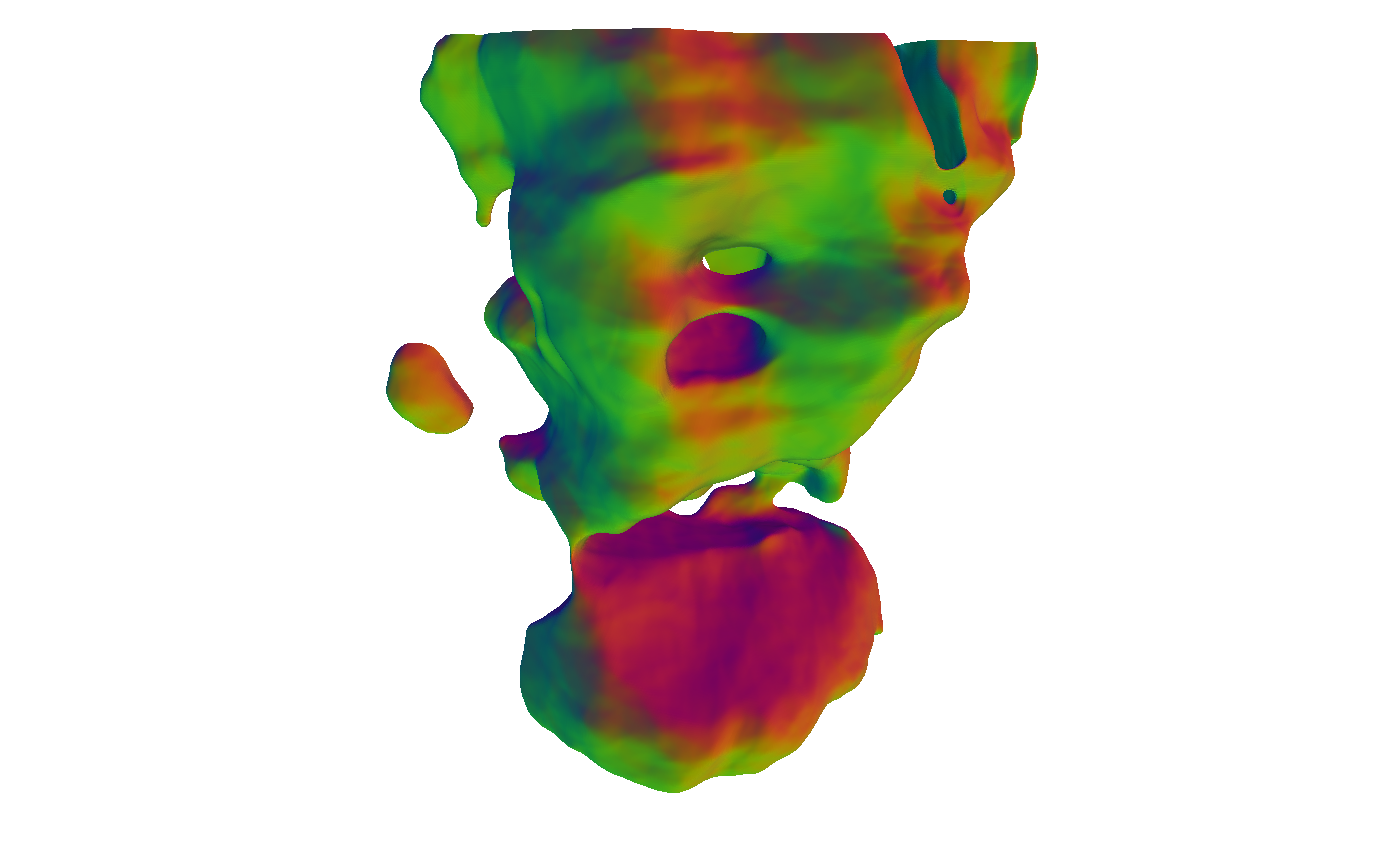} &
    \includegraphics[width=0.24\linewidth,trim={100, 30, 100, 0},clip]{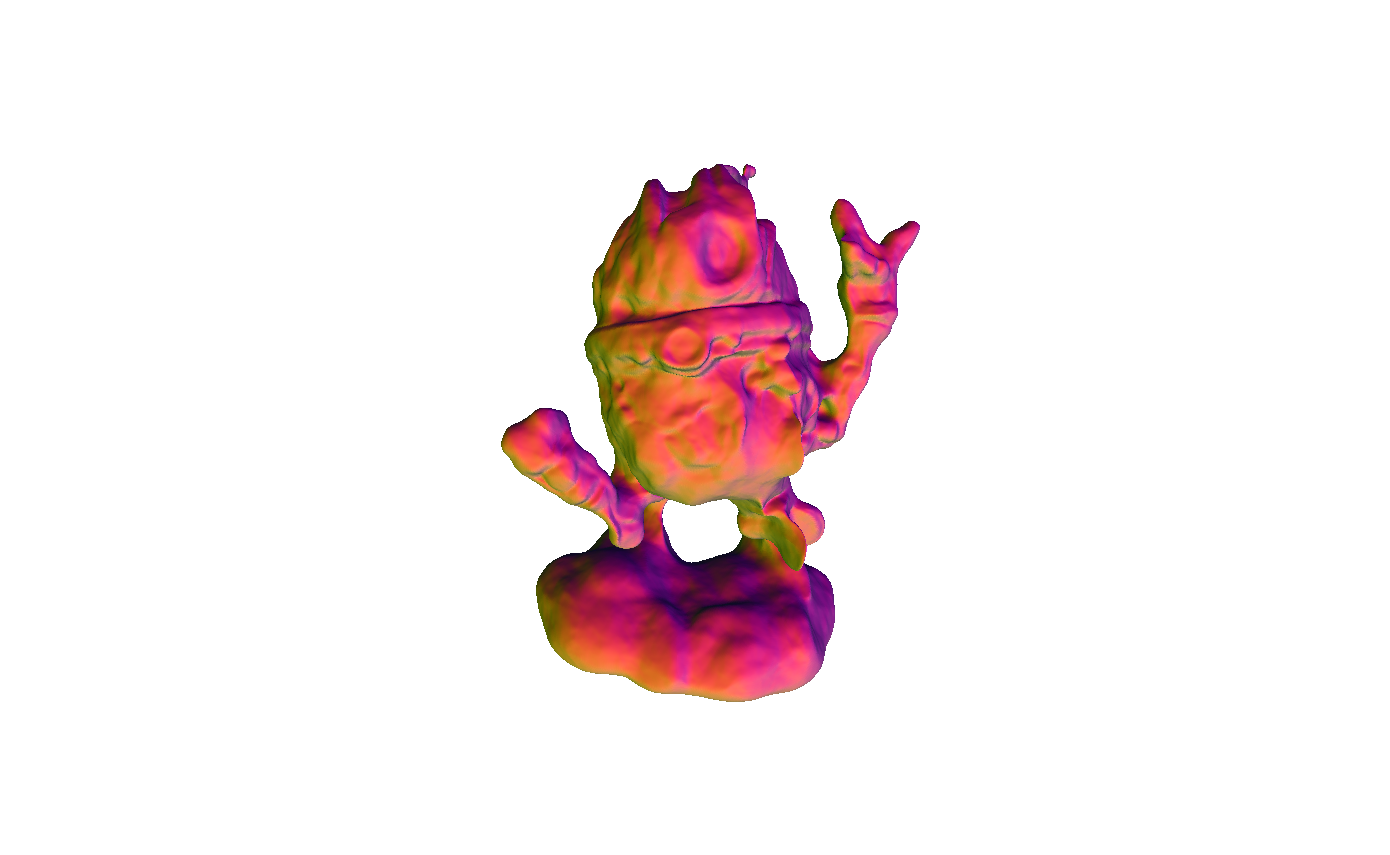}&
        \includegraphics[width=0.20\linewidth]{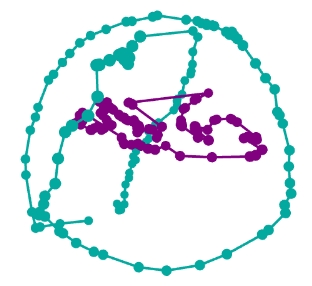} \\
    \small{Reference} & \small{BARF} & \small{Ours} & \small{Poses} \\
    \end{tabular}
    \endgroup
  \caption{\textbf{Results on egocentric sequences.}
  We visualize the shape results of BARF~\cite{barf} and ours in the middle, and compare their pose results in the last column~(\textcolor[RGB]{128,0,128}{purple}: BARF, \textcolor[RGB]{3,168,158}{cyan}: ours).
  Without proper pose initialization for free-moving objects, BARF struggles to produce reasonable results. Our method generates accurate pose and shape results in most cases.
  }
  \label{fig:evaluation_on_ml}
\end{figure}

\subsection{Results on Egocentric Sequences}

To verify the generalization ability of our method, we capture some sequences with a head-mounted device with egocentric views~(Magic Leap 2~\cite{magcileap2}), where the camera is naturally moving with the user’s head and the target is freely manipulated by the user.
We illustrate the performance of our approach in Fig.~\ref{fig:evaluation_on_ml}.
Note that our method only relies on RGB images and does not use head pose or depth images provided by the device.
The result shows that our method generalizes well to this real setting, and produces accurate results for free-moving daily objects.

\subsection{Limitation}

\begin{figure}[htbp]
  \centering
      \begingroup
  \setlength{\tabcolsep}{1pt}
    \begin{tabular}{cccc}
    \includegraphics[width=0.24\linewidth]{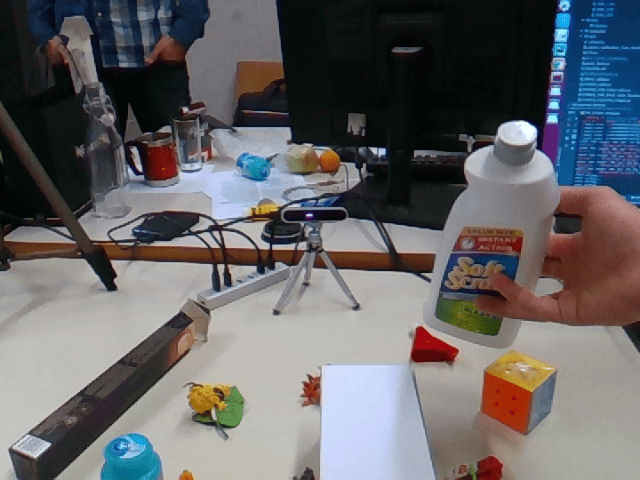} &
    \includegraphics[width=0.24\linewidth,trim={100, 0, 100, 0},clip]{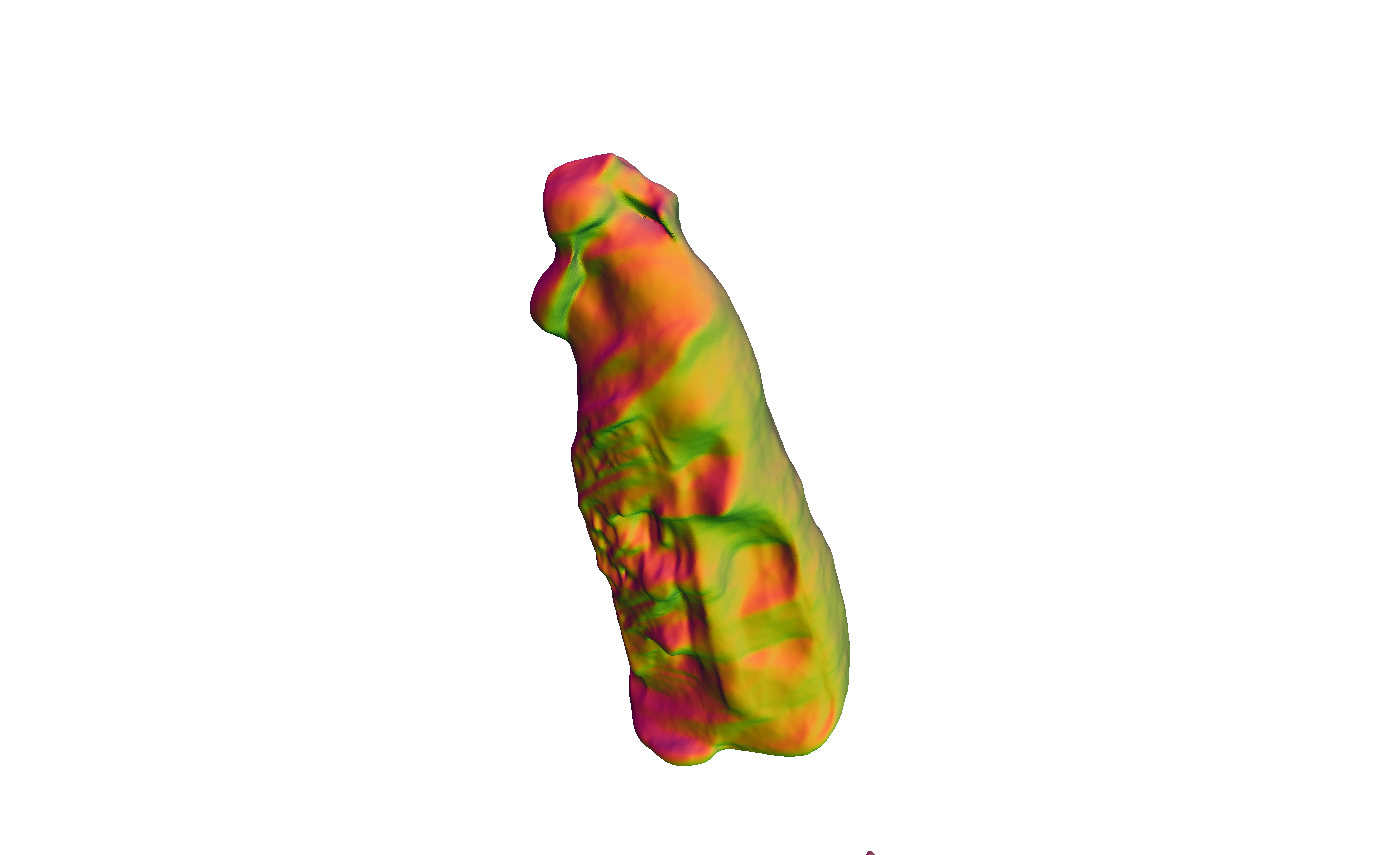} &
    \includegraphics[width=0.24\linewidth]{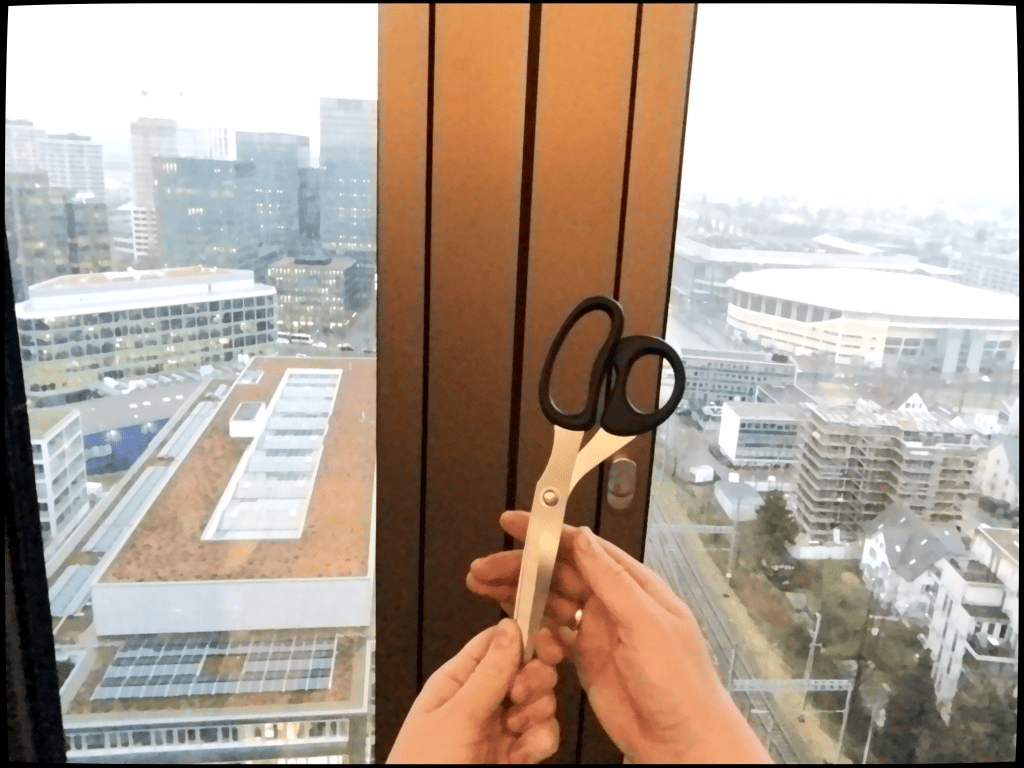} &
   \includegraphics[width=0.24\linewidth,trim={100, 0, 100, 0},clip]{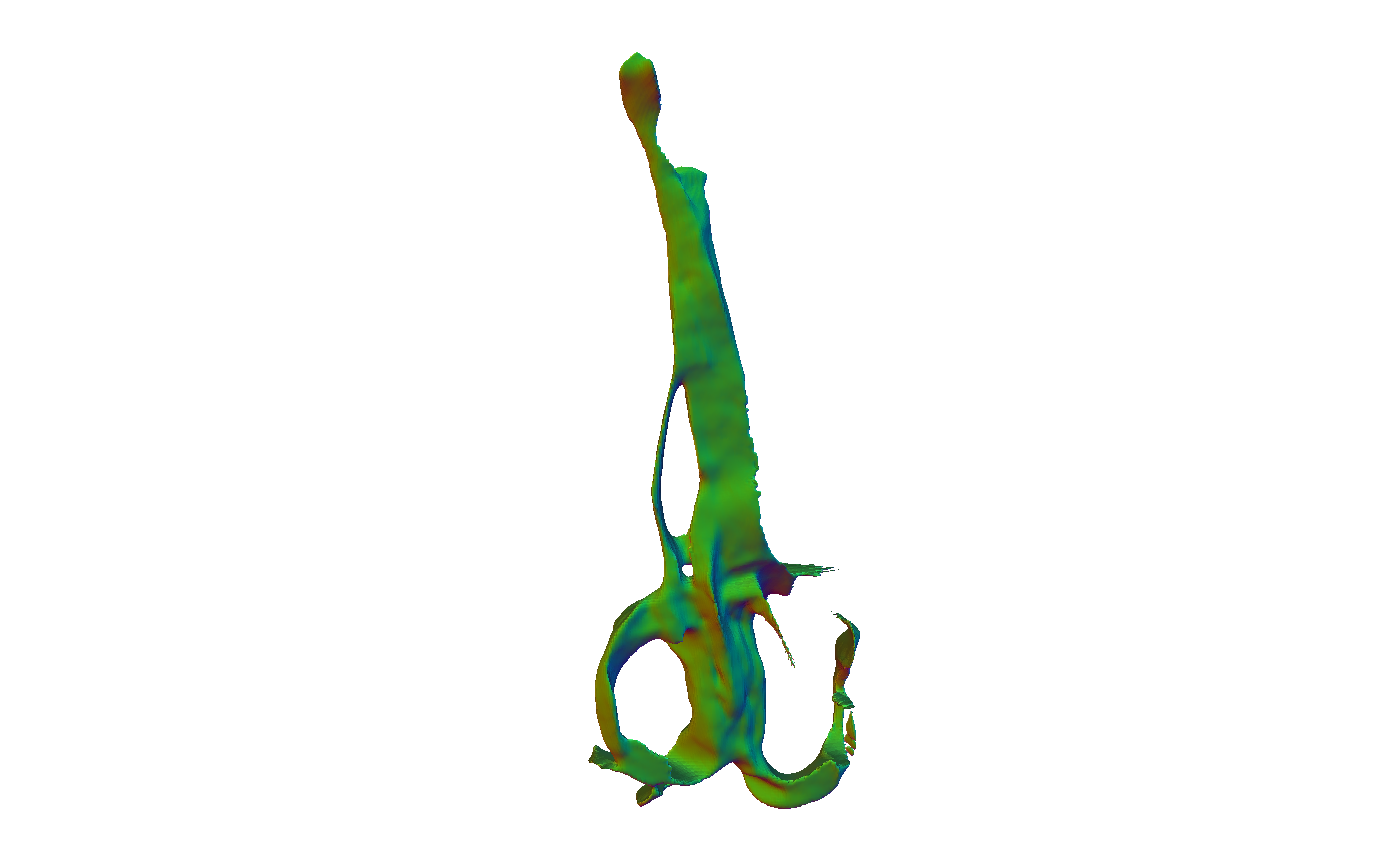} 
   \\
    \end{tabular}
    \endgroup
  \caption{\textbf{Failure cases.} The first two figures show an example where the side of the bleach cleaner is occluded for a long time caused by the firmly holding during the whole capture, resulting in deteriorated meshes. The last two figures show an example of a tiny texture-less object, where our method can generate reasonable results but still has a gap from perfection.
  }
  \label{fig:failure_cases}
\end{figure}

Although our method produces accurate pose and mesh results in most cases, it can not handle scenarios where some parts of the object is occluded for a long time during capture, as shown in Fig.~\ref{fig:failure_cases}. On the other hand, our method still can not produce accurate results for tiny texture-less object due to the lacking of enough features. Addressing this will be one of our future work.


%% file: sec/05_conclusion.tex
\section{Conclusion}

We have showed that it is possible to jointly optimize the reconstruction and pose estimation for free-moving objects without relying on any prior information, or any segmenting procedure from a monocular RGB video. It relies on the intuition that, using estimated 2D object masks, one can reformulate the joint optimization problem w.r.t. a virtual camera pointing to the object center, which simplifies the trajectory and reduces the search space of optimization significantly. Although the pose and shape are not physically-compliant in the virtual camera system, our experiments have demonstrated that optimizing w.r.t. the virtual camera yields robust initialization results and produces accurate final results after a refinement w.r.t. the real camera. Future work will focus on making the method more robust and more generic so that it can be used in more scenarios.